\pdfoutput=1

\documentclass[11pt]{article}

\usepackage[preprint]{acl}

\usepackage{times}
\usepackage{latexsym}

\usepackage[T1]{fontenc}

\usepackage[utf8]{inputenc}

\usepackage{microtype}

\usepackage{inconsolata}

\usepackage{graphicx}

\usepackage{epsfig}
\usepackage{graphicx}
\usepackage{amsmath}
\usepackage{amssymb}
\usepackage{xspace}
\usepackage{amstext}
\usepackage{multirow}
\usepackage{tabularx}
\usepackage{booktabs}   
\usepackage{tcolorbox}
\usepackage{xcolor}
\usepackage{subcaption}
\usepackage{colortbl}
\usepackage{hyperref}
\usepackage{todonotes}
\usepackage{placeins}
\usepackage{wrapfig}
\usepackage{longtable}
\usepackage{listings}

\newcommand{\rparagraph}[1]{\vspace{1.4mm}\noindent\textbf{#1.}}

\newcommand{\sparagraph}[1]{\vspace{0.0mm}\noindent\textbf{#1.}}

\newcommand{\modelname}{\texttt{Centurio}\xspace}

\newtcolorbox{promptbox}[1]{colback=gray!5!white,colframe=gray!75!black,fonttitle=\bfseries\scriptsize,fontupper=\ttfamily\footnotesize,title=#1}

\newtcolorbox{examplebox}[1]{colback=gray!5!white,colframe=gray!75!black,fonttitle=\bfseries\scriptsize,fontupper=\rmfamily\footnotesize,title=#1}

\title{\texttt{Centurio}: On Drivers of Multilingual Ability of \\ Large Vision-Language Model}

\author{\textbf{Gregor Geigle$^{12*}$ \quad Florian Schneider$^{3*}$\quad  Carolin Holtermann$^{4}$} \\
\textbf{Chris Biemann$^{3}$ \quad Radu Timofte$^{2}$ \quad Anne Lauscher$^{4}$ \quad Goran Glava\v{s}$^{1}$} \\
  $^{1}$W{\"u}NLP, $^{2}$Computer Vision Lab, CAIDAS, University of W{\"u}rzburg  \\
  $^{3}$Language Technology Group, $^{4}$Data Science Group, University of Hamburg  \\
  \texttt{(gregor.geigle|florian.schneider-1)@uni-(wuerzburg|hamburg).de} \\
  \href{https://gregor-ge.github.io/Centurio/}{gregor-ge.github.io/Centurio}
  }

\begin{document}
\maketitle
\def\thefootnote{*}\footnotetext{\,Equal contribution.}\def\thefootnote{\arabic{footnote}}
\begin{abstract}
Most Large Vision-Language Models (LVLMs) to date are trained predominantly on English data, which makes them 
struggle to understand non-English input and fail to generate output in the desired target language.
Existing efforts mitigate these issues by adding multilingual training data, but do so in a largely ad-hoc manner, lacking insight into how different training mixes tip the scale for different groups of languages. 
In this work, we present a comprehensive investigation into the training strategies for massively multilingual LVLMs. 
First, we conduct a series of multi-stage experiments spanning 13 downstream vision-language tasks and 43 languages, systematically examining:
(1) the number of training languages that can be included without degrading English performance and
(2) optimal language distributions of pre-training as well as (3) instruction-tuning data. 
Further, we (4) investigate how to improve multilingual text-in-image understanding, and introduce a new benchmark for the task. 
Surprisingly, our analysis reveals that one can (i) include as many as 100 training languages simultaneously (ii) with as little as 25-50\% of non-English data, to greatly improve multilingual performance while retaining strong English performance.
We further find that (iii) including non-English OCR data in pre-training and instruction-tuning is paramount for improving multilingual text-in-image understanding.
Finally, we put all our findings together and train \modelname, a 100-language LVLM, offering state-of-the-art performance in an evaluation covering 14 tasks and 56 languages.
%
\end{abstract}

\section{Introduction}

\begin{figure*}[ht!]
    \centering
    \includegraphics[width=0.9\linewidth,trim={0 4.8cm 0 4.8cm},clip]{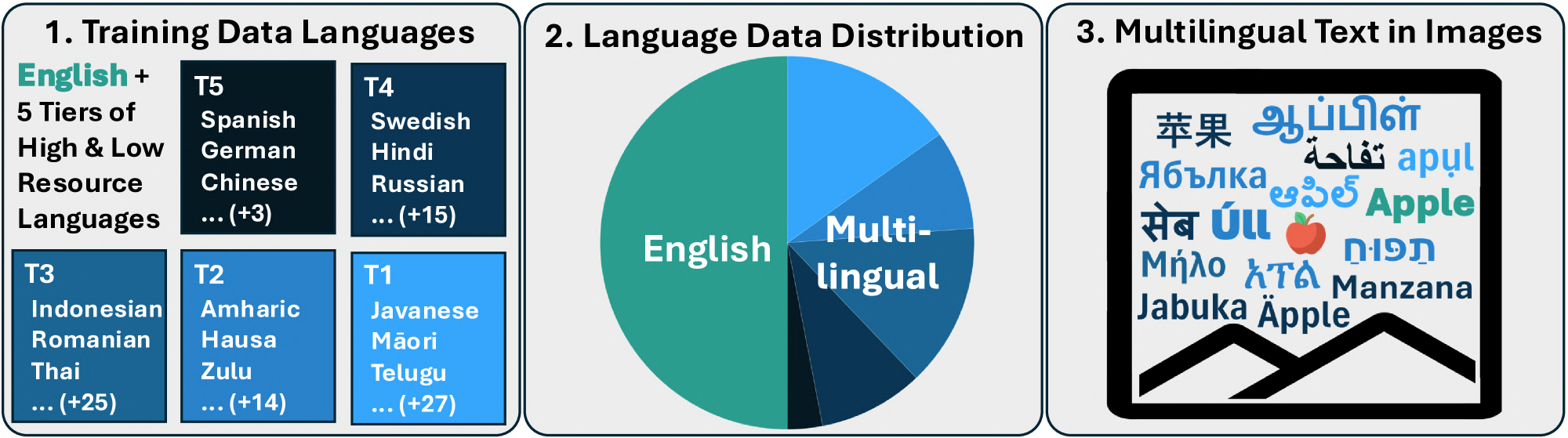}
    \caption{
    Exploring drivers of multilingual ability: 
    (1) Languages in the training data; 
    (2) Distribution of languages in the training data; 
    (3) Incorporating multilingual OCR samples to understand non-English text in images.
    }
    \label{fig:enter-label}
    \vspace{-0.5em}
\end{figure*}
Large Vision-Language Models (LVLMs) extend Large Language Models (LLMs) \cite{brown_language_2020} to natively understand images as input \cite{li_blip-2_2023,liu_visual_2023}.
This leverages the impressive language generation and reasoning abilities of recent LLMs \cite{llama_team_llama_2024,yang_qwen2_2024} for vision-language tasks like image captioning or visual question answering. 



However, most models are trained with just English data \cite{liu_improved_2023,dai_instructblip_2023,liu_llava-next_2024}.
This limits the access for speakers of other languages as the resulting models have several limitations even if the underlying LLMs exhibit multilingual capabilities: 
the models fail to understand non-English instructions \cite{schneider_m5_2024}, struggle with non-English text content in images \cite{tang_mtvqa_2024}, and often fail to reply in the correct language, i.e., they have problems with \textit{language fidelity} \cite{hinck_why_2024}.
%
%
To ameliorate these issues, LVLMs need to be trained on a multilingual data composition.
As the amount of data one can train on is always limited---by time, computing resources, financial costs, or other constraints---an effective distribution of the data across different languages is paramount. Existing multilingual LVLM work has, however, given minimal consideration to this central question of optimal training data composition \cite[e.g.,][]{geigle_mblip_2023-1,sun_parrot_2024,maaz_palo_2024}.
%

In this work, we comprehensively investigate the space of language distributions of LVLM training mixes, focusing on the presumed trade-off between the number of included languages 
and performance across languages---grouped by the amount of data available for them---under a \textit{fixed training budget}.  
%
We train several models with different data compositions obtained by machine-translating high-quality English data and benchmark them across 13 downstream tasks covering 43 diverse languages---from low-resource languages like Igbo to high-resource languages like German.
We focus on four research questions, each building on the previous one, designed to identify an optimal multilingual training mix: 
\textbf{RQ1:} What is the optimal number of training languages?
\textbf{RQ2 \& RQ3}: What is the optimal distribution of data across languages in (RQ3) pre-training data and (RQ2) instruction-tuning?
\textbf{RQ4:} How to improve the understanding of multilingual text in images?
To measure progress for RQ4, we introduce SMPQA (Synthetic Multilingual Plot Question Answering), a novel dataset for testing multilingual OCR 
capabilities, spanning 11 languages and 7 scripts.

Our findings are encouraging, albeit surprising.

\noindent\textbf{1.} We do not observe the infamous ``curse of multilinguality'' \cite{conneau_unsupervised_2020,pfeiffer2022lifting} and find that gradually increasing the number of languages incurs only a negligible ``performance tax'': scaling from 7 to 100 languages greatly improves performance for languages newly added to the training data, especially with respect to language fidelity, while largely retaining performance levels for all previously added languages.

\noindent\textbf{2.} We find that exposure to a language matters more than increasing the training portion of the language or, in particular, that the majority of the training data can still be in English, which lowers the cost of acquiring training data in other languages (e.g., via machine translation). Concretely, we find that turning 25 to 50\% of training data multilingual 
yields strong performance, with more data 
sometimes even degrading performance; in \textit{pre-training}, having a larger share of multilingual data is more beneficial, but it also saturates after 50\%.

\noindent\textbf{3.} We obtain mixed results for text-in-image problems: while incorporating (synthetic) OCR data with 5k samples per language rapidly boosts the performance for Latin-script languages, the same does not hold for languages with other scripts. 

\textbf{Finally}, to demonstrate the practical impact of our findings, we train \modelname, a massively multilingual LVLM with 100 languages, following what we found to be an ``optimal'' data distribution across languages for both training stages. 
\modelname achieves state-of-the-art results over 14 tasks
, matching the performance of popular multilingual open-weight LVLMs like Qwen2-VL \cite{wang_qwen2-vl_2024}, InternVL 2.5 \cite{chen2024expandingperformanceboundariesopensource} and Pangea \cite{yue_pangea_2024} on English and other high-resource languages while outperforming them on low(er)-resource languages.


\section{Drivers of Multilingual Ability}
\label{sec:analysis}
The design space for training (multilingual) LVLMs is extensive, ranging from the choice of the image encoder and the alignment module between the image encoder and LLM to the selection of training data. 
\cite{karamcheti_prismatic_2024,laurencon_what_2024,tong_cambrian-1_2024}.
Exhaustively searching through the cross-product of all choices 
 is not feasible. 
In this work, we focus on extensive evaluation of language distributions of training data in both pre-training and instruction-tuning.
Intuitively, this should be a major factor affecting the multilingual ability of an LVLM. 
Figure \ref{fig:enter-label} illustrates the scope of our analysis. We keep adding groups of languages---from highest- to lowest-resourced, following the ``resourceness'' tiers of \cite{joshi_state_2020}---into the training mix while keeping the data size fixed. 
Besides the number of languages, our main focus is on the division of the training budget between English and all other languages.
Finally, we posit that, besides understanding instructions and generating outputs in different languages, truly multilingual LVLMs must be able to ``understand'' multilingual text in images.
We thus pay special attention to training adaptions for multilingual text-in-image problems. 

\subsection{Experimental Setup}
\label{sec:analysis_ex_setup}
\sparagraph{Architecture}
%
For our experiments, we adopt the popular LLaVA architecture \cite{liu_visual_2023,liu_improved_2023}:
An image encoder (\texttt{SigLIP SO400/384} \cite{zhai_sigmoid_2023}) encodes images into a sequence of visual tokens which are projected with a $2$-layer MLP into the LLM embedding space; these 
tokens are then concatenated to the text tokens and fed to the LLM. 
We choose \texttt{Phi 3.5} \cite{abdin_phi-3_2024} as our LLM because it exhibits strong multilingual performance while its small size ($3.8$B parameters) allows for more computationally efficient experimentation. 
%
To show that our findings generalize to other LLMs, we repeat 
a subset of the analysis experiments with \texttt{Llama 3} ($8$B) \cite{llama_team_llama_2024} as the LLM backbone (see Appendix \ref{sec:appendix:experiments:analysis_llama}).  

\rparagraph{Training Setup}
Following previous work \cite{liu_improved_2023,tong_cambrian-1_2024}, we split the training into two phases : 
\textit{1) pre-training:} the model is trained only on image captioning, with dense image captions;
\textit{2) instruction tuning:} the model is trained on a mix of diverse vision-language tasks using several 
public datasets.
While pre-training benefits downstream performance, it is not strictly necessary for the LVLM to perform well on the downstream tasks \cite{karamcheti_prismatic_2024}.
To reduce the computational cost of our analysis (i.e.,  to avoid coupling each language distribution over pre-training data with every language distribution of instruction-tuning data), we skip pre-training while searching for an optimal language distribution for instruction tuning.
Then, with instruction-tuning data fixed, we search for an optimal language distribution for pre-training data.   
In both stages, we freeze the image encoder and only update the MLP and LLM (with LoRA \cite{hu_lora_2022-1}) weights.
We provide further details in Appendix~\ref{sec:appendix:training}.

\rparagraph{Training Data}
Our controlled experiments require comparable data over a wide range of languages.
Existing multilingual datasets, available only for some tasks and in a handful of languages\footnote{See, for example, datasets collected by \citet{yue_pangea_2024}} thus do not meet our needs.
Instead, we resort to machine translation (MT) and use the open NLLB model~\cite{costa-jussa_no_2022}\footnote{\texttt{nllb-200-distilled-1.3B} } to translate readily available English datasets.\footnote{We do \textbf{not} translate text-in-image datasets as that would result in mismatches between the instruction/output language and the English text in the image.}
While MT results in lower data quality, especially for lower-resource languages, it is the only option to obtain multilingual vision-language training data at scale.
Moreover, gains from ``low-quality'' MT data are guaranteed to be met or even surpassed with higher-quality translations (e.g., commercial MT or human translators).
%
%
Our instruction-tuning data is adapted from LLaVA-Next \cite{liu_llava-next_2024} and contains 0.77M samples.
For pre-training, we use the $1.3$M dense captions from ShareGPT4v \cite{chen_sharegpt4v_2024}.
We provide further details in Appendix~\ref{sec:appendix:train_data}.

\rparagraph{Evaluation}
We curate an extensive test suite of $13$ tasks covering $43$ languages to assess the multilingual abilities of our models.
Following~\citet{joshi_state_2020}, we cluster the tested languages into five \textit{tiers}, with \textit{T5} encompassing the high-resource languages (e.g., German, Chinese) and \textit{T1} extremely low-resource languages (e.g., Maori, Telugu).  
The tasks contained in our test suite are twofold:
(1) \textit{discriminative} tasks with questions that require binary (''yes/no'') or multiple-choice answers and 
(2) \textit{open generation} tasks, where the models need to generate output in the target language (e.g., an image caption or a free-form answer).
Generative tasks additionally evaluate
a model's \textit{language fidelity}, i.e., the ability to generate the answer in the language of the instruction.
The full list of evaluation tasks and languages, along with further details, is in Appendix~\ref{sec:appendix:evaluation}. 
We report the results for language tiers (\textit{T1--T5}), averaging the scores over all tasks and all tier languages.\footnote{While tasks use different measures, all are on the 0-100\% scale, so no task skews the average.}
We separately report English performance and \textbf{exclude} it from \textit{T5}.

\subsection{RQ1: Number of Training Languages}
\label{sec:analysis:ex1}
We first investigate on \textit{how many} languages to actually train with: does training on few high-resource languages and (zero-shot) cross-lingual transfer to unseen languages suffice, as suggested, e.g., by \citet{shaham_multilingual_2024,chen-etal-2024-monolingual,kew_turning_2023}, or do we need to explicitly include each targeted languages? 
Conversely, does training with more languages harm the per-language performance, with a smaller portion of the training data now allocated to each language?
%

\rparagraph{Setup}
%
We focus on the instruction-tuning step: 50\% of the data is kept in English\footnote{More specifically, 50\% of the 80\% of non-text-in-image data, which is excluded from translation.}, while the other 50\% split between $N$ other languages equally, i.e., each language gets $\frac{50}{N}\%$ of the data budget.
We gradually increase $N$, starting with the highest-resource tier (T5) and then including tiers of lower-resource languages (T4 to T1), one at a time.
This results in the following setups:
\textbf{T5} ($N=6$),
\textbf{T5-T4} ($N=24$),
\textbf{T5-T3} ($N=52$),
\textbf{T5-T2} ($N=69$), and finally
\textbf{L100} ($N=99$).
In \textbf{L100}, in addition to languages from \textbf{T5-T2}, we include T1 languages 
to cover XM3600 \cite{thapliyal_xm3600_2022} and otherwise randomly to reach 99 languages.

\begin{table}[t]
    \centering
\footnotesize
\begin{subtable}{\linewidth}
 \def\arraystretch{0.9}
 \resizebox{0.99\linewidth}{!}{
\begin{tabular}{lrrrrrr}
\toprule
\bf Train Lang. & \bf T1 & \bf T2 & \bf T3 & \bf T4 & \bf T5 & \bf en \\
\midrule
\multicolumn{5}{l}{\bf All tasks} \\ \midrule
English & 14.4 & 30.4 & 24.4 & 23.6 & 28.5 & 53.6 \\
T5 & 16.5 & 31.0 & 26.3 & 26.7 & 34.0 &  \underline{53.7} \\
T5-4 & 17.4 & 30.6 & 27.9 & \textbf{29.6} & 33.5 & 51.5 \\
T5-3 & \underline{17.7} & 31.4 & \textbf{32.1} & \underline{29.0} & \underline{34.1} &  52.7 \\
T5-2 & 17.0 & \textbf{34.5} & 30.0 & 28.2 & 33.4 & \textbf{54.1} \\
L100 & \textbf{19.3} & \underline{32.6} & \underline{30.7} & 28.9 & \textbf{34.4} & 52.6 \\
\midrule
\multicolumn{7}{l}{\bf Tasks unaffected by language fidelity} \\ \midrule
English & 33.0 & 32.5 & 36.3 & 38.5 & \textbf{42.9} & 55.7 \\
T5 & 35.3 & 33.2 & 36.4 & 38.7 & 42.4 & \underline{56.0} \\
T5-T4 & 35.8 & 32.6 & 37.8 & \textbf{40.1} & 42.2 &  55.7 \\
T5-T3 & \underline{35.9} & 33.6 & \textbf{40.5} & 39.7 & 42.6 &  \textbf{56.3} \\
T5-T2 & 35.2 & \textbf{36.5} & 38.5 & 39.5 & \underline{42.8} & 55.5 \\
L100 & \textbf{36.1} & \underline{34.3} & \underline{39.1} & \underline{39.8} & 42.7 & 54.6 \\
\bottomrule
\end{tabular}
}
\caption{
Scores are averaged over results from all tasks grouped by language tier.
The performance on the following tasks is affected by language fidelity: XM3600, MaXM, MTVQA.}
\label{tab:ex1:average}
\end{subtable}

\vspace{0.5em}

\begin{subtable}{\linewidth}
\centering
\def\arraystretch{0.9}
\resizebox{0.99\linewidth}{!}{
\begin{tabular}{lrrrrrr}
\toprule
\bf Train Lang. & \bf T1 & \bf T2 & \bf T3 & \bf T4 & \bf T5 & \bf en \\
\midrule
English & 0.2 & 0.2 & 0.1 & 2.4 & 6.2 & \underline{100.0} \\
T5 & 39.1 & 36.1 & 82.2 & 83.9 & \textbf{99.1} & \underline{100.0} \\
T5-T4 & 61.8 & 84.6 & 87.5 & \textbf{99.2} & \underline{98.4} & \underline{100.0} \\
T5-T3 & \underline{72.9} & 84.4 & \textbf{98.2} & 95.2 & 97.9 & \underline{100.0} \\
T5-T2 & 68.5 & \textbf{99.0} & \underline{97.9} & \underline{98.4} & 98.1 & \underline{100.0} \\
L100 & \textbf{72.9} & \underline{98.2} & 95.4 & 97.8 & 98.2 & \underline{100.0} \\
\bottomrule
\end{tabular}
}
\caption{Average language fidelity on XM3600 in \%.}
\label{tab:ex1:fidelity}
\end{subtable}

\caption{\textbf{RQ1} (\S\ref{sec:analysis:ex1}) results for models trained with different sets of languages.
We emphasize the \textbf{best} and \underline{second-best} result in each column.
\vspace{-0.5em}
}
\label{tab:ex1}
\end{table}

\rparagraph{Results}
Table~\ref{tab:ex1} summarizes the results.
Expectedly, we find that including a language (tier) in instruction-tuning improves their performance (Table~\ref{tab:ex1:average}, top half).
Nevertheless, the (negative) effect of adding new languages on performance of previously included languages is negligible, if at all present. 
This makes training massively multilingual LVLMs feasible with only minor performance drawbacks for any given language.
In-language training leads to dramatic improvements in language fidelity (i.e., the model producing the output in the correct language), as shown in Table~\ref{tab:ex1:fidelity}. 
Interestingly, the more multilingual the training, the larger the fidelity gains also for languages not included in training; explicit in-language training, expectedly, then further improves fidelity for any given language (see Table~\ref{tab:full:xm3600_fidelity} in the Appendix for detailed per language results).
Even when excluding tasks where language fidelity plays a role (Table~\ref{tab:ex1:average} bottom), we observe the same trends: 
steady improvements from in-language training, with negligible (if any) performance drops for other languages.
A subset of experiments with Llama 3 (setups: English, T5, and L100) in Table~\ref{tab:ex1:average_llama} in the Appendix confirms these trends observed with Phi 3.5: in fact, we see even larger gains over all tasks when training with more languages. 

\subsection{RQ2: Language Distribution in Instruction-Tuning}
\label{sec:analysis:ex2}
RQ1 experiments show that massively multilingual instruction-tuning data is beneficial across the board. 
We now analyze \textit{how much} of the training data should be multilingual.
On the one hand, intuitively, increasing the non-English portion of the training data budget could then lead to further gains. 
On the other hand, the gains from more multilingual training are, at some point, likely to be offset by the fact that we are adding noisy (MT-obtained) data at the expense of clean (English) data. 

\rparagraph{Setup}
We opt for the full set of $100$ languages (\textit{L100}) in this experiment due to their robust multilingual performance.
%
However, we adjust the language distribution by keeping $E$\% of the data budget in English and splitting the remaining $100-E$\% equally across the other 99 languages\footnote{We observed no benefits from an unequal allocation that up-samples low(er)-resource languages; see \S\ref{sec:appendix:experiments:stratified})}.
We consider the following six setups: $E \in \{1, 10, 25, 50, 75, 90\}$.

\begin{table}[]
    \centering
\footnotesize
 \def\arraystretch{0.97}
 \resizebox{0.99\linewidth}{!}{
\begin{tabular}{lrrrrrr}
\toprule
\bf English \% & \bf T1 & \bf T2 & \bf T3 & \bf T4 & \bf T5 & \bf en \\
\midrule
1 & 19.1 & 30.3 & 28.8 & 27.1 & 31.7 & 48.9 \\
10 & 18.1 & 32.4 & 29.4 & 27.4 & 32.5 & 50.1 \\
25 & \textbf{19.7} & \textbf{35.5} & 29.9 & 27.9 & 33.0 & 50.3 \\
50 & \underline{19.3} & \underline{32.6} & \underline{30.7} & \textbf{28.9} & \underline{34.4} & 52.6 \\
75 & 18.5 & 31.5 & \textbf{30.7} & \underline{28.4} & \textbf{34.6} & \underline{54.1} \\
90 & 15.9 & 31.2 & 27.6 & 26.9 & 34.1 &  \textbf{54.8} \\
\bottomrule
\end{tabular}
}
    \caption{\textbf{RQ2} (\S\ref{sec:analysis:ex2}) results for models trained with different ratios of English to multilingual data in the instruction-tuning phase.
    Scores are averaged over results from all tasks grouped by language tier.
    }
    \label{tab:ex2}
    \vspace{-0.5em}
\end{table}


\rparagraph{Results}
We present the results in Table~\ref{tab:ex2}.
We observe peak performance for all language tiers when between 25\% and 75\% of the training data is in English. 
%
For some tasks (e.g., XM3600, MaXM, BIN-MC), we see weaker performance with more English data, while for others (e.g., MTVQA, xGQA, MaRVL) more multilingual data leads to slight performance drops (see per-task results in \ref{sec:appendix:fullresults:analyis}).
Overall, lower-resource languages benefit from more multilingual data and, conversely, higher-resource languages benefit from more English data.
However, this is in part also a consequence of language coverage across tasks: XM3600 and BIN-MC profit from a more multilingual training mix; at the same time, they are the tasks that encompass the most low(er)-resource languages. 
%

%
Results obtained with the Llama 3 backbone (see Table~\ref{tab:ex2:average_llama} in the Appendix) follow the same pattern: we observe the best performance in T1 and T2 with $E=10$; and for T5 and English with $E=90$;
$E=50$ yields the best results overall, considering all tiers.
Our findings align with concurrent work by \citet{yue_pangea_2024}, who found that anywhere between 20 and 80\% of English data yields good global performance. Following these results, we choose $E=50$ as a robust value for the training. 

\subsection{RQ3: Language Distribution in Pre-Training}
\label{sec:analysis:ex3}
As hinted by~\citep{liu_visual_2023,liu_llava-next_2024} and explicitly demonstrated by~\citet{tong_cambrian-1_2024}, pre-training on image-caption pairs improves the LVLM's performance.
We therefore, after identifying an effective distribution of instruction-tuning data, next explore the effect of different distributions of pre-training data across languages.  
%
Specifically, we test if balancing out the English and multilingual portions delivers better performance than unbalanced distributions, that assign more training budget to English or the multilingual mix, respectively. 
%

\rparagraph{Setup}
In these experiments, we fix the instruction-tuning mix to \textit{L100} with $E_{\mathit{IT}}=50\%$ of data in English, which we found in the previous section to produce overall most balanced results. 
%
For the pre-training data mix, we select the same 100 languages, varying the portion of English image-caption pairs, $E_{\mathit{PT}} \in \{100\%, 50\%, 1\%\}$; as in instruction-tuning, the non-English data budget is equally distributed across the other 99 languages.

\begin{table}[t]
    \centering
    \footnotesize
     \def\arraystretch{0.8}
     \resizebox{0.99\linewidth}{!}{
\begin{tabular}{lrrrrrr}
\toprule
\bf English \% & \bf T1 & \bf T2 & \bf T3 & \bf T4 & \bf T5 & \bf en \\
\midrule
No pre-training & 19.3 & 32.6 & 30.7 & 28.9 & 34.4 & 52.6 \\ \midrule
100 & 19.3 & 33.3 & 32.1 & 29.4 & 34.5 &  \textbf{55.2} \\
50 & \textbf{22.8} & \textbf{39.5} & \textbf{33.8} & \underline{30.8} & \textbf{35.7} & 54.9 \\
1 & \underline{22.7} & \underline{38.9} & \underline{33.7} & \textbf{31.2} & \underline{35.4} & \underline{55.1} \\
\bottomrule
\end{tabular}
}
    \caption{\textbf{RQ3} (\S\ref{sec:analysis:ex3}) results with different English-to-multilingual ratios (\textit{L100}) for pre-training. All variants are identically instruction-tuned (\textit{L100, 50\% En.}). 
    }
    \label{tab:ex3}
    \vspace{-0.5em}
\end{table}

\rparagraph{Results} Scores in Table~\ref{tab:ex3} reveal that while English-only pre-training yields downstream benefits on English tasks, it has a largely negligible effect on other languages. 
%
The multilingual mixes 
substantially improve the performance for virtually all language tiers, with gains being the most prominent for lowest-resource languages from T2 and T1. In contrast to instruction-tuning, a very low proportion of clean English data does not result in tangible performance degradation, but it generally does not improve the multilingual performance either. We thus select  $E_{\mathit{PT}}=50\%$ as the ``optimal'' choice for subsequent experiments. 
%
Experiments with Llama 3, with 1\% and 100\% of English data (see Table~\ref{tab:ex3:llama} in the Appendix) support this finding that having a highly multilingual 
pre-training benefits multilingual downstream performance.  

\subsection{RQ4: Improving on Multilingual Text-in-Image Tasks}
\label{sec:analysis:ex4}
Finally, we focus on the models' multilingual understanding of text in images and how to improve it.
Unlike tasks based on natural images, text-in-image tasks cannot be translated trivially from English: even if the prompt and output text are translated, the text in the image remains in English. Because of this, we test how \textit{synthetic} multilingual OCR data, which can be generated at scale in any number of languages, can help improve performance.


\rparagraph{Evaluation} To this end, we introduce SMPQA (Synthetic Multilingual Plot QA) a new multilingual evaluation dataset, which focuses on two fundamental skills required in text-in-image tasks: 1) \textbf{reading} (and outputting) the text from an image and 2) \textbf{grounding} the input text (given as part of the prompt) to the corresponding text in the image (via balanced `yes/no' questions, e.g., ``Is the bar with label \textit{\$Label} the largest?''). We provide further details on the construction and examples in Appendix \ref{sec:appendix:evaluation:smpqa}.\footnote{While MTVQA and M3Exam also require OCR capabilities, they also require input image resolution that is far greater than what we use in our experiments (384px); SMPQA uses bigger letters, making performance effects from multilingual training on text-in-image understanding easier to measure.} 
We construct SMPQA to cover (i) 5 Latin-script languages, one from each tier, and (ii) 6 major languages with different non-Latin scripts. 

\rparagraph{Setup} We generate multilingual synthetic text-in-image data for training following the Synthdog approach \cite{kim_ocr-free_2022} 
(see \ref{sec:appendix:train_data:ocr} for details). We again adopt the training setup \textit{L100} with 50\% English data, both in pre-training and fine-tuning, now adding   
500k Synthdog samples to pre-training and a subset of 50k instances to the instruction-tuning mix.
As before, we select $E\in\{100, 50, 1\}$\% English samples, distributing the rest of the budget equally over the other 99 languages.
We test an additional \textit{Latin-down} distribution: we double the budget allocated to 32 non-Latin-script languages and cut the training budget for Latin-script languages (other than English) in half. 
Importantly, in these experiments we \textit{unfreeze} the image encoder and fine-tune its parameters as well. 

\begin{table}[t]
    \centering
    \footnotesize
     \def\arraystretch{0.9}
     \resizebox{0.99\linewidth}{!}{
\begin{tabular}{lrrrrrr}
\toprule
\bf Setup & \multicolumn{3}{c}{\bf SMPQA Ground} & \multicolumn{3}{c}{\bf SMPQA Read} \\
 & \bf en  & \bf Latin & \bf other & \bf en  & \bf Latin & \bf other \\
\cmidrule(lr){1-1} \cmidrule(lr){2-4} \cmidrule(lr){5-7}
No pre-training & 69.6 & 67.2 & 51.9 & 33.4 & 12.8 & 0.1 \\
No OCR & 76.1 & 73.0 & 55.3 & 41.8 & 23.1 & 0.2 \\
\cmidrule(lr){1-1} \cmidrule(lr){2-4} \cmidrule(lr){5-7}
100\% Eng. & 78.4 & 74.7 & 57.9 & \textbf{55.8} & 39.9 & 3.9 \\
50\% Eng. & \textbf{81.2} & \underline{76.7} & \underline{60.0} & 53.8 & \underline{41.8} & 7.1 \\
50\% (frozen) & 76.1 & 70.8 & 56.3 & 47.2 & 34.1 & 3.5 \\
1\% Eng. & \underline{81.0} & \textbf{78.3} & \textbf{64.1} & \underline{54.8} & \textbf{43.5} & \underline{8.0} \\
Latin down & 78.9 & 74.2 & 59.5 & 54.6 & 41.0 & \textbf{9.9} \\
\bottomrule
\end{tabular}
}
    \caption{\textbf{RQ4} (\S\ref{sec:analysis:ex4}) results of models trained with additional synthetic OCR data on SMPQA for English, Latin-script languages, and languages with other scripts. 
    \textbf{No pre-training}: from Table~\ref{tab:ex2}; \textbf{No OCR}: from Table~\ref{tab:ex3};
    \textbf{frozen}: image encoder frozen;
    \textbf{N\% Eng.}: $N\%$ of OCR data is English, rest uniform distributed over L100 languages;
    \textbf{Latin down}: 2.5k samples for all Latin-script languages, 10k samples for others.
    }
    \label{tab:ex4}
    \vspace{-0.5em}
\end{table}

\rparagraph{Results} Table~\ref{tab:ex4} summarizes the results. The models from prior experiments, \textit{No pre-training} and \textit{No OCR}, succeed for English and other Latin-script languages but utterly fail on non-Latin scripts with near-random performance. We note that the model with the pre-training step (without the additional OCR data) already performs better than the model trained just via instruction-tuning; this is likely due to the presence of images with text coupled with captions that explicitly mention this text.
Training with synthetic data greatly improves the performance across all languages even if all of the OCR data is in English (100\% Eng.).
Nonetheless, using multilingual synthetic OCR data is very effective and, importantly, does not degrade English SMPQA performance even if English constitutes only 1\% of the training data. We note that \textit{unfreezing} and training the image encoder is critical for optimal performance in \textit{all scripts}.
Despite all this, we still observe a large performance gap between Latin- and non-Latin-script languages, even if we skew the training budget towards the non-Latin scripts (\textit{Latin-down}).  
We hypothesize that orders of magnitude more text-in-image training data for other scripts are required for adequate performance.\footnote{Concurrently, in a preliminary exploration of text-in-image capabilities, \citet{yue_pangea_2024} noted steady gains with 50k samples per language but also observed worse performance for non-Latin-script languages.}
%


\section{\modelname: Applying Lessons Learned}
\label{sec:centurio}

\begin{table*}[t]
    \centering
    \footnotesize
 \def\arraystretch{0.85}
 \resizebox{0.99\linewidth}{!}{
\begin{tabular}{lrrrrrrrrrrrrrrrrrrrrrrrrrrr}
\toprule
 & \bf AVG. & \multicolumn{3}{c}{\bf XM3600} & \multicolumn{1}{c}{\bf MT-} & \multicolumn{2}{c}{\bf SMPQA G.} & \multicolumn{2}{c}{\bf SMPQA N.} & \multicolumn{2}{c}{\bf M3Exam} & \multicolumn{2}{c}{\bf xMMMU} & \multicolumn{1}{c}{\bf C-} \\
& & \bf en & \bf mul & \bf fid. & \bf VQA & \bf en & \bf mul & \bf en & \bf mul & \bf en & \bf mul & \bf en & \bf mul & \bf VQA \\
\cmidrule(lr){1-1}  \cmidrule(lr){2-2} \cmidrule(lr){3-5} \cmidrule(lr){6-6} \cmidrule(lr){7-8} \cmidrule(lr){9-10} \cmidrule(lr){11-12} \cmidrule(lr){13-14} \cmidrule(lr){15-15}
Parrot & 25.8 & 5.6 & 0.4 & 25.0 & 2.0 & 51.0 & 49.9 & 0.0 & 0.0 & 46.6 & 36.2 & 35.3 & 32.4 & 41.1 \\
PALO 7B & 28.7 & 65.9 & 13.5 & 72.0 & 5.8 & 55.5 & 52.8 & 22.4 & 2.7 & 41.0 & 29.1 & 31.8 & 30.9 & 37.1 \\
PALO 13B & 29.9 & 67.3 & 17.0 & 60.1 & 6.3 & 54.0 & 51.5 & 25.6 & 4.0 & 45.2 & 28.3 & 32.4 & 28.9 & 39.6 \\
Llama-Vision 3.2 11B & *32.3 & 35.9 & 7.2 & 33.3 & 15.2 & 91.1 & \underline{84.8} & 58.4 & 22.8 & --- & --- & --- & --- & 38.8 \\
Maya & 33.4 & 55.9 & 14.6 & 65.7 & 5.3 & 51.4 & 50.9 & 14.6 & 1.8 & 49.2 & 36.3 & 37.9 & 33.3 & 39.8 \\
Pixtral 12B & 38.1 & 26.5 & 22.1 & \textbf{96.8} & 14.1 & 91.1 & 71.0 & \underline{85.0} & 35.9 & 49.4 & 33.7 & 30.3 & 26.2 & 33.5 \\
Phi 3.5 Vision & 39.5 & 32.3 & 6.3 & 40.8 & 11.1 & \textbf{92.2} & 79.4 & 84.8 & 35.9 & 56.3 & 40.7 & 41.7 & 37.4 & 40.9 \\
Qwen2VL 2B & 41.2 & 68.8 & 5.2 & 13.2 & 19.0 & 85.0 & 83.5 & 68.8 & 47.4 & 47.9 & 40.5 & 36.8 & 35.5 & 33.6 \\
MiniCPM 2.6 & 41.7 & \textbf{87.5} & 14.2 & 92.3 & 16.1 & 89.0 & 74.3 & 80.8 & 39.3 & 55.0 & 48.2 & 39.1 & 36.5 & 34.1 \\
InternVL 2.5 4B & 45.3 & 38.9 & 17.5 & 91.0 & \textbf{25.1} & 87.0 & 78.3 & 77.8 & 47.5 & \underline{63.2} & \underline{50.3} & \underline{49.2} & 42.7 & 48.1 \\
InternVL 2.5 8B & 47.4 & 38.3 & 15.7 & 91.1 & \underline{25.0} & 91.0 & 79.2 & 80.6 & \underline{48.2} & \textbf{67.0} & \textbf{53.3} & \textbf{50.7} & \textbf{45.2} & 48.6 \\
Qwen2VL 7B & 47.7 & 50.3 & 24.6 & 90.0 & 23.2 & \underline{91.2} & \textbf{90.9} & \underline{85.0} & \textbf{64.9} & 56.1 & 49.7 & 43.0 & 40.7 & 37.6 \\
Pangea & 48.2 & 70.1 & \underline{34.6} & 87.9 & 19.3 & 87.2 & 72.2 & 72.0 & 23.8 & 58.0 & 45.5 & 43.1 & 42.0 & \textbf{55.2} \\
\cmidrule(lr){1-1}
\modelname \texttt{Aya} & \underline{48.5} & 78.4 & \textbf{39.2} & \underline{95.7} & 11.1 & 83.1 & 74.2 & 60.0 & 30.1 & 53.0 & 41.2 & 37.6 & 37.2 & 49.4 \\
\modelname \texttt{Qwen} & \textbf{51.6} & \underline{79.1} & 34.4 & 95.2 & 11.9 & 84.8 & 76.1 & 65.2 & 31.7 & 61.2 & 46.9 & 46.4 & \underline{43.0} & \underline{52.9} \\
\midrule
 & \multicolumn{2}{c}{\bf MAXM} & \multicolumn{2}{c}{\bf xGQA} & \multicolumn{2}{c}{\bf BIN-MC} & \multicolumn{2}{c}{\bf XVNLI} & \multicolumn{2}{c}{\bf MaRVL} & \multicolumn{2}{c}{\bf VGR} & \multicolumn{2}{c}{\bf VLOD} \\
 & \bf en & \bf mul & \bf en & \bf mul& \bf en & \bf mul& \bf en & \bf mul& \bf en & \bf mul& \bf en & \bf mul& \bf en & \bf mul \\
\cmidrule(lr){1-1} \cmidrule(lr){2-3} \cmidrule(lr){4-5} \cmidrule(lr){6-7} \cmidrule(lr){8-9} \cmidrule(lr){10-11} \cmidrule(lr){12-13} \cmidrule(lr){14-15}
Parrot & 28.2 & 3.6 & 37.7 & 21.2 & 30.5 & 25.7 & 28.7 & 31.4 & 63.5 & 55.1 & 59.2 & 52.9 & 0.0 & 0.0 \\
PALO 7B & 54.0 & 22.5 & 59.1 & 36.6 & 58.7 & 38.6 & 58.0 & 53.4 & 62.7 & 24.1 & 48.3 & 25.6 & 5.8 & 6.8 \\
PALO 13B & 51.7 & 33.1 & 58.0 & 27.8 & 61.4 & 41.1 & 56.6 & 53.6 & 63.8 & 33.1 & 63.3 & 26.2 & 2.5 & 4.9 \\
Llama-Vision 3.2 11B & 0.0 & 4.7 & 39.3 & 27.6 & 75.6 & 50.8 & --- & --- & --- & --- & --- & --- & --- & --- \\
Maya & 55.4 & 17.3 & 58.2 & 49.1 & 54.0 & 43.2 & 50.1 & 43.9 & 60.3 & 56.3 & 46.7 & 42.3 & 20.0 & 20.1 \\
Pixtral 12B & 59.4 & 43.4 & 59.9 & 3.8 & 71.0 & 54.2 & 60.9 & 52.7 & 67.7 & 60.7 & 55.8 & 47.7 & 9.2 & 12.4 \\
Phi 3.5 Vision & 43.6 & 17.9 & \textbf{65.2} & 38.0 & 63.1 & 36.8 & 58.9 & 53.3 & 73.4 & 46.4 & 81.7 & 50.3 & \underline{45.8} & \textbf{31.5} \\
Qwen2VL 2B & 53.7 & 26.5 & 60.5 & 38.2 & \underline{78.2} & 47.2 & 61.9 & 56.2 & 67.9 & 55.9 & 61.7 & 50.5 & 22.5 & 20.4 \\
MiniCPM 2.6 & 53.4 & 22.3 & 57.9 & 45.7 & 72.6 & 47.4 & 71.9 & 65.4 & 70.2 & 57.9 & 52.5 & 49.1 & 9.2 & 14.6 \\
InternVL 2.5 4B & 46.0 & 42.5 & 63.6 & 28.0 & 68.4 & 45.4 & 69.0 & 58.7 & 74.9 & 59.0 & 72.5 & 49.7 & 24.2 & 21.0 \\
InternVL 2.5 8B & 45.6 & 38.2 & 63.4 & 32.0 & 70.3 & 44.2 & \underline{73.5} & \underline{66.4} & 83.0 & 63.3 & \underline{87.5} & 51.6 & \textbf{57.5} & \underline{29.0} \\
Qwen2VL 7B & 54.7 & 31.2 & 62.5 & 49.3 & \textbf{80.7} & \textbf{57.5} & 62.1 & 59.6 & 69.8 & 60.2 & 60.0 & 52.9 & 5.8 & 13.2 \\
Pangea & \textbf{61.4} & \textbf{55.0} & \underline{64.6} & \textbf{60.4} & 70.3 & 52.1 & 69.0 & 65.2 & 75.8 & 70.5 & 69.2 & 58.9 & 0.0 & 6.7 \\
\cmidrule(lr){1-1}
\modelname \texttt{Aya} & 55.7 & \underline{49.3} & 59.1 & 53.2 & 69.7 & 54.7 & 65.0 & 62.4 & \underline{85.0} & \underline{77.9} & 82.5 & \underline{66.8} & 12.5 & 20.7 \\
\modelname \texttt{Qwen} & \underline{60.1} & 47.7 & 60.6 & \underline{54.8} & 72.7 & \underline{56.2} & \textbf{75.4} & \textbf{70.2} & \textbf{89.6} & \textbf{81.7} & \underline{87.5} & \textbf{73.1} & 28.3 & 27.0 \\
\bottomrule
\end{tabular}
}
    \caption{Comparison of \modelname and 13 other LVLMs across 14 tasks.
    We highlight the \textbf{best} and \underline{second-best} results.
    Scores are accuracy (CIDEr for XM3600).
    \textbf{en} \& \textbf{mul} are the English and averaged multilingual results.
    \textbf{XM3600 fid.} is the language fidelity over all languages;
    \textbf{SMPQA G. \& N} are \textbf{G}rounding and \textbf{N}aming.
    \textbf{*:} supports only single-image input.
    \textbf{AVG.}: average over all tasks.
    Details on the setup and models are provided in Appendix~\ref{sec:appendix:evaluation}.
    }
    \label{tab:baselines:per_task}
    
    \vspace{-0.5em}
\end{table*}

\begin{table}[t]
    \centering
    \footnotesize
 \def\arraystretch{0.9}
 \resizebox{0.99\linewidth}{!}{
\begin{tabular}{lrrrrrr}
\toprule
\bf Model & \bf T1 & \bf T2 & \bf T3 & \bf T4 & \bf T5 & \bf en \\
\midrule
\modelname \texttt{Aya} & 35.1 & \underline{46.4} & \underline{47.0} & \underline{46.7} & 48.3 & 60.6 \\
\modelname \texttt{Qwen} & \underline{38.1} & \textbf{51.0} & \textbf{48.3} & \textbf{47.0} & \textbf{50.9} & \textbf{66.6} \\
\cmidrule(lr){1-1}
InternVL 2.5 8B & 29.9 & 37.0 & 37.4 & 41.0 & \underline{50.5} & \underline{64.4} \\
Qwen2VL 7B & 30.6 & 36.8 & 40.5 & 46.2 & 48.0 & 56.8 \\
Pangea & \textbf{38.5} & 38.6 & 46.9 & 44.2 & 49.9 & 59.8 \\
\midrule
\multicolumn{6}{l}{\bf Without multi-image tasks (MaRVL, VGR, VLOD):} \\
\midrule
\modelname \texttt{Aya} & 35.1 & 44.5 & \underline{45.7} & \underline{46.2} & 47.7 & 60.7 \\
\modelname \texttt{Qwen} & \underline{38.1} & \textbf{49.5} & 45.6 & 45.8 & 49.6 & \textbf{66.0} \\
\cmidrule(lr){1-1}
InternVL 2.5 8B & 29.9 & 40.4 & 35.2 & 39.4 & \underline{49.7} & 62.3 \\
Qwen2VL 7B & 30.6 & 38.7 & 40.8 & \textbf{46.8} & 48.3 & 61.7 \\
Pangea & \textbf{38.5} & \underline{46.5} & \textbf{47.7} & 44.4 & \textbf{49.9} & \underline{64.9} \\
\bottomrule
\end{tabular}
}
    \caption{Comparison between \modelname and the top-3 models of Table~\ref{tab:baselines:per_task}. Scores are averages over results from all 14 tasks grouped by language tier. 
    }
    \label{tab:baselines:per_langugage}
\end{table}

Our answers to \textbf{RQ1}--\textbf{RQ4} (see  \S\ref{sec:analysis}) point to the feasibility of training massively multilingual LVLMs supporting 100 languages with a ``sweet spot'' of roughly $50\%$ of the English data being MT-translated to the languages covered.
For improving multilingual OCR capabilities, training on large-scale synthetic data with an unfrozen image encoder has proven effective.
Demonstrating the practicability of our findings, we now train state-of-the-art multilingual LVLMs applying our lessons learned, which we call \modelname. We briefly describe further design choices below.

\subsection{Design Choices}
\rparagraph{Text Encoder}
The choice of the LLM greatly impacts multilingual performance.
We benchmark several LLMs (with 7-9B parameters) following the evaluation setup described in \S\ref{sec:analysis} for L100 languages and translations for $50\%$ of the English instruct data to find candidates for \modelname (details in Appendix \ref{sec:appendix:experiments:llm}).
The best performances where obtained with Aya-Expanse \cite{dang2024ayaexpansecombiningresearch} and Qwen 2.5 \cite{yang_qwen2_2024} as backbones.

\rparagraph{Image Tiling and Projection}
Image tiling methods \cite{lin_sphinx_2024,liu_llava-next_2024} increase the image resolution by concatenating encodings of $n$ non-overlapping tiles of an input image together, which significantly helps with `reading' small text in images.
However, they also greatly increase the input length: a $2\times2$ tiling would yield 3,645 tokens per image with our model.\footnote{The whole image plus four tiles, each with 729 tokens.}
Instead, we adopt the method by \citet{shi_when_2024}, which concatenates the tokens of the whole image and the tiles along the \textit{feature dimension} before projection by the MLP. 
This gives an efficient trade-off between computing cost---the number of tokens stays constant---and performance gains for fine-grained content.

\rparagraph{Training Data}
We increase the amount of the pre-training and instruct tuning data to further improve performance beyond our analysis setup.
For pre-training, we add the 0.7M ALLaVA  captions~\cite{chen_allava_2024} to the ShareGPT-4V captions and we use all synthetic OCR data generated in  \S\ref{sec:analysis:ex4} (1.16M total: 500k English, 5k for Latin-script language, 10k for other scripts).
For instruction-tuning, we incorporate additional datasets from the Cambrian collection~\cite{tong_cambrian-1_2024} along with several text-only instruction-tuning datasets (see Appendix~\ref{sec:appendix:train_data:scaled} for a list).
We translate the data to the \textit{L100 50\% En.} setup, excluding text-heavy datasets and others that are problematic for MT.

\subsection{Results}
We compare our \modelname models against 13 other multilingual LVLMs 
across the 13 tasks used in \S\ref{sec:analysis}, and additionally evaluate them on CVQA\footnote{CVQA has a private test set and only allows limited submissions hence we left it out for our analysis experiments.}, testing the models' capabilities across 56 languages.
We provide details for all models in Appendix~\ref{sec:appendix:evaluation:baselines}.

On average, \modelname achieves the best results across the 14 tasks on their multilingual portions and additionally performs strongly on English (Table~\ref{tab:baselines:per_task}). These results prove the effectiveness of our training composition: we are able to retain high English performances while maximizing the models' multilingual capabilities.
When analyzing these results grouped by language tier (Table~\ref{tab:baselines:per_langugage}), we find that our models shine in the low-resource tiers T1\footnote{Despite 4/7 T1 CVQA languages \textbf{not} in our training data.} and T2, with competitive results for higher-resource languages---even when excluding multi-image tasks (VGR, MaRVL, VLOD), where our models greatly outperform most others.

Only for text-heavy tasks (primarily MTVQA and SMPQA), \modelname falls behind. 
While we show the importance of multilingual OCR training---\modelname succeeds at the SMPQA reading task in more languages than, for example, Pangea
---the limited input resolution and magnitudes less OCR data compared to Qwen2-VL and others 
result in comparably poor performance.

\section{Related Work}

\sparagraph{Multilingual LVLMs}
Building on the success of monolingual LVLMs like BLIP-2~\cite{li_blip-2_2023} and LLaVA~\cite{liu_visual_2023,liu_improved_2023}, researchers extended the English training protocols to include multilingual data for obtaining massively multilingual LVLMs~\citep[e.g.,][]{maaz_palo_2024, geigle_mblip_2023-1}. 
As such, Google's PaLI models~\cite{chen_pali_2022,chen_pali-x_2023} were the first closed-weight models trained on multilingual captions and VQA data with the recent open-weight PaliGemma \cite{beyer_paligemma_2024} following a similar training strategy.
\citet{geigle_mblip_2023-1} presented with mBLIP the first open model, trained with image captions and a limited mix of instruct data translated to 98 languages.
Subsequent models similarly followed an established procedure by directly translating parts of the English training data~\cite{maaz_palo_2024,hu_large_2024,alam2024maya}.
 For the concurrent Pangea, \citet{yue_pangea_2024} optimized for multicultural aspects and used a mix of machine-translated data, existing multilingual data, and synthetically generated data. 
 While they analyze the ratio between English and multilingual data, they do not vary the number of languages, fixing it at 39.
 Interestingly, most researchers either (\emph{i})~did not 
properly motivate their 
multilingual data mix~\citep[e.g.,][]{geigle_mblip_2023-1,alam2024maya,beyer_paligemma_2024}, or (\emph{ii}) did not provide any details on the training data composition~\citep[e.g.,][]{wang_qwen2-vl_2024,yao_minicpm-v_2024,chen2024expandingperformanceboundariesopensource})

\rparagraph{Multilingual OCR with LVLMs} While OCR recently gained popularity for English LVLMs \cite{lu_deepseek-vl_2024, tong_cambrian-1_2024}, multilingual OCR has largely been neglected in prior work. As an exception, Qwen2-VL~\citep{wang_qwen2-vl_2024} and InternVL 2.5~\citep{chen2024expandingperformanceboundariesopensource} exhibit excellent multilingual OCR capabilities, but no training details are known. Towards open knowledge on improving multilingual OCR, \citet{yue_pangea_2024} performed preliminary experiments leveraging data in 10 languages. 
However, such efforts are still hindered by the lack of evaluation resources: MTVQA \cite{tang_mtvqa_2024} and M3Exam \cite{zhang_m3exam_2023} only cover up to 9 languages and  conflate language understanding (in the text input) with understanding text on images. 
In this work, we push multilingual OCR research by presenting the novel SMPQA dataset dedicated to evaluation of multilingual OCR. We further explore how synthetic training data can improve models' capabilities.

\rparagraph{Multilingual Instruction Tuning of LLMs.}
 %
While older LLMs struggled in multilingual tasks ~\citep{ahuja-etal-2024-megaverse},
more recent ones like Qwen 2.5 \cite{yang_qwen2_2024}, Llama 3 \cite{llama_team_llama_2024}, Gemma 2 \cite{gemma_team_gemma_2024}, or Aya \cite{aryabumi_aya_2024} have greatly improved in that respect, making them usable in many languages besides English.
Still, current LLMs often fail to respond faithfully to the prompting language if that language is not English, especially for low-resource languages~\cite{holtermann-etal-2024-evaluating, kew-etal-2024-turning,marchisio_understanding_2024}.
To mitigate this issue, several works have analyzed the importance of multilingual instruction tuning.
\citet{weber-etal-2024-investigating} demonstrated that multilingual training is crucial for downstream performance even if the base models are pre-trained on multilingual data mixtures.
Others showed that just a small set of languages
is sufficient to improve cross-lingual transfer 
for multilingual downstream tasks significantly~\citep{shaham-etal-2024-multilingual,chen-etal-2024-monolingual,kew-etal-2024-turning}.
However, they focus on a small set of primarily higher-resource languages,
while we consider the problem in the vision-language context for a wider language selection.
%
%

In \cite{soykan2024linguistically}, the authors propose methods to select the optimal mix of languages for instruction tuning in a ''linguistically-informed manner''.
%
However, they find no general best selection, and instead a task- and model-dependent selection is necessary.
%
Therefore, in our work, we do not apply these techniques
and instead choose languages based on the taxonomy introduced by \citet{joshi_state_2020}.

\section{Conclusion}
In this study, we systematically investigated the optimal data composition for training a multilingual LVLM through four progressively refined analysis setups.
Our findings reveal that massively multilingual training with 100 languages is highly effective, achieving comparable results to configurations with fewer languages.
Moreover, only 25–50\% of the training data needs to be non-English, keeping the cost of multilingual data production low.
To enhance multilingual text understanding in images, we introduced a novel evaluation benchmark and demonstrated the importance and effectiveness of including multilingual synthetic OCR data in the training mix.
Finally, we apply our findings to train \modelname, massively multilingual LVLMs trained with 100 languages, and achieve state-of-the-art results on our evaluation suite covering 14 tasks and 56 language tasks against 13 other LVLMs.

\newpage
\section{Limitations}
\paragraph{Lack of Explicit Multicultural Training}
The focus of this work is on \textit{language understanding} in a massively multilingual setup, that is, how to train the model to maximize its ability to understand and generate text in various languages.
We do not consider the multicultural aspect, that is, training a model so that it is also more knowledgeable about concepts from the countries whose languages it can understand as measured by benchmarks like CVQA or CulturalVQA \cite{nayak_benchmarking_2024}.
While the two aspects --- multilingual and multicultural knowledge --- can be intermingled in practice, they require distinct approaches in training:
Multilingual data is necessary for multilingual language understanding, as we have shown.
However, multicultural knowledge can be learned from multilingual resources as created by, for example, by \citet{yue_pangea_2024}, but also from fully English resources like Wikipedia \cite{srinivasan_wit_2021}.

\paragraph{Using Machine-Translated Training Data}
We train our model using machine-translated (MT) data derived from high-quality English datasets.
This is advantageous because it allows us to create comparable setups for our analyses with full control over the languages and their proportions.
While the data proves effective in increasing multilingual performance, MT data, especially for low-resource languages, can be of low quality and, even in higher-resource languages, might exhibit unwanted ``translationese'' artifacts.
This can negatively impact the quality of generated text in a way that the metrics employed in our evaluation suite do not adequately measure.
While native multilingual training data is available, it is not available for all tasks or languages equally, or, for most languages, not at all.
Future work should consider evaluation setups to quantify the effect the MT data has on the final model, work on better MT pipelines, or create more data through native speakers.

\paragraph{Using Synthetically Generated OCR Data}
The text-heavy, ``real-world'' tasks in some datasets of our instruction tuning mix, which cover diverse image types such as plots, scans, application screenshots, or screenshots of webpages, are still entirely in English.
Due to the issues that arise when translating such samples, we do not translate them.
Hence, our methods to improve the understanding of multilingual texts in images are limited to only using synthetically generated images.
While we have seen that our synthetic data positively impacts the performance of models on the respective tasks, future work should explore methods for collecting or generating more diverse data in different languages beyond our synthetic OCR data.

Another limitation regarding OCR capabilities is our relatively small image input resolution compared to models like Qwen2-VL or InternVL 2.5 --- both of which support image inputs in native resolution at the cost of thousands of tokens per image ---, which limits the performance of \modelname for images with small text.

\section*{Acknowledgments}
Simulations were performed with computing resources
granted by WestAI under project 9148.

Simulations were performed with computing resources from Julia 2.
Julia 2 was funded as DFG project as “Forschungsgroßgerät nach Art 91b GG” under INST 93/1145-1 FUGG

To work of Gregor Geigle was in part supported by the Alexander von Humboldt Foundation.

The work of Carolin Holtermann and Anne Lauscher is funded under the Excellence Strategy of the German Federal Government and the States.

\bibliography{custom,flo,anthology}

\appendix

\section{Training Setup}
\label{sec:appendix:training}

All models are trained with the following hyperparameters:
AdamW optimizer \cite{loshchilov_decoupled_2019} with cosine learning rate schedule and $3\%$ linear warmup;
LORA \cite{hu_lora_2022-1} is used with rank $256$ and $\alpha 512$ and applied to all matrices in the LLM -- the LLM is otherwise frozen;
the image encoder is frozen in the first three experiments and jointly trained with the model otherwise;
weight decay is 0;
batch size is $32$ using gradient accumulation;
learning rate is $1e-6$ for the image encoder, $1e-4$ for LORA and the MLP in general except when training \modelname, we use $5e-5$ during pretraining and $3e-5$ during instruct tuning.
Models are always trained for one epoch on the entire data.
The training loss is causal language modeling and we mask both image and input prompt tokens for calculating the loss.

For going from the pretraining to the instruct tuning phase, we found it best to continue training the same LORA adapter; merging the LORA weights after pretraining and initializing a new adapter gave worse results.

Hyperparameter (LORA rank \& $\alpha$, learning rates, weight decay) were tuned for Phi 3.5 and transferred to the other LLMs.

All models were trained with $4$ H100 GPUs.
Training \modelname took $\approx6$ days (half for pre-training, half for instruct tuning).

Training of one Phi 3.5 model for \S\ref{sec:analysis} takes 8-10h for instruct tuning, and for pre-training 12 to 20h (with synthetic OCR data and unfrozen image encoder).

\subsection{Training Languages}
\label{sec:appendix:train_langs}
We list the 100 languages used in training in Table~\ref{tab:training_data:languages}.
\begin{table*}[htbp]
    \centering
    \begin{minipage}[t]{0.45\textwidth}
        \scriptsize
        \centering
            \begin{tabular}{llllr}
            \toprule
            Name & Script & ISO-639 & Flores-200 & Tier \\
            \midrule
            Arabic & Arabic & \texttt{ar} & \texttt{arb\_Arab} & $5$ \\
            Chinese & Trad. Han & \texttt{zh} & \texttt{zho\_Hant} & $5$ \\
            English & Latin & \texttt{en} & \texttt{eng\_Latn} & $5$ \\
            French & Latin & \texttt{fr} & \texttt{fra\_Latn} & $5$ \\
            German & Latin & \texttt{de} & \texttt{deu\_Latn} & $5$ \\
            Japanese & Japanese & \texttt{ja} & \texttt{jpn\_Jpan} & $5$ \\
            Spanish & Latin & \texttt{es} & \texttt{spa\_Latn} & $5$ \\
            Basque & Latin & \texttt{eu} & \texttt{eus\_Latn} & $4$ \\
            Catalan & Latin & \texttt{ca} & \texttt{cat\_Latn} & $4$ \\
            Croatian & Latin & \texttt{hr} & \texttt{hrv\_Latn} & $4$ \\
            Czech & Latin & \texttt{cs} & \texttt{ces\_Latn} & $4$ \\
            Dutch & Latin & \texttt{nl} & \texttt{nld\_Latn} & $4$ \\
            Finnish & Latin & \texttt{fi} & \texttt{fin\_Latn} & $4$ \\
            Hindi & Devanagari & \texttt{hi} & \texttt{hin\_Deva} & $4$ \\
            Hungarian & Latin & \texttt{hu} & \texttt{hun\_Latn} & $4$ \\
            Italian & Latin & \texttt{it} & \texttt{ita\_Latn} & $4$ \\
            Korean & Hangul & \texttt{ko} & \texttt{kor\_Hang} & $4$ \\
            Persian & Arabic & \texttt{fa} & \texttt{pes\_Arab} & $4$ \\
            Polish & Latin & \texttt{pl} & \texttt{pol\_Latn} & $4$ \\
            Portuguese & Latin & \texttt{pt} & \texttt{por\_Latn} & $4$ \\
            Russian & Cyrillic & \texttt{ru} & \texttt{rus\_Cyrl} & $4$ \\
            Serbian & Cyrillic & \texttt{sr} & \texttt{srp\_Cyrl} & $4$ \\
            Swedish & Latin & \texttt{sv} & \texttt{swe\_Latn} & $4$ \\
            Turkish & Latin & \texttt{tr} & \texttt{tur\_Latn} & $4$ \\
            Vietnamese & Latin & \texttt{vi} & \texttt{vie\_Latn} & $4$ \\
            Afrikaans & Latin & \texttt{af} & \texttt{afr\_Latn} & $3$ \\
            Bangla & Bengali & \texttt{bn} & \texttt{ben\_Beng} & $3$ \\
            Belarusian & Cyrillic & \texttt{be} & \texttt{bel\_Cyrl} & $3$ \\
            Bosnian & Latin & \texttt{bs} & \texttt{bos\_Latn} & $3$ \\
            Bulgarian & Cyrillic & \texttt{bg} & \texttt{bul\_Cyrl} & $3$ \\
            Cebuano & Latin & \texttt{ceb} & \texttt{ceb\_Latn} & $3$ \\
            Danish & Latin & \texttt{da} & \texttt{dan\_Latn} & $3$ \\
            Egyptian Arabic & Arabic & \texttt{ar-eg} & \texttt{arz\_Arab} & $3$ \\
            Estonian & Latin & \texttt{et} & \texttt{est\_Latn} & $3$ \\
            Galician & Latin & \texttt{gl} & \texttt{glg\_Latn} & $3$ \\
            Georgian & Georgian & \texttt{ka} & \texttt{kat\_Geor} & $3$ \\
            Greek & Greek & \texttt{el} & \texttt{ell\_Grek} & $3$ \\
            Indonesian & Latin & \texttt{id} & \texttt{ind\_Latn} & $3$ \\
            Kazakh & Cyrillic & \texttt{kk} & \texttt{kaz\_Cyrl} & $3$ \\
            Latin & Latin & \texttt{la} & \texttt{NO} & $3$ \\
            Latvian & Latin & \texttt{lv} & \texttt{lvs\_Latn} & $3$ \\
            Lithuanian & Latin & \texttt{lt} & \texttt{lit\_Latn} & $3$ \\
            Malay & Latin & \texttt{ms} & \texttt{zsm\_Latn} & $3$ \\
            Romanian & Latin & \texttt{ro} & \texttt{ron\_Latn} & $3$ \\
            Slovak & Latin & \texttt{sk} & \texttt{slk\_Latn} & $3$ \\
            Slovenian & Latin & \texttt{sl} & \texttt{slv\_Latn} & $3$ \\
            Tagalog & Latin & \texttt{tl} & \texttt{tgl\_Latn} & $3$ \\
            Tamil & Tamil & \texttt{ta} & \texttt{tam\_Taml} & $3$ \\
            Thai & Thai & \texttt{th} & \texttt{tha\_Thai} & $3$ \\
            Ukrainian & Cyrillic & \texttt{uk} & \texttt{ukr\_Cyrl} & $3$ \\
            \bottomrule
        \end{tabular}
    \end{minipage}
    \hspace{0.2cm}
    \begin{minipage}[t]{0.45\textwidth}
        \scriptsize
        \centering
        \begin{tabular}{llllr}
            \toprule
            Name & Script & ISO-639 & Flores-200 & Tier \\
            \midrule
        Urdu & Arabic & \texttt{ur} & \texttt{urd\_Arab} & $3$ \\
        Uzbek & Latin & \texttt{uz} & \texttt{uzn\_Latn} & $3$ \\
        Hebrew & Hebrew & \texttt{iw}\texttt{he} & \texttt{heb\_Hebr} & $3$ \\
        Amharic & Ethiopic & \texttt{am} & \texttt{amh\_Ethi} & $2$ \\
        Haitian & Latin & \texttt{ht} & \texttt{hat\_Latn} & $2$ \\
        Hausa & Latin & \texttt{ha} & \texttt{hau\_Latn} & $2$ \\
        Icelandic & Latin & \texttt{is} & \texttt{isl\_Latn} & $2$ \\
        Irish & Latin & \texttt{ga} & \texttt{gle\_Latn} & $2$ \\
        Lao & Lao & \texttt{lo} & \texttt{lao\_Laoo} & $2$ \\
        Maltese & Latin & \texttt{mt} & \texttt{mlt\_Latn} & $2$ \\
        Marathi & Devanagari & \texttt{mr} & \texttt{mar\_Deva} & $2$ \\
        Punjabi & Gurmukhi & \texttt{pa} & \texttt{pan\_Guru} & $2$ \\
        Sanskrit & Devanagari & \texttt{sa} & \texttt{san\_Deva} & $2$ \\
        Swahili & Latin & \texttt{sw} & \texttt{swh\_Latn} & $2$ \\
        Tigrinya & Ethiopic & \texttt{ti} & \texttt{tir\_Ethi} & $2$ \\
        Tswana & Latin & \texttt{tn} & \texttt{tsn\_Latn} & $2$ \\
        Wolof & Latin & \texttt{wo} & \texttt{wol\_Latn} & $2$ \\
        Xhosa & Latin & \texttt{xh} & \texttt{xho\_Latn} & $2$ \\
        Yoruba & Latin & \texttt{yo} & \texttt{yor\_Latn} & $2$ \\
        Zulu & Latin & \texttt{zu} & \texttt{zul\_Latn} & $2$ \\
        Albanian & Latin & \texttt{sq} & \texttt{als\_Latn} & $1$ \\
        Assamese & Bengali & \texttt{as} & \texttt{asm\_Beng} & $1$ \\
        Azerbaijani & Arabic & \texttt{azb} & \texttt{azb\_Arab} & $1$ \\
        Bambara & Latin & \texttt{bm} & \texttt{bam\_Latn} & $1$ \\
        Burmese & Myanmar & \texttt{my} & \texttt{mya\_Mymr} & $1$ \\
        Esperanto & Latin & \texttt{eo} & \texttt{epo\_Latn} & $1$ \\
        Igbo & Latin & \texttt{ig} & \texttt{ibo\_Latn} & $1$ \\
        Javanese & Latin & \texttt{jv} & \texttt{jav\_Latn} & $1$ \\
        Khmer & Khmer & \texttt{km} & \texttt{khm\_Khmr} & $1$ \\
        Kikuyu & Latin & \texttt{ki} & \texttt{kik\_Latn} & $1$ \\
        Lingala & Latin & \texttt{ln} & \texttt{lin\_Latn} & $1$ \\
        Luxembourgish & Latin & \texttt{lb} & \texttt{ltz\_Latn} & $1$ \\
        Maori & Latin & \texttt{mi} & \texttt{mri\_Latn} & $1$ \\
        Norwegian & Latin & \texttt{no} & \texttt{nob\_Latn} & $1$ \\
        Occitan & Latin & \texttt{oc} & \texttt{oci\_Latn} & $1$ \\
        Quechua & Latin & \texttt{qu} & \texttt{quy\_Latn} & $1$ \\
        Samoan & Latin & \texttt{sm} & \texttt{smo\_Latn} & $1$ \\
        Sango & Latin & \texttt{sg} & \texttt{sag\_Latn} & $1$ \\
        Sardinian & Latin & \texttt{sc} & \texttt{srd\_Latn} & $1$ \\
        Scottish Gaelic & Latin & \texttt{gd} & \texttt{gla\_Latn} & $1$ \\
        Sindhi & Arabic & \texttt{sd} & \texttt{snd\_Arab} & $1$ \\
        Somali & Latin & \texttt{so} & \texttt{som\_Latn} & $1$ \\
        Swati & Latin & \texttt{ss} & \texttt{ssw\_Latn} & $1$ \\
        Telugu & Telugu & \texttt{te} & \texttt{tel\_Telu} & $1$ \\
        Tibetan & Tibetan & \texttt{bo} & \texttt{bod\_Tibt} & $1$ \\
        Tok Pisin & Latin & \texttt{tpi} & \texttt{tpi\_Latn} & $1$ \\
        Tsonga & Latin & \texttt{ts} & \texttt{tso\_Latn} & $1$ \\
        Twi & Latin & \texttt{tw} & \texttt{twi\_Latn} & $1$ \\
        Waray & Latin & \texttt{war} & \texttt{war\_Latn} & $1$ \\
        Welsh & Latin & \texttt{cy} & \texttt{cym\_Latn} & $1$ \\
        \bottomrule
        \end{tabular}
    \end{minipage}
    \caption{The list of 100 languages used in our training experiments. The ''Tier'' column represents the tier in the taxonomy proposed by \citet{joshi_state_2020}, where a higher tier indicates more available resources, i.e., data, in the respective language.}
    \label{tab:training_data:languages}
\end{table*}

\section{Training Data}
\label{sec:appendix:train_data}

\subsection{For Analysis Experiments}

\begin{table}[t]
    \centering
     \def\arraystretch{0.97}
     \resizebox{0.99\linewidth}{!}{
    \begin{tabular}{lrrl}
\toprule
     Dataset  &  Size (Images) & Translated?\\
 \midrule
\multicolumn{4}{l}{\bf Natural Image:}\\
LLaVA Instruct \cite{liu_visual_2023}	& 160k & yes & \\
VQAv2 \cite{goyal_making_2017} &	83k & yes & \\
GQA \cite{hudson_gqa_2019} &	72k & yes & \\
OKVQA \cite{marino_ok-vqa_2019} &	9k & yes & \\
A-OKVQA	\cite{schwenk_-okvqa_2022-1} & 30k & yes & \\
RefCOCO \cite{kazemzadeh_referitgame_2014,mao_generation_2016}	& 48k & yes & \\
VG	\cite{krishna_visual_2017} & 86k & yes & \\
MSCOCO \cite{lin_microsoft_2014}	& 50k (subset) & yes & \\
 \midrule
\multicolumn{4}{l}{\bf Multiple Images:}\\
NLVR \cite{suhr_corpus_2019-1} & 86k  & yes & \\
Spot-the-difference \cite{jhamtani_learning_2018} &	8k & yes & \\
 \midrule
\multicolumn{4}{l}{\bf OCR:}\\
OCRVQA	\cite{mishra_ocr-vqa_2019} & 50k (subset) & no & \\
DocVQA \cite{mathew_docvqa_2021} &	10k & no & \\
AI2D \cite{kembhavi_diagram_2016} &	3k & no & \\
ChartQA \cite{masry_chartqa_2022} &	18k & no & \\
DVQA \cite{kafle_dvqa_2018} &	50k (subset) & no & \\
ScienceQA \cite{lu_learn_2022} &	6k & no & \\
\midrule
\textbf{Total} & 766k \\
\bottomrule
    \end{tabular}
    }
    \caption{List of datasets included in the \textit{instruct tuning phase} in our analysis experiments. All sizes are based on unique images; examples about the same image are packed into one sequence.}
    \label{tab:dataset:info}
\end{table}

The collections of datasets used in the instruct tuning phase for the analysis experiments (\S\ref{sec:analysis}) is adapted from LLaVA-Next \cite{liu_llava-next_2024}.
As multiple evaluation datasets contain multiple images in the input (MaRVL, VGR, VLOD, M3Exam, xMMMU), we include additional datasets to improve capabilities for this situation.
See Table~\ref{tab:dataset:info} for the full list.

\subsection{For Training \modelname}
\label{sec:appendix:train_data:scaled}

\begin{table}[t]
    \centering
     \def\arraystretch{0.97}
 \resizebox{0.99\linewidth}{!}{
    \begin{tabular}{lrr}
\toprule
     Dataset  &  Size (Images) & Translated?\\
 \midrule
\multicolumn{3}{l}{\bf Natural Image:}\\
ALLaVA Instruct$^1$ \cite{chen_allava_2024} & 760k & yes \\
LVIS Instruct4V \cite{wang_see_2023} & 223k & yes \\
Visual7W \cite{zhu_visual7w_2016} & 14k & no \\
VizWiz QA \cite{gurari_vizwiz_2018} & 21k & no \\
TallyQA \cite{acharya_tallyqa_2019}  & 133k & yes \\
SketchyVQA \cite{tu_how_2023} & 4k & yes \\
OODVQA \cite{tu_how_2023} & 3k & no \\
\midrule
\multicolumn{3}{l}{\bf OCR:}\\
ScienceQA (Cambrian version) & 6k & no \\
AI2D (Cambrian version) & 4k & no \\
Rendered Text $^2$ & 10k & no \\
ScreenQA \cite{hsiao_screenqa_2022} & 33k & no \\
LLaVAR \cite{zhang_llavar_2023} & 20k & no \\
ArxivQA \cite{li_multimodal_2024} & 54k & no \\
Chart2Text \cite{obeid_chart--text_2020} & 25k & no \\
InfographicVQA \cite{mathew_infographicvqa_2022} & 2k & no \\
VisText \cite{tang_vistext_2023} & 10k & no \\
TQA \cite{kembhavi_are_2017} & 1k & no \\
STVQA \cite{biten_scene_2019} & 17k & no \\
TAT-QA \cite{zhu_tat-qa_2021} & 2k & no \\
TabMWP \cite{lu_dynamic_2023} & 23k & no \\
HiTab \cite{cheng_hitab_2022} & 2k & no \\
IconQA \cite{lu_iconqa_2021} & 27k & no \\
VisualMRC \cite{tanaka_visualmrc_2021} & 3k & no \\
RobuT \cite{zhao_robut_2023} & 113k & no \\
FinQA \cite{chen_finqa_2021} & 5k & no \\
\midrule
\multicolumn{3}{l}{\bf Math \& Code:}\\
WebSight \cite{laurencon_unlocking_2024}  & 10k & yes \\
Design2Code \cite{si_design2code_2024} & 0k & yes \\
DaTikz \cite{belouadi_automatikz_2024} & 48k & no \\
CLEVR \cite{johnson_clevr_2017}  & 70k & yes \\
CLEVR-Math \cite{lindstrom_clevr-math_2022}  & 70k & yes \\
Geo170k \cite{gao_g-llava_2023} & 9k & no \\
GeomVerse \cite{kazemi_geomverse_2023} & 9k & no \\
Inter-GPS \cite{lu_inter-gps_2021} & 1k & no \\
MathVision \cite{wang_measuring_2024} & 3k & no \\
Raven \cite{zhang_raven_2019} & 42k & no \\
\midrule
\multicolumn{3}{l}{\bf Text (no images):}\\
Aya Dataset \cite{singh_aya_2024} & 202k & -- \\
Tagengo-GPT4 \cite{devine_tagengo_2024} & 70k & -- \\
Magpie$^2$ \cite{xu_magpie_2024} & 400k & -- \\
\midrule
\bf Total & 2.47M \\
\bottomrule
    \end{tabular}
    }
    \caption{Datasets used on top of the datasets from Table~\ref{tab:dataset:info} for the instruct tuning phase of \modelname.
    $^1$: also contains web-scraped images from LAION \cite{schuhmann_laion-5b_2022} which contain textual elements.
    $2$:\url{https://huggingface.co/datasets/wendlerc/RenderedText}.
    $^2$: Combining \textit{magpie-ultra-v0.1} (50k), \textit{Magpie-Qwen2-Pro-200K-English} (200k), \textit{Magpie-Llama-3.1-Pro-MT-300K-Filtered} (150k subset).
    }
    \label{tab:dataset:info:scaled}
\end{table}

For training \modelname, we combine the datasets from Table~\ref{tab:dataset:info} with additional datasets listed in Table~\ref{tab:dataset:info:scaled}.

\subsection{Synthetic OCR Data}
\label{sec:appendix:train_data:ocr}
We use the official Synthdog code\footnote{\url{https://github.com/clovaai/donut/tree/master/synthdog}} to generate the samples using the Google Noto font and with images from the ImageNet train split as background.
Text is sampled from Wikipedias of the respective languages.

We consider the following 32 languages as not using the Latin script: \textit{am, ar, as, azb, be, bg, bn, bo, el, fa, he, hi, ja, ka, kk, km, ko, lo, mr, my, pa, ru, sa, sd, sr, ta, te, th, ti, uk, ur, zh}.

\section{Evaluation Setup}
\label{sec:appendix:evaluation}
This section describes the details of our evaluation setup.
\subsection{Generation Parameters}
In all experiments of our test suite, we use greedy decoding (\texttt{temperature=0.0; do\_sample=False;}). 
\subsection{Metrics}
Depending on the dataset and task, we employ either CIDEr~\citep{vedantam_cider_2015}, the exact match accuracy, or a relaxed match accuracy (see Table~\ref{appendix:sec:test_suite_datasets}).

For the relaxed match accuracy, we consider an answer correct if it starts with the correct choice letter. For example, answers like ''A.'' are also counted as correct if the gold label is ''A''.
\subsection{Prompts}
\label{appendix:sec:evaluation:prompts}
We list the prompts for each dataset in our test suite used for all models in Figure~\ref{appendix:fig:evaluation:prompts}.
\begin{figure*}[ht!]\tiny
    \centering
    \begin{promptbox}{SMPQA}
    <IMG>\{QUESTION\}\textbackslash{n}Answer the question using a single word or phrase.
    \end{promptbox}
    \begin{promptbox}{CVQA}
    <IMG>\{QUESTION\}\textbackslash{n}There are several options:\textbackslash{n}A. \{OPTION A\}\textbackslash{n}B. \{OPTION B\}\textbackslash{n}C. \{OPTION C\}\textbackslash{n}D. \{OPTION D\}\textbackslash{n}Answer with the option's letter from the given choices directly.
    \end{promptbox}
    \begin{promptbox}{xMMMU}
    \{QUESTION\}\textbackslash{n}There are several options:\textbackslash{n}A. \{OPTION A\}\textbackslash{n}B. \{OPTION B\}\textbackslash{n}C. \{OPTION C\}\textbackslash{n}D. \{OPTION D\}\textbackslash{n}Answer with the option's letter from the given choices directly.
    \end{promptbox}
    \begin{promptbox}{MTVQA}
    <IMG>\{QUESTION\}\textbackslash{n}Answer the question using a single word or phrase.\textbackslash{n}Answer in \{LANGUAGE\}.
    \end{promptbox}
    \begin{promptbox}{M3Exam}
    \{QUESTION\}\textbackslash{n}Options:\textbackslash{n}A. \{OPTION A\}\textbackslash{n}B. \{OPTION B\}\textbackslash{n}C. \{OPTION C\}\textbackslash{n}D. \{OPTION D\}\textbackslash{n} Answer with the option's letter from the given choices directly.
    \end{promptbox}
    \begin{promptbox}{BIN-MC}
    <IMG>Which of these choices (in English) is shown in the image?\textbackslash{n} Choices:\textbackslash{n}A. \{CHOICE A\}\textbackslash{n}B. \{CHOICE B\}\textbackslash{n}C. \{CHOICE C\}\textbackslash{n}D. \{CHOICE D\}\textbackslash{n} Answer with the letter from the given choices directly.
    \end{promptbox}
    \begin{promptbox}{xGQA}
    <IMG>\{QUESTION\}?\textbackslash{n}Answer the question using a single word or phrase.\textbackslash{n}Answer in English.
    \end{promptbox}
    \begin{promptbox}{MaXM}
    <IMG>\{QUESTION\}?\textbackslash{n}Answer the question using a single word or phrase.\textbackslash{n}Answer in \{LANGUAGE\}.
    \end{promptbox}
    \begin{promptbox}{MaRVL}
    <IMG>Given the two images <IMG><IMG>, is it correct to say ``\{HYPOTHESIS\}''? Answer yes or no.'
    \end{promptbox}
    \begin{promptbox}{XVNLI}
    <IMG>Is it guaranteed true that ``\{HYPOTHESIS\}''? Yes, no, or maybe? Answer in English.
    \end{promptbox}
    \begin{promptbox}{M5-VGR}
    Given the two images <IMG><IMG>, is it correct to say ``\{HYPOTHESIS\}''? Answer yes or no.'
    \end{promptbox}
    \begin{promptbox}{M5-VLOD}
    Based on the 5 images <IMG><IMG><IMG><IMG><IMG> ordered from top-left to bottom-right, which image does not match the hypothesis ``\{HYPOTHESIS\}''? Choose one from [A, B, C, D, E] and only output a single letter:
    \end{promptbox}
    \begin{promptbox}{XM3600}
    Briefly describe the image in \{LANGUAGE\} in one sentence.
    \end{promptbox}
    \caption{Prompts used for the different datasets of our test suite. For M3Exam and xMMMU, the questions contain images at individual positions, and also the options can consist of images. In total, a sample of M3Exam can contain up to 8 images and 8 options, and a sample of xMMMU can contain up to 4 images and 4 options.}
    \label{appendix:fig:evaluation:prompts}
\end{figure*}
\subsection{Datasets}
\label{appendix:sec:test_suite_datasets}
In the following, datasets included in our test suite are briefly introduced.
An overview is provided in Table~\ref{appendix:tab:eval_suite_datasets}.
Details about the languages covered by the datasets are listed in Table~\ref{appendix:tab:eval_suite_langs}.
\begin{table*}[htb!]
    \small
    \centering
     \def\arraystretch{0.97}
     \resizebox{0.99\linewidth}{!}{
\begin{tabular}{llllllr}
\toprule
Dataset & Task & Visual Input & Textual Input & Target Output & Metric & \#Lang. \\
\midrule
MaXM & VQA & Single-Image & Question (TL) & WoP (TL) & E. Acc. & $6$ \\
xGQA & VQA & Single-Image & Question (TL) & WoP (EN) & E. Acc. & $8$ \\
XVNLI & VNLI & Single-Image & Hypothesis (TL) & 'yes' / 'no' / 'maybe' & E. Acc. & $5$ \\
M5B-VLOD & VLOD & Multi-Image & Hypothesis (TL) & LoC & R. Acc. & $12$ \\
M5B-VGR & VGR & Multi-Image & Hypothesis (TL) & 'yes' / 'no' & E. Acc. & $12$ \\
MaRVL & VGR & Multi-Image & Hypothesis (TL) & 'yes' / 'no' & E. Acc. & $6$ \\
MTVQA & TH VQA & Single-Image & Question (TL) & WoP (TL) & E. Acc. & $9$ \\
SMPQA - Name & TH VQA & Single-Image & Question (TL) & WoP (TL) & E. Acc. & $11$ \\
SMPQA - Ground & TH VGR & Single-Image & Question (TL) & 'yes' / 'no' & E. Acc. & $11$ \\
M3Exam & TH MC VQA & Single or Multi-Image & Question (TL) & LoC & R. Acc. & $7$ \\
MMMU & TH MC VQA & Single or Multi-Image & Question (EN) & LoC & R. Acc. & $1$ \\
xMMMU & TH MC VQA & Single or Multi-Image & Question (TL) & LoC & R. Acc. & $7$ \\
BabelImageNet-MC & MC VQA & Single-Image & Question (TL) & LoC & R. Acc. & $20$ \\
CVQA & MC VQA & Single-Image & Question (TL) & LoC & R. Acc. & $39$ \\
XM3600 & Captioning & Single-Image & Prompt (EN) & Caption (TL) & CIDEr & $36$ \\
\bottomrule
\end{tabular}
    }
    \caption{List of datasets contained in our test suite. In the Task column, ''VQA'' ''VNLI'', ''VLOD'', ''VGR'', ''TH'', and ''MC'' are acronyms for ''Visual Question Answering'', ''Visual Natural Language Inference'', ''Visio-Linguistic Outlier Detection'', ''Visually Grounded Reasoning'', ''Text-Heavy'', and ''Multiple-Choice'', respectively. In the ''Textual Input'' and ''Target Output'' columns,  the acronyms ''WoP'', ''LoC'', ''TL'', and ''EN'' stand for ''(Single) Word or Phrase'', ''Letter of the correct Choice'', ''Target Language'', and ''English'', respectively. Further, ''E. Acc.'' is ''Exact Accuracy'' and ''R. Acc.'' is ''Relaxed Accuracy''.
    CVQA is not used in  \S\ref{sec:analysis} due to its hidden test set with limited submissions.
    }
    \label{appendix:tab:eval_suite_datasets}
\end{table*}
\begin{table*}[htb!]
    \small
    \centering
     \def\arraystretch{0.97}
     \resizebox{0.99\linewidth}{!}{
        \begin{tabular}{llllp{30em}}
        \toprule
        Name & Tier & ISO-639-3 & ISO-639-1 & Datasets \\
        \midrule
        Afrikaans & 3 & afr & af & BabelImageNet-MC, M3Exam \\
        Amharic & 2 & amh & am & BabelImageNet-MC, CVQA, M5B-VGR, M5B-VLOD \\
        Arabic & 5 & ara & ar & MTVQA, SMPQA, XM3600, xMMMU, XVNLI \\
        Bengali & 3 & ben & bn & CVQA, M5B-VGR, M5B-VLOD, xGQA, XM3600 \\
        Berber (macrolanguage) & 0 & ber & - & M5B-VGR, M5B-VLOD \\
        Breton & 1 & bre & br & CVQA \\
        Bulgarian & 3 & bul & bg & CVQA \\
        Chinese & 5 & zho & zh & CVQA, M3Exam, MaRVL, MaXM, SMPQA, xGQA, XM3600 \\
        Croatian & 4 & hrv & hr & BabelImageNet-MC, XM3600 \\
        Cusco Quechua & 1 & quz & - & XM3600 \\
        Czech & 4 & ces & cs & BabelImageNet-MC, XM3600 \\
        Danish & 3 & dan & da & XM3600 \\
        Dutch & 4 & nld & nl & BabelImageNet-MC, XM3600 \\
        Egyptian Arabic & 3 & arz & - & CVQA \\
        English & 5 & eng & en & BabelImageNet-MC, M3Exam, M5B-VGR, M5B-VLOD, MaRVL, MaXM, MME, MMMU, SMPQA, xGQA, XM3600, xMMMU, XVNLI \\
        Filipino & 3 & fil & - & CVQA, M5B-VGR, M5B-VLOD, XM3600 \\
        Finnish & 4 & fin & fi & BabelImageNet-MC, XM3600 \\
        French & 5 & fra & fr & MaXM, MTVQA, XM3600, xMMMU, XVNLI \\
        German & 5 & deu & de & M5B-VGR, M5B-VLOD, MTVQA, SMPQA, xGQA, XM3600 \\
        Hausa & 2 & hau & ha & BabelImageNet-MC, M5B-VGR, M5B-VLOD \\
        Hebrew & 3 & heb & he & XM3600 \\
        Hindi & 4 & hin & hi & M5B-VGR, M5B-VLOD, MaXM, SMPQA, XM3600, xMMMU \\
        Hungarian & 4 & hun & hu & BabelImageNet-MC, XM3600 \\
        Igbo & 1 & ibo & ig & CVQA \\
        Indonesian & 3 & ind & id & CVQA, MaRVL, SMPQA, xGQA, XM3600, xMMMU \\
        Irish & 2 & gle & ga & CVQA \\
        Italian & 4 & ita & it & M3Exam, MTVQA, SMPQA, XM3600 \\
        Japanese & 5 & jpn & ja & BabelImageNet-MC, CVQA, MTVQA, XM3600, xMMMU \\
        Javanese & 1 & jav & jv & CVQA \\
        Kanuri & 0 & kau & kr & CVQA \\
        Kinyarwanda & 1 & kin & rw & CVQA \\
        Korean & 4 & kor & ko & CVQA, SMPQA, xGQA, XM3600 \\
        Malay (macrolanguage) & 3 & msa & ms & CVQA \\
        Maori & 1 & mri & mi & BabelImageNet-MC, XM3600 \\
        Mi-gkabau & 1 & min & - & CVQA \\
        Modern Greek & 3 & ell & el & BabelImageNet-MC, XM3600 \\
        Mongolian & 1 & mon & mn & CVQA \\
        Norwegian & 1 & nor & no & BabelImageNet-MC, CVQA, XM3600 \\
        Oromo & 1 & orm & om & CVQA \\
        Persian & 4 & fas & fa & BabelImageNet-MC, XM3600 \\
        Polish & 4 & pol & pl & BabelImageNet-MC, XM3600 \\
        Portuguese & 4 & por & pt & CVQA, M3Exam, xGQA, XM3600, xMMMU \\
        Romanian & 3 & ron & ro & BabelImageNet-MC, CVQA, MaXM, XM3600 \\
        Russian & 4 & rus & ru & CVQA, M5B-VGR, M5B-VLOD, MTVQA, SMPQA, xGQA, XM3600, XVNLI \\
        Sinhala & 0 & sin & si & CVQA \\
        Spanish & 5 & spa & es & BabelImageNet-MC, CVQA, XM3600, XVNLI \\
        Sundanese & 1 & sun & su & CVQA \\
        Swahili (macrolanguage) & 2 & swa & sw & CVQA, M5B-VGR, M5B-VLOD, MaRVL, XM3600 \\
        Swedish & 4 & swe & sv & XM3600 \\
        Tamil & 3 & tam & ta & BabelImageNet-MC, CVQA, MaRVL \\
        Telugu & 1 & tel & te & BabelImageNet-MC, CVQA, XM3600 \\
        Thai & 3 & tha & th & M3Exam, M5B-VGR, M5B-VLOD, MaXM, MTVQA, SMPQA, XM3600 \\
        Turkish & 4 & tur & tr & MaRVL, XM3600 \\
        Ukrainian & 3 & ukr & uk & XM3600 \\
        Urdu & 3 & urd & ur & CVQA \\
        Vietnamese & 4 & vie & vi & M3Exam, MTVQA, XM3600 \\
        Zulu & 2 & zul & zu & BabelImageNet-MC, M5B-VGR, M5B-VLOD, SMPQA \\
        \midrule
        Unique Languages & \multicolumn{4}{l}{56 (43 without CVQA)} \\
        \bottomrule
    \end{tabular}
    }
    \caption{List of languages covered in the datasets of our test suite. The ''Tier'' column represents the tier in the taxonomy proposed by \citet{joshi_state_2020}, where a higher tier indicates more available resources, i.e., data, in the respective language.
    CVQA is not used in  \S\ref{sec:analysis} due to its hidden test set with limited submissions.}
    \label{appendix:tab:eval_suite_langs}
\end{table*}
\paragraph{xGQA}
The xGQA dataset~\citep{pfeiffer_xgqa_2022} is a cross-lingual visual question-answering dataset.
It extends the well-known English-only GQA dataset~\citep{hudson_gqa_2019} by manually translating the questions in the balanced \textit{test-dev} set.
Each of the $9666$ questions is available in eight languages covering five scripts, while the answers are in English only.
The dataset holds $300$ unique images from Visual Genome~\citep{krishna_visual_2017}.
%
\paragraph{MaXM}
The MaXM dataset was introduced by~\citet{changpinyo_maxm_2022} and is a VQA dataset comprising seven languages in five scripts.
In MaXM, the questions and their respective answers are in the same language.
The images are a subset of the XM3600~\citep{thapliyal_xm3600_2022} dataset and are chosen to match a region where the language of the question-answer pair is spoken.
This ensures cultural diversity in the images in addition to the language diversity in the question-answer texts.
%
%
\paragraph{XVNLI}
The XVNLI dataset~\citep{bugliarello_iglue_2022} introduces the task of Cross-lingual Visual Natural Language Inference where a model needs to predict whether a textual hypothesis \textit{entails}, \textit{contradicts}, or is \textit{neutral} concerning a visual premise.
XVNLI comprises five languages covering three scripts and $357$ unique images from Visual Genome.
It is based on a combination of the text-only SNLI~\citep{bowman_large_2015} dataset and its cross-lingual~\citep{agic_cli_2018} and cross-modal~\citep{xie_visual_entailment_2019} equivalents.
%
%
\paragraph{MaRVL}
The MaRVL dataset~\citep{liu_visually_2021} aims to benchmark models on Multicultural Reasoning over Vision and Language.
A task sample comprises two images, a textual statement, and a binary true or false answer grounded in the images.
MaRVL comprises five languages covering three scripts and $4914$ culturally diverse images that match the respective languages.
The images in a sample are chosen to match the culture of the annotator who has written the textual statement in his or her native language.
%
%
%
\paragraph{XM3600}
The XM3600 dataset~\citep{thapliyal_xm3600_2022} is a large multilingual image captioning dataset comprising $36$ languages with $261375$ captions covering $13$ different scripts for $100$ unique images per language.
The images are selected to match the language's cultural background, ensuring cultural and linguistic diversity.
The captions were not automatically translated but manually created by professional annotators who are native speakers of the respective language.

We only use a subset of 500/3600 images (selected randomly) per language when evaluating XM3600 due to its size.

%
%
\paragraph{Babel-ImageNet (multiple-choice) (BIN-MC)} 
Babel-ImageNet~\cite{geigle_babel-imagenet_2023} translates the labels of ImageNet \cite{deng_imagenet_2009} to nearly 300 languages, which allows us to test if models are capable of recognizing and linking the diverse objects of ImageNet to their correct label in the tested language.
Testing all 300 languages would be too expensive, instead we use it to deepen our evaluation in languages appearing in only 1 or 2 other datasets, plus English and select few high-resource languages.
Also, we only use 10 images per class instead of 50, again, to keep computational cost reasonable.

We formulate the task as a multiple-choice problem, following the approach by \citet{geigle_african_2024} to mine hard negative options from the total label pool.
This avoids problems of unclear or underspecified answers that appear in a traditional open-ended VQA formulation.
We mine negatives with the English labels, filtering out all candidates not translated by Babel-ImageNet in the target language, that is, in the end, we select the three most similar negative labels that appear in the Babel-ImageNet labels of a given language.
%
\paragraph{SMPQA}
We propose \textbf{SMPQA} (Synthetic Multilingual Plot QA) as a novel test dataset for evaluating multilingual OCR capabilities in images -- bar plots and pie charts to be specific -- in 11 languages, covering different scripts and resource levels. See \S\ref{sec:appendix:evaluation:smpqa} for details.
%
%
%
\paragraph{M5B-VGR}
The M5B-VGR dataset is a Visually Grounded Reasoning dataset similar to MaRVL and was introduced by~\cite{schneider_m5_2024}.
A sample comprises two images, a textual statement, and a binary true or false answer grounded in the images.
It comprises 12 languages covering 7 scripts and culturally diverse photos taken in regions where the respective language is spoken.
The images are sampled from the Dollar Street~\cite{gaviria2022dollar} dataset.
For each language, there are $120$ samples.
%
%
\paragraph{M5B-VLOD}
The M5B-VLOD (Visio-Linguistic Outlier Detection) dataset was introduced by~\cite{schneider_m5_2024}.
A sample comprises five images and a textual statement that is true for all but one of the images.
The task is to find the outlier image, that is, the image that does not match the statement.
It comprises the same 12 languages as M5B-VGR and images sampled with a similar strategy from the same dataset.
For each language, there are $120$ samples.
%
%
\paragraph{MTVQA}
The MTVQA dataset was introduced by~\cite{tang_mtvqa_2024} and comprises text-heavy Visual Question Answering (VQA) tasks.
It features human expert annotations across 9 diverse languages, consisting of a total of 6778 question-answer pairs across 2116 images.
The images primarily contain text in the respective language and the question (and answer) related to that text.
The images are sampled from different publicly available datasets.
%

%
\paragraph{CVQA}
The CVQA dataset was introduced by~\cite{romero_cvqa_2024} and is a multilingual, culturally nuanced VQA benchmark that includes a diverse set of languages, many of them underrepresented and understudied in NLP.
It consists of 10000 questions across 30 countries, covering 31 languages, and in 39 distinct country-language pairs (e.g., the dataset includes 7 different splits for Spanish because it contains 7 countries where Spanish is spoken).
The images in the dataset were manually gathered by human annotators to match and depict the culture of the respective country-language pair.
A sample consists of one image and a question related to the image in the respective language.
The authors did not release the test set publicly but allowed up to 5 daily submissions to their leaderboard to obtain evaluation results.
%
%
\paragraph{M3Exam}
The M3Exam dataset was introduced by~\cite{zhang_m3exam_2023}.
It contains real-world exam questions in 9 languages, which are either text-only or multi-modal.
In our test suite, we only consider samples that require at least one image.
Further, due to the low number of resulting samples for Swhalili and Javanese, we only include the remaining 7 languages.
The remaining samples consist of multiple-choice questions in the target language and up to 8 images that can appear both in the question and the answer options.
Further, the number of options ranges from 4 to 8 depending on the individual sample.
%

%
\paragraph{xMMMU}
The xMMMU was introduced by~\cite{yue2024pangea} and consists of college-level multiple-choice VQA samples across seven languages.
It was automatically translated using GPT4o from a subset of 300 randomly selected questions from the MMMU~\cite{yue_mmmu_2023} validation split.
\subsection{Details for SMPQA}
\label{sec:appendix:evaluation:smpqa}
\begin{figure}[th]
    \centering
\begin{subfigure}{1\linewidth}
    \includegraphics[width=\textwidth]{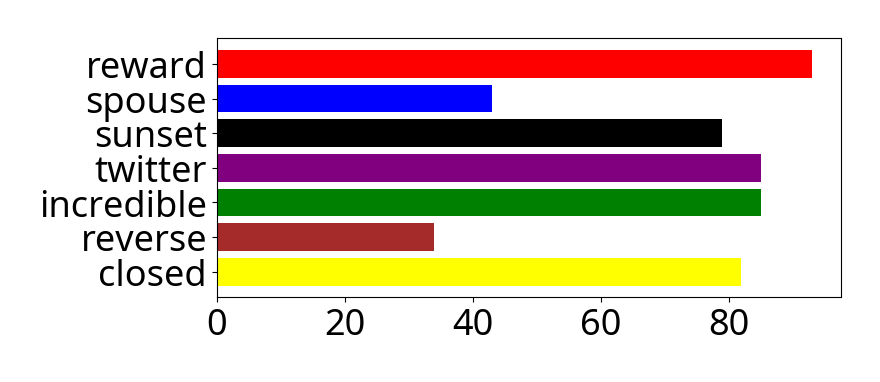}
    \caption{Example of a bar plot in SMPQA for English. \\
    \textbf{Questions for Grounding:}
            "Is the bar with label 'reward' the biggest?",
            "Is the bar with label 'incredible' the biggest?",
            "Is the bar with label 'reverse' the smallest?",
            "Is the bar with label 'sunset' the smallest?",
            "Is the bar with label 'closed' colored in yellow?",
            "Is the bar with label 'closed' colored in purple?",
            "Is the bar with label 'twitter' colored in purple?",
            "Is the bar with label 'twitter' colored in red?" \\
\textbf{Questions for Reading:}
            "What is the label of the biggest bar?",
            "What is the label of the smallest bar?",
            "What is the label of the yellow bar?",
            "What is the label of the red bar?",
            "What is the label of the purple bar?"
    }
    \label{fig:smpqa:bar}
\end{subfigure} 
\begin{subfigure}{1\linewidth}
    \includegraphics[width=\textwidth]{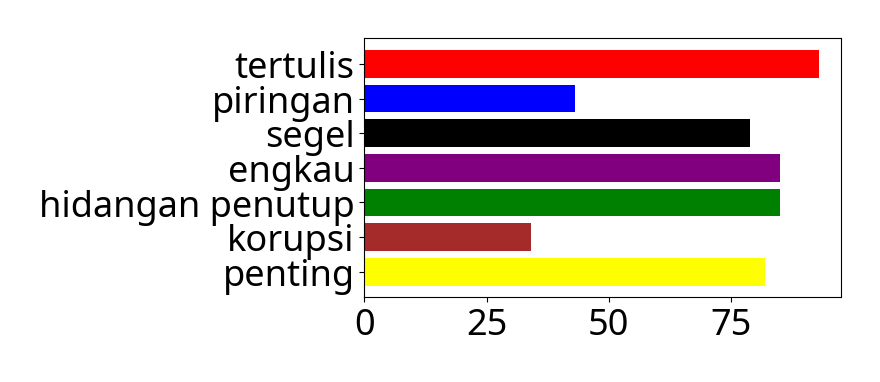}
    \caption{The same plot in Indonesian. Note that all questions refer to the same parts of the plot as the English version just with different words for labels. \\
    \textbf{Questions for Grounding:}
            "Is the bar with label 'tertulis' the biggest?",
            "Is the bar with label 'hidangan penutup' the biggest?",
            "Is the bar with label 'korupsi' the smallest?",
            "Is the bar with label 'segel' the smallest?",
            "Is the bar with label 'penting' colored in yellow?",
            "Is the bar with label 'penting' colored in purple?",
            "Is the bar with label 'engkau' colored in purple?",
            "Is the bar with label 'engkau' colored in red?"\\
\textbf{Questions for Reading:}
            "What is the label of the biggest bar?",
            "What is the label of the smallest bar?",
            "What is the label of the yellow bar?",
            "What is the label of the red bar?",
            "What is the label of the purple bar?"
    }
    \label{fig:smpqa:pie}
\end{subfigure}
    \caption{Examples of one plot configuration in SMPQA for English and Indonesian.}
    \label{fig:smpqa}
\end{figure}

We propose \textbf{SMPQA} (Synthetic Multilingual Plot QA) as a test dataset for evaluating multilingual OCR capabilities, that is capabilities to identify and read text in various languages in images, specifically bar plots and pie charts.

We test the capabilities in two directions: i) \textit{grounding} requires the model to ground a given label in the user prompt to the corresponding part in the plot to answer a yes/no question (``Is the bar with label \$X the biggest?''); ii) \textit{reading} requires the model to output the label of a specified part of the plot (``What is the label of the biggest slice?''). 
The questions are simple by design, requiring minimal reasoning, math, or multi-hop capabilities, as to isolate solely the OCR capabilities in the tested language.
We show example plots and questions in Figure~\ref{fig:smpqa}.

We use exact match accuracy for both tasks. For reading, edit distance to the correct word would be a fine-grained alternative but since word lengths differ between languages -- Chinese can be 1-2 characters while Indonesian can be >10 -- we opt against it to more easily compare results between languages.
To have a fair comparison between languages, we construct the dataset in a way that plots and questions about them are identical between languages (except for labels in the respective languages, obviously).

\paragraph{Construction:} SMPQA is constructed with a deterministic pipeline yielding identical results for each language.

\noindent 1. We define a list of diverse pie charts and bar plots by randomly sampling the number of bars/slices, the size of each, their colors, the plot size and aspect ratio, and vertical/horizontal orientation for bar plots, and exploding some slices in pie charts.
For each plot type, we define 50 configurations, so we have 100 plots/images in total per language.

\noindent 2. Using word lists of common words in the languages, we sample words for use as labels for the bars and pie slices to fill and ultimately render the pre-defined plots. This means the plots are identical between languages except for the labels and some size adjustments caused by different word lengths.

\noindent 3. For each plot, we use templates to generate 5 questions for reading and 8 questions for grounding (with balanced `yes' and `no' as answers). The questions are always the same for a plot, so each language has the same questions, just with different labels.

\paragraph{Language Selection:}
We selected the languages as follows: For Latin-script languages, we chose English and one language from Tier 5 to 2 to have both high- and low-resource languages: German, Italian, Indonesian, and Zulu.
For non-Latin scripts, we select 6 languages to represent scripts with high usage in the world: Russian (Cyrillic), Chinese, Korean (Hangul), Hindi (Devanagari), Arabic, and Thai.

We note that our dataset construction can easily be extended to other languages if needed (as long as word lists are available) to test, for example, more scripts (Telugu, Greek, Hebrew, ...) or languages using the Latin script with heavy use of diacritics (Vietnamese, Turkish, ...).
This makes SMPQA an ideal starting point for probing OCR capabilities in diverse languages.

\subsection{Baseline Models}
\label{sec:appendix:evaluation:baselines}
We list the evaluated baseline models in Table~\ref{tab:appendix:baselines}.
In all baseline evaluation experiments, we use greedy decoding (\texttt{temperature=0.0; do\_sample=False;}).
Further, we do not pre-process the images in any way and use the provided code for inference with the respective model.
\begin{table}[t]
    \centering
     \def\arraystretch{0.97}
     \resizebox{0.99\linewidth}{!}{
    \begin{tabular}{ll}
\toprule
HuggingFace Model ID & Params \\
\midrule
\href{https://www.huggingface.com/Qwen/Qwen2-VL-2B-Instruct}{Qwen/Qwen2-VL-2B-Instruct}~\citep{wang2024qwen2vl} & $2$B \\
\href{https://www.huggingface.com/Qwen/Qwen2-VL-7B-Instruct}{Qwen/Qwen2-VL-7B-Instruct}~\citep{wang2024qwen2vl} & $7$B \\
\href{https://www.huggingface.com/microsoft/Phi-3.5-vision-instruct}{microsoft/Phi-3.5-vision-instruct}~\citep{abdin2024phi} & $4$B \\
\href{https://www.huggingface.com/neulab/Pangea-7B-hf}{neulab/Pangea-7B-hf}~\citep{yue2024pangea} & $7$B \\
\href{https://www.huggingface.com/openbmb/MiniCPM-V-2_6}{openbmb/MiniCPM-V-2\_6}~\citep{yao2024minicpmv} & $8$B \\
\href{https://www.huggingface.com/meta-llama/Llama-3.2-11B-Vision-Instruct}{meta-llama/Llama-3.2-11B-Vision-Instruct}~\citep{aimeta_llama_2024} & $11$B \\
\href{https://www.huggingface.com/mistralai/Pixtral-12B-2409}{mistralai/Pixtral-12B-2409}~\citep{agrawal2024pixtral} & $12$B \\
\href{https://www.huggingface.com/AIDC-AI/Parrot-7B}{AIDC-AI/Parrot-7B}~\citep{sun2024parrot} & $7$B \\
\href{https://www.huggingface.com/MBZUAI/PALO-7B}{MBZUAI/PALO-7B}~\citep{maaz2024palo} & $7$B \\
\href{https://www.huggingface.com/MBZUAI/PALO-13B}{MBZUAI/PALO-13B}~\citep{maaz2024palo} & $13$B \\
\href{https://www.huggingface.com/OpenGVLab/InternVL2_5-4B}{OpenGVLab/InternVL2\_5-4B}~\citep{chen2024internvl2_5} & $4$B \\
\href{https://www.huggingface.com/OpenGVLab/InternVL2_5-8B}{OpenGVLab/InternVL2\_5-8B}~\citep{chen2024internvl2_5} & $8$B \\
\href{https://www.huggingface.com/maya-multimodal/maya}{maya-multimodal/maya}~\citep{alam2024maya} & $8$B \\
\bottomrule
\end{tabular}
}
\caption{List of models considered in our evaluation experiments.}
    \label{tab:appendix:baselines}
\end{table}

We use relaxed match accuracy for all tasks even if \modelname uses exact match for a fairer comparison because some models struggled with replying just `yes'/`no' and similar issues.

\section{Additional Experiments}

\subsection{Analysis Results with Llama 3}
\label{sec:appendix:experiments:analysis_llama}

We report the results with Llama 3 when repeating the experiments of \S\ref{sec:analysis:ex1}, \ref{sec:analysis:ex2}, \ref{sec:analysis:ex3} in Table~\ref{tab:ex1:average_llama}, \ref{tab:ex2:average_llama}, \ref{tab:ex3:llama}.

\begin{table}[t]
 \def\arraystretch{0.97}
 \resizebox{0.99\linewidth}{!}{
\begin{tabular}{lrrrrrr}
\toprule
\bf Train Lang. & \bf T1 & \bf T2 & \bf T3 & \bf T4 & \bf T5 & \bf en \\
\midrule
English & 16.1 & \underline{34.7} & 26.3 & 24.3 & 26.2 &  \underline{56.4} \\
T5 & \underline{19.1} & 32.5 & \underline{29.3} & \underline{27.2} & \underline{35.5} & 54.3 \\
L100 & \textbf{31.1} & \textbf{43.0} & \textbf{39.4} & \textbf{35.9} & \textbf{36.4} & \textbf{56.6} \\
\midrule
\multicolumn{6}{l}{\bf Without tasks affected by language fidelity:} \\
English & 36.6 & \underline{37.1} & 39.0 & 39.6 & 40.0 &  \underline{54.6} \\
T5 & \underline{38.8} & 34.8 & \underline{40.1} & \underline{40.2} & \underline{40.4} & 53.5 \\
L100 & \textbf{46.3} & \textbf{44.0} & \textbf{45.0} & \textbf{42.8} & \textbf{42.9} &  \textbf{55.3} \\
\bottomrule
\end{tabular}
}
\caption{Experimental setup of Table~\ref{tab:ex1} repeated with Llama 3 and the setups: just English, T5 languages, and L100 languages.}
\label{tab:ex1:average_llama}
\end{table}

\begin{table}[t]
     \def\arraystretch{0.97}
     \resizebox{0.99\linewidth}{!}{
\begin{tabular}{lrrrrrr}
\toprule
\bf English \% & \bf T1 & \bf T2 & \bf T3 & \bf T4 & \bf T5 & \bf en \\
\midrule
10 & \textbf{32.9} & \textbf{43.1} & \underline{38.7} & \underline{35.4} & 35.4 & 54.2 \\
50 & \underline{31.1} & \underline{43.0} & \textbf{39.4} & \textbf{35.9} & \textbf{36.4} & \underline{56.6} \\
90 & 26.9 & 38.7 & 36.9 & 34.2 & \underline{35.8} & \textbf{56.6} \\
\bottomrule
\end{tabular}
}
    \caption{Experimental setup of Table~\ref{tab:ex2} repeated with Llama 3 and the setups: 10, 50, and 90\% English instruct tune data.}
    \label{tab:ex2:average_llama}
\end{table}

\begin{table}[t]
    \centering
    \footnotesize
     \def\arraystretch{0.97}
     \resizebox{0.99\linewidth}{!}{
\begin{tabular}{lrrrrrr}
\toprule
\bf English \% & \bf T1 & \bf T2 & \bf T3 & \bf T4 & \bf T5 & \bf en \\
\midrule
No pretrain & 31.1 & 43.0 & 39.4 & 35.9 & 36.4 & 56.6 \\
100 & \underline{33.9} & \underline{44.7} & \underline{43.3} & \underline{39.9} & \underline{39.9} & \underline{60.8} \\
1 & \textbf{37.8} & \textbf{47.4} & \textbf{45.0} & \textbf{41.1} & \textbf{40.7} & \textbf{61.4} \\
\bottomrule
\end{tabular}
}
    \caption{Results of Table~\ref{tab:ex3} repeated with Llama 3 and the setups: 1 and 100\% English pre-train data.}
    \label{tab:ex3:llama}
\end{table}

\subsection{Non-Uniform Language Allocation}
\label{sec:appendix:experiments:stratified}

In our experiments in \S\ref{sec:analysis}, we distribute the non-English portion of the data uniformly over all languages.
We now consider two \textit{stratified} distributions that upsample low-resource languages.
A language with taxonomy $i$ will get allocated the following portion of the non-English data:
\begin{equation}
    p(i) = \frac{f(i)}{\sum_{j\in\mathrm{Train Languages}} f(j)}
\end{equation}
with $f(i)=\frac{1}{i}$ for \textbf{Stratified-1} and $f(i)=\frac{1}{\exp i}$ for \textbf{Stratified-2}.
This effectively doubles the allocated data for T1 languages, and divides the data for T5 languages by a factor 3 or 20 (depending on Stratified-1 or -2).

Results are reported in Table~\ref{tab:appendix:ex_stratified}.
We do observe a small decrease for T5 and T4 languages, and also for T3 with Stratified-2, but results for T1 and T2 languages stay relatively constant despite more data.
This suggests that higher-resource languages can be quite sample efficient even with what amounts to a few hundred samples (at least in the instruct tuning phase) but as the stratified distributions fail to improve lower-resource languages, there is little reason in practice to not use the uniform distribution which makes no assumptions about the resource-level of a language.

\begin{table}[t]
    \centering
     \def\arraystretch{0.97}
     \resizebox{0.99\linewidth}{!}{
\begin{tabular}{lrrrrrr}
\toprule
\bf Distribution & \bf T1 & \bf T2 & \bf T3 & \bf T4 & \bf T5 & \bf en \\
\midrule
Uniform & \underline{18.9} & \underline{32.6} & \underline{30.7} & \textbf{28.8} & \textbf{34.4} & \underline{52.6} \\
Stratified-1 & 18.6 & 32.5 & \textbf{30.7} & \underline{28.0} & 33.8 & \textbf{53.0} \\
Stratified-2 & \textbf{19.2} & \textbf{32.6} & 29.5 & 27.4 & \underline{33.9} & 52.0 \\
\bottomrule
\end{tabular}
}
    \caption{Comparison between our uniform allocation of data compared to two stratified allocations that upsample low-resource languages.}
    \label{tab:appendix:ex_stratified}
\end{table}

\subsection{LLM Comparison}
\label{sec:appendix:experiments:llm}
\begin{table}[t]
    \centering
     \def\arraystretch{0.97}
     \resizebox{0.99\linewidth}{!}{
\begin{tabular}{lrrrrrr}
\toprule
\bf LLM & \bf T1 & \bf T2 & \bf T3 & \bf T4 & \bf T5 & \bf en \\
\midrule
\texttt{Phi-3.5-mini-instruct} & 18.9 & 32.6 & 30.7 & 28.8 & 34.4 & 52.6 \\
\texttt{gemma-2-9b-it} & 29.2 & 40.9 & 36.4 & 33.5 & 35.3 & 52.8 \\
\texttt{Meta-Llama-3-8B-Instruct} & \textbf{31.1} & \underline{43.0} & 39.4 & 35.9 & 36.4 & 56.6 \\
\texttt{Qwen2.5-7B-Instruct} & \underline{30.7} & \textbf{43.7} & \underline{42.0} & \underline{38.1} & \underline{40.5} & \textbf{62.7} \\
\texttt{aya-expanse-8b} & 28.3 & 42.5 & \textbf{43.0} & \textbf{39.8} & \textbf{40.9} & \underline{59.9} \\
\bottomrule
\end{tabular}
}
    \caption{Comparison between different LLM backbones all trained with the instruct tuning data with L100 languages and 50\% English (as in \S\ref{sec:analysis:ex2}).}
    \label{tab:appendix:ex_llm}
\end{table}

We train several recent 7-9B parameter LLMs on the instruct tuning data mix used in our analysis with L100 languages and 50\% English. All models are trained with the same hyperparameters.
We compare Llama 3 \cite{llama_team_llama_2024}\footnote{While Llama 3.1 has officially better multilingual capabilities, we found Llama 3 to work better.}, Gemma 2 \cite{gemma_team_gemma_2024}, Aya-Expanse \cite{aryabumi_aya_2024,dang2024ayaexpansecombiningresearch}, and Qwen 2.5 \cite{yang_qwen2_2024}.

Table~\ref{tab:appendix:ex_llm} shows that Qwen and Aya yield the overall best results with Aya slightly ahead in T3-T5 and Qwen (with Llama 3) better in T1 and T2.
Qwen achievs by far the best English results.
Qwen is also notably strong in the exam tasks (M3Exam, xMMMU) and is the only model with better-than-guessing results on average for VLOD.



\section{Qualitative Examples}
\label{sec:appendix:qualitative}

\begin{figure*}
    \centering
    \tiny
    \includegraphics[width=0.5\linewidth]{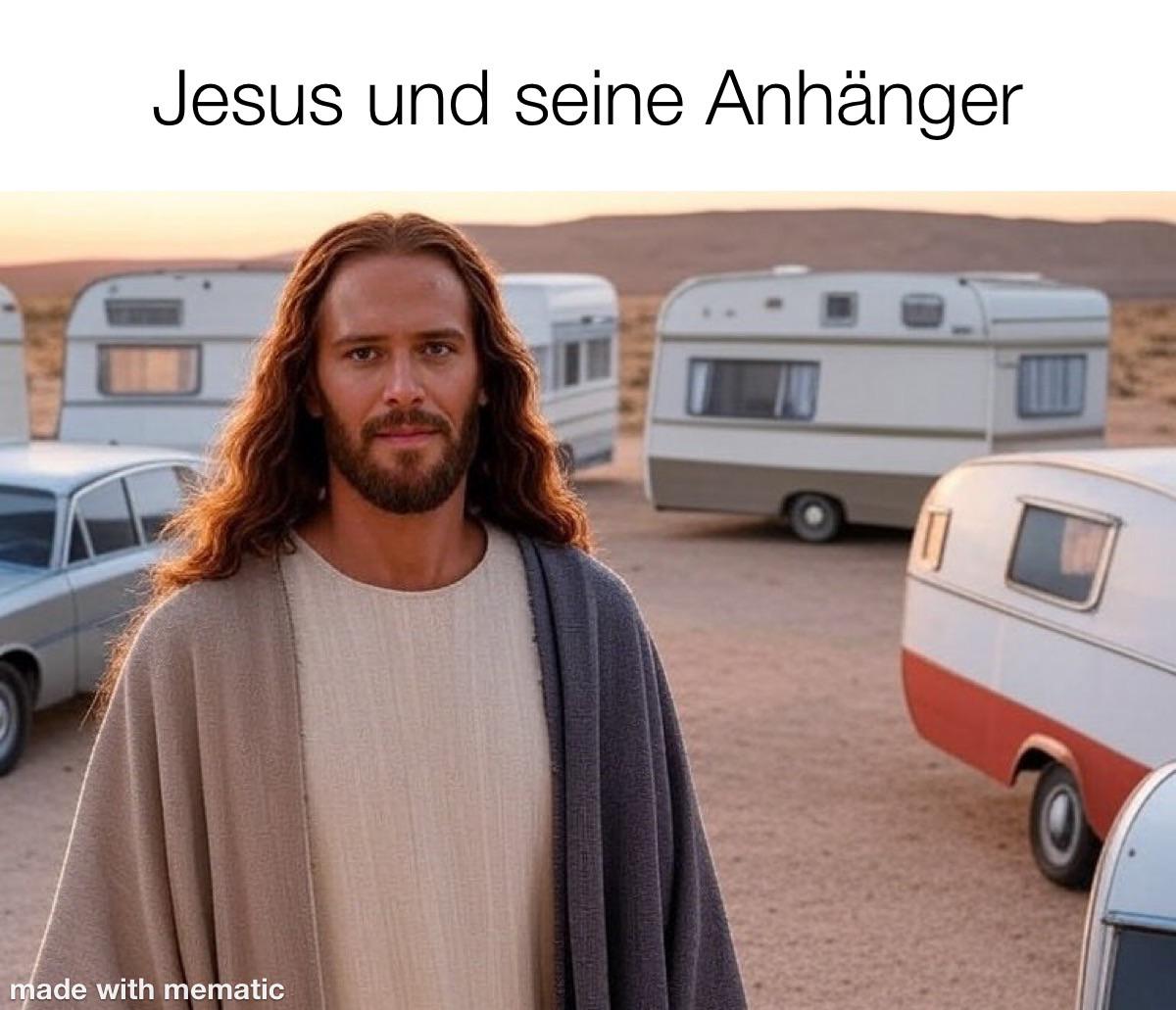}
    \begin{examplebox}{Can you explain the joke of this German meme?}
    \textbf{\modelname \texttt{Aya}}:
The humor in this image comes from juxtaposing a biblical figure, Jesus, with an everyday scene that is both ordinary and unexpected. The man's appearance as Jesus, complete with long hair and a beard, contrasts with the mundane setting of a caravan park populated by caravans (trailers). This creates a humorous effect because it takes something familiar (Jesus) and places it into a context that is not only unfamiliar but also somewhat banal or trivial. It plays on the idea of finding divinity in the most unexpected places, which can be seen as a commentary on how we perceive holiness or significance. Additionally, the caption "Jesus und seine Anhänger" further emphasizes the absurdity by referencing followers, implying that even in such an unremarkable location, one could find disciples or supporters, thus adding another layer to the comedic effect.
\\\\
    \textbf{\modelname \texttt{Qwen}}: The joke in this German meme plays on a pun. The phrase "Jesus und seine Anhänger" translates to "Jesus and his followers." However, the word "Anhänger," which means "followers," is also slang for caravans or RVs (recreational vehicles) in some contexts. Therefore, the image humorously suggests that Jesus has modern-day followers who travel together in caravans instead of following him spiritually as he did with his original disciples. This juxtaposition creates a humorous effect by combining religious iconography with contemporary culture.

    \end{examplebox}
    \caption{\modelname explaining a German meme. Both models correctly read the caption but only \texttt{Qwen} `understands' the pun and gets the explanation (mostly) correct.}
    \label{fig:qualitative_examples:1}
\end{figure*}

\begin{figure*}
    \centering
    \tiny
    \includegraphics[width=0.5\linewidth]{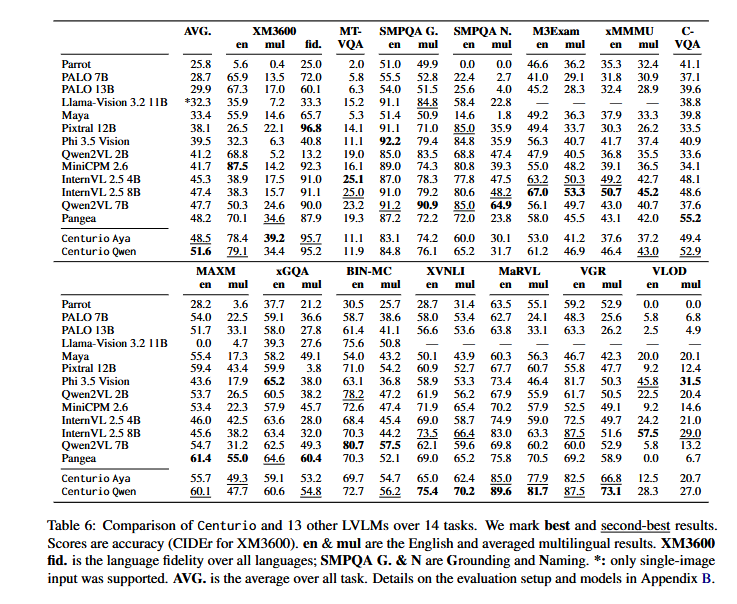}
    \begin{examplebox}{Summarize the table for me.}
    \textbf{\modelname \texttt{Aya}}:
This is a detailed summary of the table:

    The table compares Centurio and 13 other LLM models on 14 different tasks, using CIDEr as an evaluation metric.
    It presents accuracy scores (AVG., MTXMUL) and their standard deviations (SD.) across various datasets like SQuAD G \& N, Image captioning, etc.
    Models are listed in descending order based on AVG score, with Paratrol being at the top followed by others like PALO-TB, Lama Vision, etc.
    There's also information about model names, task descriptions, and additional metrics like BIN\_MC, XNLI, MARVL, VGRU, VLOD, and C.Q.A.
    A note explains that only one image input was used for evaluating language fidelity over all languages, and it mentions "CENTURIO QEN" which might be a specific version or variant of the Centurio model.
    The bottom part of the table provides further details about the evaluation setup and models' performance on these tasks.
    Overall, this table seems to evaluate the Centurio model against several others in terms of its ability to perform well on multiple NLP tasks.
\\\\
    \textbf{\modelname \texttt{Qwen}}: Here a summary:

    The best model is Centurio Aya.\\
    The worst model is Palo 7B.\\
    The average score of all models is around 50\%.\\
    The most accurate model (Centurio Aya) has an accuracy over 63\% in all tasks, while the least accurate one (Palo 7B) has an accuracy under 28\%.\\
    In multilingual results, the best model is also the most accurate one and vice versa.\\
    The most accurate model has an overall language fidelity over 94\%, whereas the least accurate one has an overall language fidelity below 13\%.\\
    All models have an input supported AVG. above 40\%.\\
    The highest number of errors made by any model was 128, which occurred on the Inten/VL 2.5 SB task.\\
    The lowest number of errors made by any model was 1, which occurred on the MiniCPM 2.6 task. 
    \end{examplebox}
    \begin{examplebox}{What model is best? Explain why.}
    \textbf{\modelname \texttt{Aya}}:
Centurio! QEN, with an AVG., MAMM/U 32.0 and C.QAL 41.8. It has the highest average over all tasks compared to other models listed in table 6, indicating its superior performance across multiple linguistic tests.
\\\\
    \textbf{\modelname \texttt{Qwen}}: Centurio Aya Centurio'Queen. It has the highest accuracy score (65.7) and lowest error rate (20.3).

    \end{examplebox}
    \caption{\modelname struggle with images containing small text.}
    \label{fig:qualitative_examples:2}
\end{figure*}

\begin{figure*}
    \centering
    \tiny
    \includegraphics[width=0.3\linewidth]{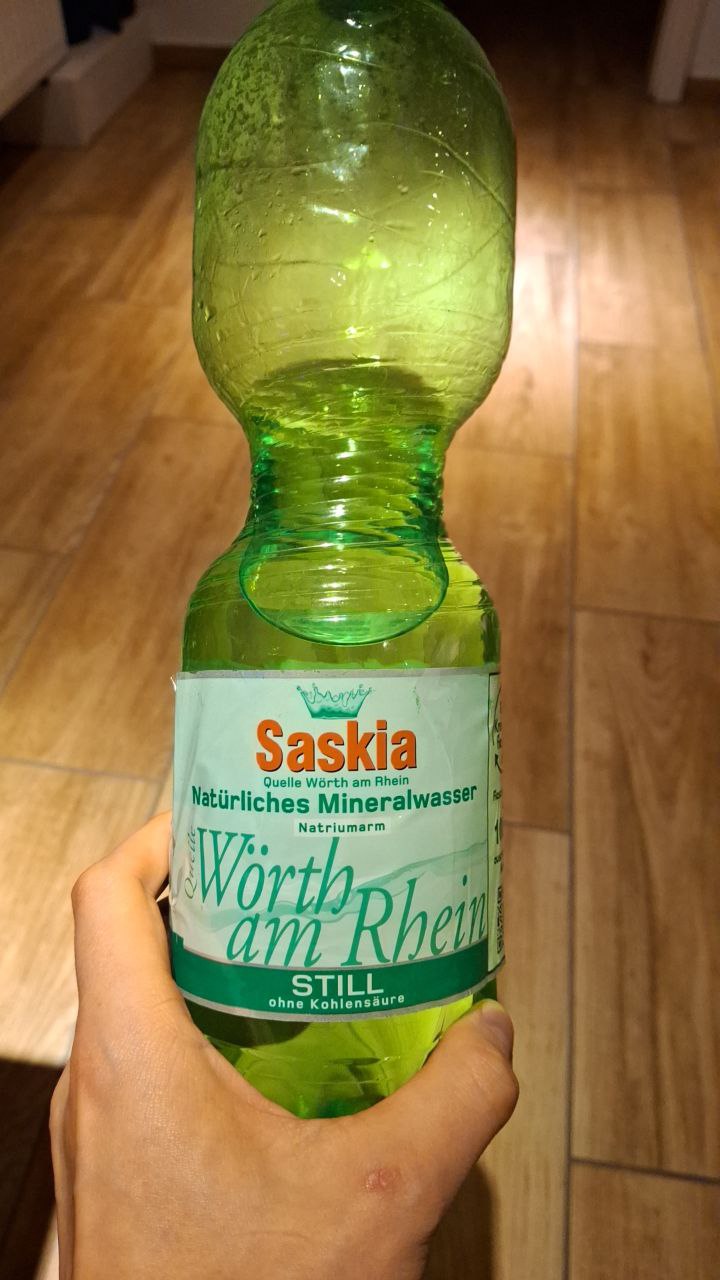}
    \begin{examplebox}{What is written on this bottle?}
    \textbf{\modelname \texttt{Aya}}: The text on the bottle, in German, reads "Sasxia Natürliches Mineralwasser" which translates to "Sasxia Natural Mineral Water." Below that, it says "Wort am Rhein," likely referring to a specific source or brand name associated with the Rhine River. The phrase "ohne Kohlensäure" indicates that the water contains no carbonation. There's also some additional information such as size ("0,5L") and possibly an expiration date or batch number at the bottom of the label, but these are not clearly visible in the image provided.
\\\\
    \textbf{\modelname \texttt{Qwen}}: The text on the label of the bottle reads "Saskia," followed by a description in German that translates to "Natural Mineral Water" and "Mineral water from Württemberg am Rhein." Below, it states "still," indicating the type of mineral water. The bottom part mentions "ohne Kohlenstoffe," which means "without carbonation," confirming that the water is still (non-carbonated). Additionally, there's a logo or emblem above the brand name "Saskia," but due to the angle and focus of the image, specific details about the logo cannot be discerned. There are also some numbers at the very top right corner of the label, likely related to product information such as batch number or size, but they are not clearly visible. The overall design suggests a natural, possibly organic origin for the mineral water. However, without clearer visibility of all elements, more detailed information would require further examination or context. If you're looking for precise nutritional facts or chemical composition, those specifics might not be available from the visual content alone. For accurate information regarding the contents or any health-related aspects of the beverage, consulting the official product packaging or manufacturer’s website would be advisable.

Please note: This analysis focuses solely on what can be seen within the provided image and does not include speculative or unverifiable claims beyond what is directly observable. Any additional interpretation goes beyond the scope of the given visual data.

If you need assistance with something else or have another question, feel free to ask!

In summary:

    Brand Name: Saskia
    Type: Natural Mineral Water
    Origin: Württemberg am Rhein
    Carbonation Status: Still (Non-Carbonated)
    Additional Information: Without Carbon Dioxide (Kohlensäure)
    Label Design Suggests Organic/Healthy Origin
    Specific Nutritional Facts or Chemical Composition Not Clearly Visible in Image

For comprehensive details, refer to the actual product packaging or manufacturer's resources.
    \end{examplebox}
    \caption{\modelname summarizing the label of a water bottle. While mostly correct, there are some OCR mistakes, e.g., with the city name ('Wörth`)}
    \label{fig:qualitative_examples:3}
\end{figure*}

\begin{figure*}
    \centering
    \tiny
    \includegraphics[width=0.5\linewidth]{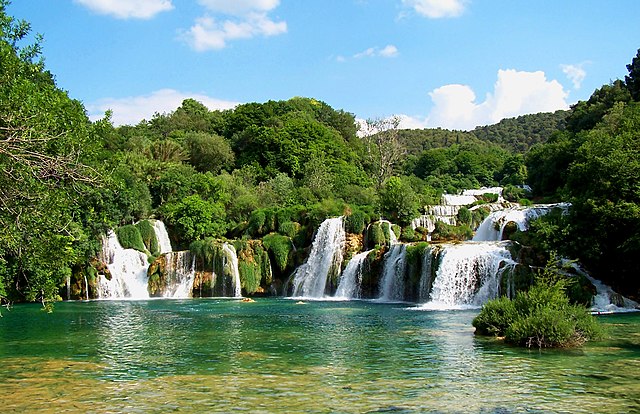}
    \begin{examplebox}{Opišite sliku što detaljnije. (\textit{Describe the image in as much detail as possible.})}
    \textbf{\modelname \texttt{Aya}}: Slika prikazuje mirno i živopisno prizoru vodopada koji se nalazi u bujnim zelenim krajoliku. Vodopad, s više padina, stvara dinamičnu scenu sa čistim plavom vodi koja kontrastira sa okolnom zelenilom. Okolište je gusto naseljen drvećem, a nekoliko oblaka može se vidjeti na nebu, dodajući dubinu sceni.
    (\textit{The image depicts a peaceful and picturesque scene of a waterfall set in a lush green landscape. The waterfall, with multiple slopes, creates a dynamic scene with its clear blue water contrasting with the surrounding greenery. The surroundings are densely populated with trees, and a few clouds can be seen in the sky, adding depth to the scene.})
\\\\
    \textbf{\modelname \texttt{Qwen}}: Slika prikazuje mirnu vodopadsku scenu, karakterizirana svojim više padavina koji se spuštaju u čisto plavo-zeleno jezero ispod njih. Voda izgleda glatko i neometano, sugerirajući dubinu i čistost. Okolišna vegetacija je bogata i živahna, ukazujući na dobro održavan ekosistem sa obiljem svježe vode. Nebo je jasno i plavo, s minimalnim oblaci, što sugeriše sunčan dan bez vidljive atmosferske promene ili udaljenog događaja. Sama vodopada nema ljudskih elemenata kao što su mostovi, klupi ili barijeri, što bi moglo značiti da je lokacija priloživa i možda nepovremena za putnike.
    (\textit{The image depicts a tranquil waterfall scene, characterized by its multiple falls cascading into a clear blue-green lake below. The water appears smooth and undisturbed, suggesting depth and purity. The surrounding vegetation is rich and vibrant, indicating a well-maintained ecosystem with an abundance of fresh water. The sky is clear and blue, with minimal clouds, suggesting a sunny day with no visible atmospheric change or distant event. The waterfall itself has no human elements such as bridges, benches, or barriers, which could indicate that the location is approachable and perhaps uninhabited by travelers.})

    \end{examplebox}
    \caption{\modelname describing an image in Croatian. Translations by Google Translate are provided in brackets for the reader.}
    \label{fig:qualitative_examples:4}
\end{figure*}

\begin{figure*}
    \centering
    \tiny
    \includegraphics[width=0.5\linewidth]{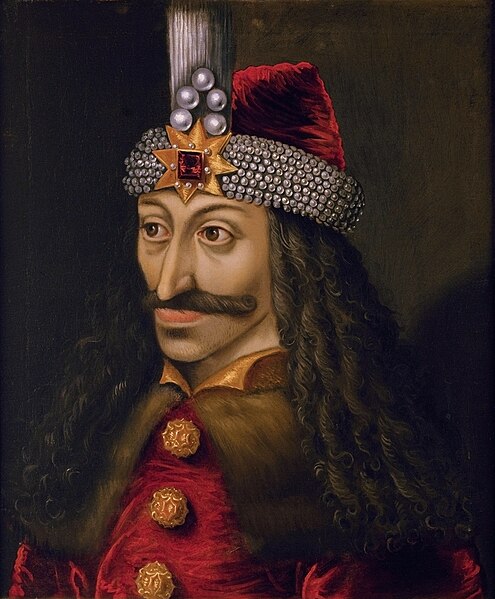}
    \begin{examplebox}{Cine este prezentat în acest portret? (\textit{Who is shown in this portrait?})}
    \textbf{\modelname \texttt{Aya}}: Portretul prezintă un om care pare a fi o figură istorică sau fictivă, probabil de rang înalt datorită îmbrăcămintei sale regale și a bijuteriilor. Stilul picturii sugerează că ar putea fi din perioada Renașterii sau barocă, având în vedere detaliile meticuloase ale costumului și tehnica picturilor cu ulei. Fără context suplimentar, nu este posibil să se identifice persoana exactă reprezentată, dar hainele și accesoriile sunt caracteristice celor care au deținut poziții semnificative în istoria europeană.
    (\textit{The portrait depicts a man who appears to be a historical or fictional figure, likely of high rank due to his regal attire and jewelry. The style of the painting suggests that it may be from the Renaissance or Baroque period, given the meticulous detail of the costume and the oil painting technique. Without further context, it is not possible to identify the exact person depicted, but the clothing and accessories are characteristic of those who held significant positions in European history.})
\\\\
    \textbf{\modelname \texttt{Qwen}}: Persoana din portret pare a fi un om de natare, probabil o figură istorică sau nobilă, având în vedere îmbrăcămintea și simbolismul capului. În mod specific, el poartă o coroană cu modele complicate, care ar putea indica putere sau statut. Stila îmbrăcămintei sugerează că ar putea fi din secolul al XVIII-lea, ceea ce indică faptul că figura poate fi asociată cu această epocă. Fără context suplimentar, nu este posibil să se determine identitatea exactă, dar caracteristicile sale distincte sugerează că ar putea fi un lider politic, un regiune sau un alt individ important dintr-o perioadă istorică. Portretul servește pentru a transmite autoritatea și statusul sa prin utilizarea elementelor visuale ale îmbrăcămintei și accesoriilor. Este important de menționat că, deși detaliul exact al persoanei este neclar, natura portretului sugerează că ea are un rol semnificativ în istoria respectivă. Portretul servește pentru a transmite autoritatea și statusul sa prin utilizarea elementelor vizuale ale îmbrăcămintei și accesoriilor.
    (\textit{The person in the portrait appears to be a man of character, perhaps a historical or noble figure, given the clothing and symbolism of the head. Specifically, he is wearing an intricately patterned crown, which could indicate power or status. The style of the clothing suggests that it may be from the 18th century, indicating that the figure can be associated with this era. Without additional context, it is not possible to determine the exact identity, but his distinctive features suggest that he may be a political leader, a regional leader, or another important individual from a historical period. The portrait serves to convey his authority and status through the use of visual elements of his clothing and accessories. It is important to note that while the exact detail of the person is unclear, the nature of the portrait suggests that he has a significant role in the history in question. The portrait serves to convey his authority and status through the use of visual elements of his clothing and accessories.})

    \end{examplebox}
    \caption{\modelname answering a question in Romanian at length. Still, neither model correctly identifies the famous portrait of Vlad III Dracula and both models are wrong with their guess of creation time (16th century). Translations by Google Translate are provided in brackets for the reader.}
    \label{fig:qualitative_examples:5}
\end{figure*}

We provide some qualitative examples of our \modelname models.
Figure \ref{fig:qualitative_examples:1}, \ref{fig:qualitative_examples:2}, and \ref{fig:qualitative_examples:3} show results with (non-English) text in images with English prompts.
Figure~\ref{fig:qualitative_examples:4} and \ref{fig:qualitative_examples:5} show examples for multilingual prompts (and responses).

\section{Full Results}
\label{sec:appendix:fullresults}

We report the full results of all tasks with all language-specific results.
\textbf{Avg.} refers to the average without English.
Metric are the same as in \S\ref{sec:appendix:evaluation}.

\subsection{Analysis Experiments}
\label{sec:appendix:fullresults:analyis}

We report the full results for all models trained for \S\ref{sec:analysis} (and also the LLMs tested for \S\ref{sec:appendix:experiments:llm}).

The following holds for all Tables:
Models of the form \textbf{`Phi 3.5 - T5 50'} are to be interpreted as using the LLM Phi 3.5 with the T5 languages and 50\% English with analog interpretation for other rows.

\textbf{`Phi 3.5 - PT 1'} means the model was pretrained with 1\% English and then instruct-tuned with the L100 50\% English mix (see \S\ref{tab:ex3}).

\textbf{`Phi 3.5 - OCR 1'} means the model was pretrained with 50\% English for the captions and 1\% English for the OCR data and then instruct-tuned with the L100 50\% English mix (see \S\ref{tab:ex4}).

\paragraph{BIN-MC}

Table~\ref{tab:full:bin-mc}

\begin{table*}[t]
    \centering
    \footnotesize
    \centering
     \def\arraystretch{0.97}
     \resizebox{0.99\linewidth}{!}{
\begin{tabular}{lrrrrrrrrrrrrrrrrrrrrr}
\toprule
 & \bf en & \bf avg. & \bf af & \bf am & \bf cs & \bf el & \bf es & \bf fa & \bf fi & \bf ha & \bf hr & \bf hu & \bf ja & \bf mi & \bf nl & \bf no & \bf pl & \bf ro & \bf ta & \bf te & \bf zu \\
\midrule
Phi 3.5 - English & 64.7 & 38.1 & 43.3 & 29.7 & 41.5 & 35.5 & 55.9 & 33.6 & 36.4 & 24.5 & 43.3 & 39.0 & 49.8 & 27.8 & 47.3 & 44.2 & 41.4 & 42.8 & 30.0 & 27.1 & 31.1 \\
Phi 3.5 - T5 50 & 66.0 & 39.6 & 46.0 & 30.3 & 43.1 & 36.3 & 56.3 & 33.4 & 36.5 & 35.1 & 45.1 & 40.7 & 50.9 & 30.1 & 48.4 & 46.2 & 41.1 & 43.1 & 31.0 & 29.6 & 29.1 \\
Phi 3.5 - T5-4 50 & 65.2 & 40.6 & 46.8 & 29.6 & 44.6 & 37.9 & 59.1 & 36.7 & 37.5 & 29.0 & 46.4 & 42.5 & 52.0 & 31.1 & 50.7 & 47.4 & 43.0 & 43.5 & 31.5 & 29.0 & 32.4 \\
Phi 3.5 - T5-3 50 & 65.5 & 40.6 & 50.0 & 28.8 & 43.3 & 37.4 & 58.6 & 34.4 & 38.4 & 33.0 & 46.1 & 41.4 & 50.9 & 31.2 & 49.7 & 47.0 & 41.9 & 43.8 & 32.5 & 29.6 & 32.7 \\
Phi 3.5 - T5-2 50 & 64.8 & 39.1 & 47.2 & 25.6 & 41.9 & 35.8 & 57.9 & 34.0 & 36.0 & 29.8 & 44.8 & 39.5 & 50.0 & 30.5 & 47.7 & 45.8 & 41.2 & 42.4 & 30.4 & 29.2 & 33.9 \\
Phi 3.5 - L100 50 & 64.7 & 39.9 & 48.1 & 28.2 & 42.8 & 36.8 & 57.2 & 34.7 & 37.0 & 28.2 & 44.7 & 40.4 & 51.2 & 31.6 & 47.8 & 46.4 & 40.9 & 43.8 & 30.6 & 30.1 & 37.5 \\
Llama 3 - English & 65.4 & 40.9 & 44.0 & 28.2 & 46.9 & 42.2 & 53.0 & 42.4 & 38.7 & 31.1 & 47.6 & 46.3 & 48.6 & 30.1 & 48.2 & 47.4 & 44.0 & 44.9 & 31.6 & 32.5 & 29.3 \\
\bottomrule
\end{tabular}
}
    \caption{BIN-MC}
    \label{tab:full:bin-mc}
\end{table*}

\paragraph{M3Exam}
Table~\ref{tab:full:m3exam}

\begin{table*}[t]
    \centering
    \footnotesize
    \centering
     \def\arraystretch{0.97}
     \resizebox{0.99\linewidth}{!}{
\begin{tabular}{lrrrrrrrr}
\toprule
 & \bf en & \bf avg. & \bf af & \bf zh & \bf it & \bf pt & \bf th & \bf vi \\
\midrule
Phi 3.5 - English & 52.9 & 32.7 & 32.5 & 37.0 & 49.6 & 39.7 & 25.4 & 12.2 \\
Phi 3.5 - T5 50 & 51.2 & 35.3 & 39.9 & 35.9 & 46.4 & 39.7 & 28.2 & 21.7 \\
Phi 3.5 - T5-4 50 & 52.2 & 34.2 & 40.5 & 32.4 & 49.1 & 38.6 & 25.2 & 19.1 \\
Phi 3.5 - T5-3 50 & 51.3 & 35.3 & 43.6 & 34.0 & 47.4 & 37.3 & 27.9 & 21.7 \\
Phi 3.5 - T5-2 50 & 49.2 & 33.7 & 39.3 & 32.9 & 45.1 & 38.4 & 22.2 & 24.3 \\
Phi 3.5 - L100 50 & 50.8 & 36.0 & 39.3 & 36.1 & 50.9 & 40.1 & 26.2 & 23.5 \\
Llama 3 - English & 46.1 & 32.5 & 38.6 & 32.6 & 41.6 & 35.0 & 25.9 & 20.9 \\
\bottomrule
\end{tabular}
}
    \caption{M3Exam}
    \label{tab:full:m3exam}
\end{table*}

\paragraph{VGR}
Table~\ref{tab:full:vgr}

\begin{table*}[t]
    \centering
    \footnotesize
    \centering
     \def\arraystretch{0.97}
     \resizebox{0.99\linewidth}{!}{
\begin{tabular}{lrrrrrrrrrrrrr}
\toprule
 & \bf en & \bf avg. & \bf am & \bf ber & \bf bn & \bf de & \bf fil & \bf ha & \bf hi & \bf ru & \bf sw & \bf th & \bf zu \\
\midrule
Phi 3.5 - English & 80.8 & 54.1 & 45.0 & 50.8 & 41.5 & 71.7 & 55.8 & 41.7 & 62.7 & 85.0 & 35.8 & 68.3 & 36.2 \\
Phi 3.5 - T5 50 & 75.8 & 50.9 & 49.2 & 49.2 & 40.7 & 72.5 & 55.0 & 42.5 & 54.2 & 60.8 & 37.5 & 60.8 & 37.9 \\
Phi 3.5 - T5-4 50 & 83.3 & 55.1 & 51.7 & 43.3 & 49.2 & 70.8 & 65.8 & 42.5 & 61.9 & 70.8 & 38.3 & 75.0 & 36.2 \\
Phi 3.5 - T5-3 50 & 83.3 & 56.6 & 43.3 & 50.8 & 50.8 & 74.2 & 69.2 & 42.5 & 57.6 & 76.7 & 43.3 & 71.7 & 42.2 \\
Phi 3.5 - T5-2 50 & 81.7 & 57.5 & 45.8 & 52.5 & 44.1 & 73.3 & 64.2 & 39.2 & 59.3 & 73.3 & 60.0 & 60.8 & 59.5 \\
Phi 3.5 - L100 50 & 76.7 & 56.4 & 46.7 & 46.7 & 54.2 & 71.7 & 60.0 & 45.0 & 57.6 & 70.8 & 57.5 & 65.8 & 44.0 \\
Llama 3 - English & 82.5 & 56.3 & 66.7 & 30.8 & 49.2 & 77.5 & 50.8 & 48.3 & 63.6 & 75.8 & 46.7 & 70.0 & 39.7 \\
\bottomrule
\end{tabular}
}
    \caption{VGR}
    \label{tab:full:vgr}
\end{table*}

\paragraph{VLOD}
Table~\ref{tab:full:vlod}

\begin{table*}[t]
    \centering
    \footnotesize
    \centering
     \def\arraystretch{0.97}
     \resizebox{0.99\linewidth}{!}{
\begin{tabular}{lrrrrrrrrrrrrr}
\toprule
 & \bf en & \bf avg. & \bf am & \bf ber & \bf bn & \bf de & \bf fil & \bf ha & \bf hi & \bf ru & \bf sw & \bf th & \bf zu \\
\midrule
Phi 3.5 - English & 16.7 & 21.3 & 20.8 & 20.8 & 19.2 & 16.7 & 25.8 & 28.3 & 17.0 & 12.5 & 25.0 & 26.7 & 22.0 \\
Phi 3.5 - T5 50 & 23.3 & 20.0 & 15.0 & 18.3 & 20.8 & 21.7 & 16.7 & 20.0 & 23.2 & 27.5 & 22.3 & 15.8 & 18.6 \\
Phi 3.5 - T5-4 50 & 17.5 & 18.2 & 19.2 & 20.8 & 13.3 & 20.8 & 17.5 & 16.7 & 21.4 & 26.7 & 16.1 & 10.0 & 17.8 \\
Phi 3.5 - T5-3 50 & 25.8 & 19.8 & 16.7 & 17.5 & 21.7 & 21.7 & 20.0 & 21.7 & 23.2 & 20.8 & 18.8 & 16.7 & 18.6 \\
Phi 3.5 - T5-2 50 & 21.7 & 20.5 & 21.7 & 18.3 & 16.7 & 22.5 & 27.5 & 27.5 & 17.9 & 21.7 & 17.0 & 13.3 & 21.2 \\
Phi 3.5 - L100 50 & 18.3 & 19.5 & 16.7 & 20.8 & 19.2 & 25.8 & 20.0 & 16.7 & 25.0 & 20.8 & 13.4 & 17.5 & 18.6 \\
Llama 3 - English & 12.5 & 20.8 & 18.3 & 21.7 & 20.0 & 10.8 & 24.2 & 29.2 & 15.2 & 12.5 & 28.6 & 29.2 & 19.5 \\
\bottomrule
\end{tabular}
}
    \caption{VLOD}
    \label{tab:full:vlod}
\end{table*}

\paragraph{MaRVL}
Table~\ref{tab:full:marvl}

\begin{table*}[t]
    \centering
    \footnotesize
    \centering
     \def\arraystretch{0.97}
     \resizebox{0.99\linewidth}{!}{
\begin{tabular}{lrrrrrrr}
\toprule
 & \bf en & \bf avg. & \bf id & \bf sw & \bf ta & \bf tr & \bf zh \\
\midrule
Phi 3.5 - English & 82.1 & 61.4 & 65.6 & 50.8 & 53.3 & 63.8 & 73.2 \\
Phi 3.5 - T5 50 & 81.5 & 61.8 & 66.4 & 53.4 & 53.7 & 61.6 & 73.8 \\
Phi 3.5 - T5-4 50 & 81.2 & 64.3 & 68.7 & 52.3 & 54.3 & 70.2 & 76.2 \\
Phi 3.5 - T5-3 50 & 81.5 & 65.9 & 70.8 & 56.4 & 56.7 & 68.9 & 76.7 \\
Phi 3.5 - T5-2 50 & 79.7 & 66.4 & 70.2 & 62.2 & 57.5 & 66.7 & 75.4 \\
Phi 3.5 - L100 50 & 79.6 & 64.4 & 69.0 & 59.0 & 53.6 & 67.5 & 73.0 \\
Llama 3 - English & 85.2 & 65.0 & 68.8 & 52.5 & 54.3 & 69.7 & 79.8 \\
\bottomrule
\end{tabular}
}
    \caption{MaRVL}
    \label{tab:full:marvl}
\end{table*}

\paragraph{MaXM}
Table~\ref{tab:full:maxm}

\begin{table*}[t]
    \centering
    \footnotesize
    \centering
     \def\arraystretch{0.97}
     \resizebox{0.99\linewidth}{!}{
\begin{tabular}{lrrrrrrrr}
\toprule
 & \bf en & \bf avg. & \bf fr & \bf hi & \bf he & \bf ro & \bf th & \bf zh \\
\midrule
Phi 3.5 - English & 53.0 & 9.2 & 14.3 & 11.9 & 7.9 & 7.2 & 7.0 & 7.2 \\
Phi 3.5 - T5 50 & 51.3 & 25.6 & 41.0 & 30.6 & 17.5 & 15.6 & 27.5 & 21.5 \\
Phi 3.5 - T5-4 50 & 51.0 & 33.1 & 45.4 & 50.7 & 27.0 & 23.7 & 32.5 & 19.5 \\
Phi 3.5 - T5-3 50 & 53.7 & 36.7 & 41.0 & 45.9 & 33.0 & 36.6 & 40.4 & 23.5 \\
Phi 3.5 - T5-2 50 & 53.4 & 35.9 & 42.3 & 48.0 & 33.3 & 35.1 & 32.8 & 23.8 \\
Phi 3.5 - L100 50 & 54.4 & 36.6 & 43.0 & 48.0 & 30.8 & 35.1 & 39.1 & 23.5 \\
Llama 3 - English & 55.4 & 7.7 & 9.2 & 10.9 & 6.7 & 4.5 & 8.3 & 6.8 \\
\bottomrule
\end{tabular}
}
    \caption{MaXM}
    \label{tab:full:maxm}
\end{table*}

\paragraph{MTVQA}
Table~\ref{tab:full:mtvqa}

\begin{table*}[t]
    \centering
    \footnotesize
    \centering
     \def\arraystretch{0.97}
     \resizebox{0.99\linewidth}{!}{
\begin{tabular}{lrrrrrrrrrr}
\toprule
 & \bf avg. & \bf ar & \bf de & \bf fr & \bf it & \bf ja & \bf ko & \bf ru & \bf th & \bf vi \\
\midrule
Phi 3.5 - English & 3.2 & 0.9 & 6.5 & 9.3 & 8.1 & 0.8 & 0.7 & 1.6 & 0.0 & 1.1 \\
Phi 3.5 - T5 50 & 5.7 & 1.7 & 12.0 & 15.9 & 10.1 & 2.4 & 3.8 & 2.6 & 0.9 & 1.8 \\
Phi 3.5 - T5-4 50 & 5.9 & 2.7 & 14.0 & 15.1 & 9.6 & 3.5 & 3.8 & 1.9 & 0.9 & 1.6 \\
Phi 3.5 - T5-3 50 & 5.8 & 2.0 & 13.5 & 14.6 & 9.4 & 3.9 & 3.8 & 2.4 & 0.9 & 2.0 \\
Phi 3.5 - T5-2 50 & 6.6 & 5.3 & 15.9 & 15.1 & 9.4 & 4.1 & 3.8 & 2.5 & 0.4 & 2.7 \\
Phi 3.5 - L100 50 & 6.3 & 2.8 & 15.8 & 16.8 & 8.9 & 3.9 & 2.7 & 2.8 & 0.4 & 2.9 \\
Llama 3 - English & 3.2 & 0.3 & 6.9 & 8.0 & 8.7 & 0.7 & 0.5 & 0.7 & 0.4 & 2.7 \\
\bottomrule
\end{tabular}
}
    \caption{MTVQA}
    \label{tab:full:mtvqa}
\end{table*}

\paragraph{xGQA}
Table~\ref{tab:full:xgqa}

\begin{table*}[t]
    \centering
    \footnotesize
    \centering
     \def\arraystretch{0.97}
     \resizebox{0.99\linewidth}{!}{
\begin{tabular}{lrrrrrrrrr}
\toprule
 & \bf en & \bf avg. & \bf bn & \bf de & \bf id & \bf ko & \bf pt & \bf ru & \bf zh \\
\midrule
Phi 3.5 - English & 59.7 & 37.2 & 4.9 & 47.8 & 33.2 & 38.2 & 47.1 & 42.1 & 47.2 \\
Phi 3.5 - T5 50 & 54.1 & 34.1 & 2.6 & 44.6 & 34.3 & 36.3 & 43.8 & 36.4 & 41.0 \\
Phi 3.5 - T5-4 50 & 52.0 & 37.4 & 5.7 & 45.6 & 38.7 & 40.4 & 45.2 & 43.4 & 42.7 \\
Phi 3.5 - T5-3 50 & 54.8 & 40.6 & 22.7 & 46.5 & 42.1 & 39.8 & 46.0 & 43.6 & 43.6 \\
Phi 3.5 - T5-2 50 & 57.8 & 45.3 & 27.4 & 50.3 & 46.0 & 46.4 & 48.6 & 49.5 & 48.9 \\
Phi 3.5 - L100 50 & 56.6 & 45.1 & 27.0 & 51.4 & 44.8 & 44.9 & 50.8 & 48.2 & 48.7 \\
Llama 3 - English & 61.9 & 39.2 & 13.2 & 49.0 & 35.6 & 39.1 & 44.9 & 44.1 & 48.4 \\
\bottomrule
\end{tabular}
}
    \caption{xGQA}
    \label{tab:full:xgqa}
\end{table*}

\paragraph{XM3600}
Table~\ref{tab:full:xm3600}. Language fidelity for \S\ref{sec:analysis:ex1} in Table~\ref{tab:full:xm3600_fidelity}.

\begin{table*}[t]
    \centering
    \footnotesize
    \centering
     \def\arraystretch{0.97}
     \resizebox{0.99\linewidth}{!}{
\begin{tabular}{lrrrrrrrrrrrrrrrrrrrrrrrrrrrrrrrrrrrrr}
\toprule
 & \bf en & \bf avg. & \bf ar & \bf bn & \bf cs & \bf da & \bf de & \bf el & \bf es & \bf fa & \bf fi & \bf fil & \bf fr & \bf he & \bf hi & \bf hr & \bf hu & \bf id & \bf it & \bf ja & \bf ko & \bf mi & \bf nl & \bf no & \bf pl & \bf pt & \bf quz & \bf ro & \bf ru & \bf sv & \bf sw & \bf te & \bf th & \bf tr & \bf uk & \bf vi & \bf zh \\
\midrule
Phi 3.5 - English & 33.6 & 1.2 & 0.0 & 0.0 & 0.7 & 1.1 & 1.7 & 0.0 & 10.5 & 0.0 & 0.4 & 1.6 & 4.4 & 0.0 & 0.0 & 0.5 & 0.6 & 1.5 & 9.2 & 0.1 & 0.0 & 0.0 & 1.6 & 1.5 & 0.7 & 1.9 & 0.1 & 0.7 & 0.4 & 1.1 & 0.6 & 0.0 & 0.2 & 0.3 & 0.1 & 0.6 & 0.0 \\
Phi 3.5 - T5 50 & 33.0 & 9.5 & 7.8 & 0.6 & 3.9 & 8.2 & 24.7 & 0.6 & 34.4 & 0.4 & 1.8 & 1.9 & 39.1 & 3.4 & 3.5 & 2.6 & 4.0 & 7.9 & 28.5 & 27.6 & 1.6 & 0.1 & 20.4 & 8.8 & 4.8 & 30.2 & 0.7 & 5.4 & 17.5 & 11.8 & 1.3 & 0.0 & 4.0 & 3.2 & 5.7 & 2.6 & 12.8 \\
Phi 3.5 - T5-4 50 & 25.2 & 11.8 & 6.9 & 1.0 & 13.8 & 10.8 & 24.4 & 1.4 & 27.3 & 8.6 & 7.0 & 3.6 & 31.3 & 3.5 & 9.7 & 9.2 & 10.5 & 8.4 & 24.9 & 27.9 & 3.1 & 2.1 & 21.7 & 12.0 & 14.1 & 24.0 & 0.5 & 6.0 & 22.5 & 24.1 & 2.1 & 0.0 & 5.8 & 9.4 & 4.6 & 18.8 & 10.6 \\
Phi 3.5 - T5-3 50 & 32.7 & 13.6 & 6.5 & 5.9 & 13.8 & 18.2 & 25.4 & 7.0 & 31.0 & 7.1 & 5.7 & 11.9 & 30.7 & 7.6 & 7.2 & 9.4 & 8.2 & 22.9 & 26.0 & 27.3 & 3.0 & 1.8 & 28.5 & 14.1 & 13.6 & 22.1 & 0.4 & 11.0 & 16.7 & 23.8 & 1.5 & 0.0 & 12.9 & 7.6 & 10.0 & 15.9 & 20.2 \\
Phi 3.5 - T5-2 50 & 29.9 & 11.1 & 4.5 & 5.5 & 10.7 & 12.4 & 20.4 & 6.4 & 20.4 & 5.7 & 5.5 & 10.1 & 24.8 & 7.4 & 6.7 & 8.0 & 7.1 & 17.5 & 18.5 & 27.0 & 2.5 & 2.0 & 18.9 & 11.2 & 10.0 & 18.1 & 0.4 & 7.6 & 21.1 & 17.8 & 7.0 & 0.0 & 12.6 & 7.9 & 10.3 & 12.7 & 9.5 \\
Phi 3.5 - L100 50 & 31.0 & 13.2 & 5.6 & 3.3 & 10.9 & 18.0 & 26.4 & 4.5 & 30.9 & 4.1 & 4.4 & 11.1 & 38.5 & 6.7 & 6.3 & 7.3 & 7.8 & 22.4 & 30.6 & 23.7 & 2.5 & 2.6 & 24.2 & 18.8 & 10.8 & 29.0 & 1.1 & 8.1 & 18.6 & 17.8 & 8.1 & 1.5 & 10.2 & 7.2 & 8.3 & 14.7 & 15.6 \\
\bottomrule
\end{tabular}
}
    \caption{XM3600}
    \label{tab:full:xm3600}
\end{table*}

\begin{table*}[t]
    \centering
    \footnotesize
    \centering
     \def\arraystretch{0.97}
     \resizebox{0.99\linewidth}{!}{
\begin{tabular}{lrrrrrrrrrrrrrrrrrrrrrrrrrrrrrrrrrrrr}
\toprule
 & \bf ar & \bf bn & \bf cs & \bf da & \bf de & \bf el & \bf en & \bf es & \bf fa & \bf fi & \bf fil & \bf fr & \bf he & \bf hi & \bf hr & \bf hu & \bf id & \bf it & \bf ja & \bf ko & \bf mi & \bf nl & \bf no & \bf pl & \bf pt & \bf quz & \bf ro & \bf ru & \bf sv & \bf sw & \bf te & \bf th & \bf tr & \bf uk & \bf vi & \bf zh \\
\midrule
Phi 3.5 - L100 & 93.8 & 99.0 & 99.0 & 91.8 & 100.0 & 88.3 & 100.0 & 100.0 & 98.8 & 100.0 & 96.7 & 100.0 & 99.6 & 97.7 & 76.2 & 100.0 & 95.3 & 100.0 & 99.2 & 99.8 & 100.0 & 100.0 & 98.4 & 100.0 & 97.7 & 0.2 & 100.0 & 99.8 & 98.6 & 98.2 & 93.2 & 99.8 & 100.0 & 92.6 & 100.0 & 96.5 \\
Phi 3.5 - T5-2 & 94.7 & 100.0 & 99.4 & 98.4 & 100.0 & 99.8 & 100.0 & 100.0 & 100.0 & 100.0 & 99.6 & 100.0 & 100.0 & 98.8 & 79.1 & 100.0 & 88.5 & 100.0 & 99.8 & 100.0 & 97.3 & 100.0 & 85.7 & 100.0 & 99.2 & 27.5 & 100.0 & 99.8 & 99.6 & 99.0 & 63.5 & 100.0 & 100.0 & 96.9 & 100.0 & 93.9 \\
Phi 3.5 - T5-3 & 92.2 & 99.8 & 99.2 & 97.9 & 100.0 & 99.8 & 100.0 & 100.0 & 100.0 & 100.0 & 99.6 & 100.0 & 99.8 & 99.0 & 77.9 & 100.0 & 91.8 & 100.0 & 100.0 & 100.0 & 95.9 & 100.0 & 75.2 & 100.0 & 52.1 & 37.5 & 100.0 & 100.0 & 99.2 & 84.4 & 82.8 & 100.0 & 100.0 & 96.5 & 100.0 & 95.3 \\
Phi 3.5 - T5-4 & 96.5 & 96.1 & 99.2 & 78.3 & 100.0 & 98.4 & 100.0 & 100.0 & 100.0 & 100.0 & 99.6 & 100.0 & 99.4 & 98.6 & 89.8 & 100.0 & 98.0 & 100.0 & 100.0 & 100.0 & 95.7 & 100.0 & 71.5 & 100.0 & 100.0 & 12.7 & 100.0 & 100.0 & 100.0 & 84.6 & 67.2 & 99.0 & 100.0 & 30.9 & 100.0 & 94.1 \\
Phi 3.5 - T5 & 98.2 & 97.3 & 78.7 & 87.5 & 100.0 & 50.2 & 100.0 & 99.8 & 2.3 & 76.4 & 49.4 & 100.0 & 96.7 & 99.2 & 73.6 & 99.2 & 95.1 & 100.0 & 100.0 & 99.8 & 1.6 & 99.8 & 92.0 & 96.5 & 100.0 & 0.2 & 96.3 & 100.0 & 98.6 & 36.1 & 62.5 & 90.2 & 94.9 & 91.6 & 39.8 & 96.7 \\
\bottomrule
\end{tabular}
}
    \caption{XM3600 language fidelity (\S\ref{tab:ex1:fidelity})}
    \label{tab:full:xm3600_fidelity}
\end{table*}

\paragraph{XVNLI}
Table~\ref{tab:full:xvnli}

\begin{table*}[t]
    \centering
    \footnotesize
    \centering
     \def\arraystretch{0.97}
     \resizebox{0.99\linewidth}{!}{
\begin{tabular}{lrrrrrr}
\toprule
 & \bf en & \bf avg. & \bf ar & \bf es & \bf fr & \bf ru \\
\midrule
Phi 3.5 - English & 59.6 & 55.0 & 52.3 & 54.9 & 57.6 & 55.2 \\
Phi 3.5 - T5 50 & 59.9 & 51.8 & 49.7 & 51.3 & 55.2 & 51.0 \\
Phi 3.5 - T5-4 50 & 58.9 & 48.3 & 47.7 & 47.1 & 51.2 & 47.3 \\
Phi 3.5 - T5-3 50 & 58.6 & 50.5 & 46.6 & 51.3 & 52.6 & 51.3 \\
Phi 3.5 - T5-2 50 & 58.5 & 53.6 & 50.7 & 54.7 & 55.1 & 54.0 \\
Phi 3.5 - L100 50 & 59.6 & 53.3 & 49.9 & 53.7 & 56.8 & 52.6 \\
Llama 3 - English & 46.1 & 36.3 & 33.4 & 37.0 & 36.5 & 38.2 \\
\bottomrule
\end{tabular}
}
    \caption{XVNLI}
    \label{tab:full:xvnli}
\end{table*}

\paragraph{xMMMU}
Table~\ref{tab:full:xmmmu}

\begin{table*}[t]
    \centering
    \footnotesize
    \centering
     \def\arraystretch{0.97}
     \resizebox{0.99\linewidth}{!}{
\begin{tabular}{lrrrrrrrr}
\toprule
 & \bf en & \bf avg. & \bf ar & \bf fr & \bf hi & \bf id & \bf ja & \bf pt \\
\midrule
Phi 3.5 - English & 38.4 & 36.2 & 36.2 & 41.9 & 29.9 & 35.4 & 34.2 & 39.7 \\
Phi 3.5 - T5 50 & 36.7 & 36.2 & 31.5 & 38.9 & 31.6 & 37.0 & 34.9 & 43.4 \\
Phi 3.5 - T5-4 50 & 37.0 & 33.9 & 33.2 & 39.9 & 29.2 & 32.3 & 31.2 & 37.7 \\
Phi 3.5 - T5-3 50 & 37.3 & 35.8 & 32.5 & 39.3 & 32.3 & 37.0 & 36.8 & 37.0 \\
Phi 3.5 - T5-2 50 & 37.6 & 35.1 & 32.2 & 40.3 & 32.6 & 34.7 & 32.3 & 38.7 \\
Phi 3.5 - L100 50 & 36.6 & 32.0 & 28.5 & 35.9 & 27.8 & 32.0 & 31.2 & 36.7 \\
Llama 3 - English & 33.2 & 32.4 & 30.9 & 34.2 & 30.6 & 32.7 & 30.5 & 35.7 \\
\bottomrule
\end{tabular}
}
    \caption{xMMMU}
    \label{tab:full:xmmmu}
\end{table*}

\paragraph{SMPQA - Ground}
Table~\ref{tab:full:smpqa_ground}

\begin{table*}[t]
    \centering
    \footnotesize
    \centering
     \def\arraystretch{0.97}
     \resizebox{0.99\linewidth}{!}{
\begin{tabular}{lrrrrrrrrrrrrrrrrrrrrrrrrrr}
\toprule
 & \bf en & \bf avg. & \bf avg. Latin & \bf avg. other & \bf ar & \bf de & \bf hi & \bf id & \bf it & \bf ko & \bf ru & \bf th & \bf zh & \bf zu \\
\midrule
Phi 3.5 - English & 65.8 & 55.8 & 62.3 & 51.5 & 50.2 & 63.5 & 58.5 & 61.4 & 64.0 & 49.0 & 52.1 & 49.1 & 49.8 & 60.2 \\
Phi 3.5 - T5 50 & 75.2 & 60.2 & 70.9 & 53.1 & 50.2 & 70.8 & 65.4 & 71.8 & 71.6 & 49.8 & 54.1 & 51.0 & 48.0 & 69.4 \\
Phi 3.5 - T5-4 50 & 74.2 & 60.8 & 71.4 & 53.7 & 52.2 & 71.5 & 65.5 & 72.8 & 73.1 & 51.1 & 53.9 & 49.6 & 49.6 & 68.4 \\
Phi 3.5 - T5-3 50 & 70.4 & 58.7 & 67.7 & 52.8 & 51.6 & 66.9 & 61.0 & 69.6 & 67.2 & 50.0 & 53.6 & 48.9 & 51.4 & 67.0 \\
Phi 3.5 - T5-2 50 & 68.4 & 56.2 & 64.2 & 50.8 & 49.5 & 64.5 & 58.4 & 65.4 & 64.9 & 50.0 & 50.5 & 48.8 & 47.9 & 62.0 \\
Phi 3.5 - L100 50 & 69.6 & 58.0 & 67.2 & 51.9 & 49.9 & 68.0 & 62.4 & 69.0 & 67.9 & 48.6 & 52.5 & 49.6 & 48.4 & 64.1 \\
Llama 3 - English & 72.0 & 60.5 & 69.6 & 54.4 & 53.5 & 69.9 & 67.2 & 71.1 & 70.9 & 48.9 & 57.5 & 50.1 & 49.4 & 66.5 \\
\bottomrule
\end{tabular}
}
    \caption{SMPQA Ground}
    \label{tab:full:smpqa_ground}
\end{table*}

\paragraph{SMPQA - Name}
Table~\ref{tab:full:smpqa_name}

\begin{table*}[t]
    \centering
    \footnotesize
    \centering
     \def\arraystretch{0.97}
     \resizebox{0.99\linewidth}{!}{
\begin{tabular}{lrrrrrrrrrrrrrrrrrrrrrrrrrr}
\toprule
 & \bf en & \bf avg. & \bf avg. Latin & \bf avg. other & \bf ar & \bf de & \bf hi & \bf id & \bf it & \bf ko & \bf ru & \bf th & \bf zh & \bf zu \\
\midrule
Phi 3.5 - English & 36.2 & 5.0 & 12.4 & 0.0 & 0.0 & 17.4 & 0.0 & 12.6 & 15.2 & 0.0 & 0.0 & 0.0 & 0.0 & 4.4 \\
Phi 3.5 - T5 50 & 36.4 & 5.4 & 13.6 & 0.0 & 0.0 & 21.2 & 0.0 & 13.2 & 16.0 & 0.0 & 0.0 & 0.0 & 0.0 & 3.8 \\
Phi 3.5 - T5-4 50 & 35.0 & 5.8 & 14.4 & 0.0 & 0.0 & 20.0 & 0.0 & 14.6 & 16.6 & 0.0 & 0.0 & 0.0 & 0.0 & 6.4 \\
Phi 3.5 - T5-3 50 & 34.6 & 5.8 & 14.4 & 0.0 & 0.0 & 16.0 & 0.0 & 16.6 & 20.4 & 0.0 & 0.0 & 0.0 & 0.0 & 4.8 \\
Phi 3.5 - T5-2 50 & 35.8 & 5.8 & 14.5 & 0.0 & 0.0 & 18.0 & 0.0 & 14.8 & 19.6 & 0.0 & 0.0 & 0.0 & 0.0 & 5.6 \\
Phi 3.5 - L100 50 & 33.4 & 5.2 & 12.8 & 0.1 & 0.0 & 17.4 & 0.0 & 14.0 & 14.6 & 0.0 & 0.2 & 0.2 & 0.0 & 5.2 \\
Llama 3 - English & 41.0 & 8.5 & 21.1 & 0.0 & 0.0 & 24.4 & 0.0 & 21.6 & 23.8 & 0.0 & 0.0 & 0.2 & 0.0 & 14.8 \\
\bottomrule
\end{tabular}
}
    \caption{SMPQA Name}
    \label{tab:full:smpqa_name}
\end{table*}

\subsection{Comparison with \modelname}
\label{sec:appendix:fullresults:baselines}

\paragraph{BIN-MC}
Table~\ref{tab:full:baseline:bin-mc}

\begin{table*}[t]
    \centering
    \footnotesize
    \centering
     \def\arraystretch{0.97}
     \resizebox{0.99\linewidth}{!}{
\begin{tabular}{lrrrrrrrrrrrrrrrrrrrrr}
\toprule
 & \bf en & \bf avg. & \bf af & \bf am & \bf cs & \bf el & \bf es & \bf fa & \bf fi & \bf ha & \bf hr & \bf hu & \bf ja & \bf mi & \bf nl & \bf no & \bf pl & \bf ro & \bf ta & \bf te & \bf zu \\
\midrule
\modelname Aya & 69.7 & 54.7 & 63.6 & 29.4 & \underline{66.2} & \textbf{67.8} & 65.1 & \textbf{60.0} & 43.3 & \textbf{37.5} & 63.6 & 49.8 & 66.7 & \underline{37.0} & 62.4 & 59.1 & \underline{62.6} & \underline{64.0} & \underline{46.9} & 50.9 & \underline{42.6} \\
\modelname Qwen & 72.7 & \underline{56.2} & \underline{65.3} & \textbf{47.4} & 62.2 & 56.7 & \underline{67.0} & 53.6 & 48.8 & \underline{36.7} & 65.4 & 54.1 & \underline{67.6} & \textbf{39.1} & \underline{63.7} & 63.6 & 60.4 & 58.5 & 45.2 & \underline{63.4} & \textbf{49.5} \\
Parrot & 30.5 & 25.7 & 26.0 & 22.8 & 26.1 & 25.5 & 27.3 & 25.9 & 26.4 & 23.7 & 25.3 & 25.6 & 26.7 & 25.4 & 28.0 & 26.6 & 26.5 & 26.8 & 25.5 & 23.9 & 24.0 \\
PALO 13B & 61.4 & 41.1 & 48.4 & 25.9 & 47.9 & 35.8 & 53.2 & 37.5 & 42.7 & 26.1 & 52.3 & 47.9 & 49.1 & 31.0 & 48.9 & 51.2 & 46.1 & 46.5 & 28.9 & 32.2 & 28.3 \\
PALO 7B & 58.7 & 38.6 & 44.2 & 28.4 & 43.6 & 33.5 & 49.9 & 36.9 & 39.1 & 24.5 & 49.6 & 45.4 & 48.8 & 27.8 & 45.1 & 45.8 & 42.0 & 44.0 & 26.7 & 30.1 & 28.3 \\
InternVL 2.5 4B & 68.4 & 45.4 & 53.2 & 31.3 & 53.2 & 42.3 & 60.8 & 45.4 & 38.3 & 26.3 & 55.2 & 42.1 & 60.5 & 29.5 & 56.6 & 53.7 & 53.1 & 49.7 & 35.3 & 50.1 & 26.5 \\
InternVL 2.5 8B & 70.3 & 44.2 & 54.4 & 29.1 & 52.8 & 43.3 & 57.8 & 40.5 & 41.3 & 25.8 & 55.6 & 44.9 & 57.3 & 30.0 & 51.8 & 54.8 & 50.3 & 48.9 & 33.2 & 41.2 & 27.3 \\
\bottomrule
\end{tabular}
}
    \caption{BIN-MC}
    \label{tab:full:baseline:bin-mc}
\end{table*}

\paragraph{M3Exam}
Table~\ref{tab:full:baseline:m3exam}

\begin{table*}[t]
    \centering
    \footnotesize
    \centering
     \def\arraystretch{0.97}
     \resizebox{0.99\linewidth}{!}{
\begin{tabular}{lrrrrrrrr}
\toprule
 & \bf en & \bf avg. & \bf af & \bf zh & \bf it & \bf pt & \bf th & \bf vi \\
\midrule
\modelname Aya & 53.0 & 41.2 & \underline{52.8} & 51.4 & 47.7 & 27.4 & 27.8 & 40.3 \\
\modelname Qwen & 61.2 & 46.9 & 50.9 & 55.6 & 49.0 & 31.9 & 29.6 & 64.1 \\
Parrot & 46.6 & 36.2 & 38.0 & 37.8 & 36.8 & 25.9 & 23.5 & 55.1 \\
PALO 13B & 45.2 & 28.3 & 33.1 & 31.3 & 36.5 & 19.3 & 20.2 & 29.2 \\
PALO 7B & 41.0 & 29.1 & 34.4 & 31.5 & 32.7 & 21.8 & 21.1 & 33.4 \\
InternVL 2.5 4B & \underline{63.2} & \underline{50.3} & 46.0 & \underline{60.9} & \underline{50.3} & 34.9 & \underline{39.1} & 70.4 \\
InternVL 2.5 8B & \textbf{67.0} & \textbf{53.3} & \textbf{57.7} & \textbf{61.7} & \textbf{53.2} & 33.0 & \underline{39.1} & \textbf{75.2} \\
Qwen2-VL 2B & 47.9 & 40.5 & 38.0 & 51.6 & 36.4 & \underline{36.2} & 26.1 & 54.9 \\
Qwen2-VL 7B & 56.1 & 49.7 & 50.9 & 58.6 & 46.8 & 34.7 & 38.3 & 69.0 \\
Maya & 49.2 & 36.3 & 48.5 & 46.4 & 36.6 & 25.9 & 20.0 & 40.3 \\
Phi 3.5 Vision & 56.3 & 40.7 & 51.5 & 54.4 & 44.1 & 25.2 & 24.3 & 44.4 \\
Pixtral 12B & 49.4 & 33.7 & 39.9 & 53.6 & 34.4 & 19.5 & 7.0 & 47.7 \\
Pangea & 58.0 & 45.5 & 50.3 & 58.6 & 49.0 & 32.2 & 27.8 & 55.3 \\
MiniCPM 2.6 & 55.0 & 48.2 & 44.2 & 54.6 & 44.3 & \textbf{36.9} & 38.3 & \underline{70.8} \\

\bottomrule
\end{tabular}
}
    \caption{M3Exam}
    \label{tab:full:baseline:m3exam}
\end{table*}

\paragraph{VGR}
Table~\ref{tab:full:baseline:vgr}

\begin{table*}[t]
    \centering
    \footnotesize
    \centering
     \def\arraystretch{0.97}
     \resizebox{0.99\linewidth}{!}{
\begin{tabular}{lrrrrrrrrrrrrr}
\toprule
 & \bf en & \bf avg. & \bf am & \bf ber & \bf bn & \bf de & \bf fil & \bf ha & \bf hi & \bf ru & \bf sw & \bf th & \bf zu \\
\midrule
\modelname Aya & 82.5 & \underline{66.8} & \underline{71.7} & 54.2 & \underline{59.3} & 73.3 & \underline{59.2} & \underline{65.0} & \underline{71.2} & 75.8 & \underline{67.5} & 72.5 & \underline{65.5} \\
\modelname Qwen & \underline{87.5} & \textbf{73.1} & \textbf{77.5} & 49.2 & \textbf{62.7} & \textbf{80.8} & \textbf{78.3} & \textbf{76.7} & \textbf{72.9} & \underline{85.0} & \textbf{70.0} & \textbf{81.7} & \textbf{69.0} \\
Parrot & 59.2 & 52.9 & 45.0 & 64.2 & 53.4 & 63.3 & 49.2 & 41.7 & 62.7 & 62.5 & 35.8 & 67.5 & 36.2 \\
PALO 13B & 63.3 & 26.2 & 25.0 & 55.0 & 0.8 & 44.2 & 47.5 & 40.0 & 0.0 & 5.8 & 32.5 & 0.0 & 37.1 \\
PALO 7B & 48.3 & 25.6 & 40.8 & \underline{75.0} & 0.0 & 0.0 & 49.2 & 40.0 & 0.0 & 0.0 & 39.2 & 0.0 & 37.9 \\
InternVL 2.5 4B & 72.5 & 49.7 & 43.3 & 50.0 & 40.7 & 62.5 & 56.7 & 41.7 & 42.4 & 63.3 & 35.8 & 74.2 & 36.2 \\
InternVL 2.5 8B & \underline{87.5} & 51.6 & 43.3 & 50.0 & 41.5 & 64.2 & 49.2 & 41.7 & 59.3 & 75.8 & 36.7 & 68.3 & 37.1 \\
Qwen2-VL 2B & 61.7 & 50.5 & 44.2 & 50.0 & 43.2 & 65.0 & 53.3 & 41.7 & 61.0 & 52.5 & 38.3 & 67.5 & 38.8 \\
Qwen2-VL 7B & 60.0 & 52.9 & 48.3 & 50.0 & 46.6 & 60.0 & 50.0 & 46.7 & 48.3 & 63.3 & 58.3 & 60.8 & 49.1 \\
Maya & 46.7 & 42.3 & 43.3 & 48.3 & 33.9 & 50.8 & 51.7 & 40.8 & 42.4 & 45.8 & 34.2 & 38.3 & 35.3 \\
Phi 3.5 Vision & 81.7 & 50.3 & 45.8 & 49.2 & 56.8 & \underline{73.3} & 54.2 & 41.7 & 56.8 & \textbf{85.8} & 38.3 & 15.0 & 36.2 \\
Pixtral 12B & 55.8 & 47.7 & 51.7 & 32.5 & 47.5 & 63.3 & 51.7 & 44.2 & 16.1 & 54.2 & 65.8 & 53.3 & 44.0 \\
Pangea & 69.2 & 58.9 & 45.8 & \textbf{90.0} & 53.4 & 61.7 & 55.0 & 41.7 & 60.2 & 74.2 & 54.2 & \underline{75.8} & 36.2 \\
MiniCPM 2.6 & 52.5 & 49.1 & 45.0 & 55.8 & 49.2 & 45.8 & 48.3 & 40.8 & 44.1 & 59.2 & 48.3 & 65.8 & 37.9 \\
\bottomrule
\end{tabular}
}
    \caption{VGR}
    \label{tab:full:baseline:vgr}
\end{table*}

\paragraph{VLOD}
Table~\ref{tab:full:baseline:vlod}

\begin{table*}[t]
    \centering
    \footnotesize
    \centering
     \def\arraystretch{0.97}
     \resizebox{0.99\linewidth}{!}{
\begin{tabular}{lrrrrrrrrrrrrr}
\toprule
 & \bf en & \bf avg. & \bf am & \bf ber & \bf bn & \bf de & \bf fil & \bf ha & \bf hi & \bf ru & \bf sw & \bf th & \bf zu \\
\midrule
\modelname Aya & 12.5 & 20.7 & 18.3 & 21.7 & 20.0 & 11.7 & 24.2 & \underline{29.2} & 15.2 & 10.8 & 28.6 & 29.2 & 19.5 \\
\modelname Qwen & 28.3 & 27.0 & 18.3 & 20.0 & \textbf{33.3} & 32.5 & 29.2 & 22.5 & 25.0 & 22.5 & \underline{30.4} & \underline{30.0} & \textbf{33.1} \\
Parrot & 0.0 & 0.0 & 0.0 & 0.0 & 0.0 & 0.0 & 0.0 & 0.0 & 0.0 & 0.0 & 0.0 & 0.0 & 0.0 \\
PALO 13B & 2.5 & 4.9 & 6.7 & 5.0 & 6.7 & 5.0 & 5.8 & 2.5 & 3.6 & 4.2 & 5.4 & 5.0 & 4.2 \\
PALO 7B & 5.8 & 6.8 & 8.3 & 9.2 & 10.0 & 5.8 & 6.7 & 4.2 & 9.8 & 5.0 & 4.5 & 5.8 & 5.1 \\
InternVL 2.5 4B & 24.2 & 21.0 & 18.3 & \underline{26.7} & 17.5 & 20.8 & 20.0 & 23.3 & 22.3 & 20.0 & 23.2 & 20.0 & 18.6 \\
InternVL 2.5 8B & \textbf{57.5} & \underline{29.0} & \underline{25.0} & 22.5 & \underline{25.8} & \textbf{38.3} & \textbf{36.7} & 25.8 & \textbf{41.1} & \textbf{35.8} & 15.2 & \underline{30.0} & 22.9 \\
Qwen2-VL 2B & 22.5 & 20.4 & 17.5 & 20.0 & 13.3 & 26.7 & 25.0 & 24.2 & 20.5 & 16.7 & 21.4 & 15.8 & 23.7 \\
Qwen2-VL 7B & 5.8 & 13.2 & 14.2 & 15.8 & 13.3 & 11.7 & 10.0 & 15.0 & 12.5 & 12.5 & 13.4 & 13.3 & 13.6 \\
Maya & 20.0 & 20.1 & 20.0 & 25.8 & 19.2 & 20.8 & 15.0 & 25.8 & 17.9 & 23.3 & 21.4 & 15.8 & 16.1 \\
Phi 3.5 Vision & \underline{45.8} & \textbf{31.5} & \textbf{27.5} & \textbf{29.2} & 23.3 & \underline{36.7} & \underline{30.0} & \textbf{31.7} & \underline{33.9} & \underline{29.2} & \textbf{37.5} & \textbf{35.8} & \underline{31.4} \\
Pixtral 12B & 9.2 & 12.4 & 17.5 & 13.3 & 10.0 & 16.7 & 10.0 & 16.7 & 3.6 & 14.2 & 8.9 & 12.5 & 13.6 \\
Pangea & 0.0 & 6.7 & 0.0 & 0.8 & 0.0 & 20.8 & 24.2 & 15.8 & 6.2 & 0.8 & 3.6 & 0.8 & 0.8 \\
MiniCPM 2.6 & 9.2 & 14.6 & 11.7 & 19.2 & 12.5 & 10.8 & 10.0 & 22.5 & 10.7 & 12.5 & 19.6 & 11.7 & 19.5 \\
\bottomrule
\end{tabular}
}
    \caption{VLOD}
    \label{tab:full:baseline:vlod}
\end{table*}

\paragraph{MaRVL}
Table~\ref{tab:full:baseline:marvl}

\begin{table*}[t]
    \centering
    \footnotesize
    \centering
     \def\arraystretch{0.97}
     \resizebox{0.99\linewidth}{!}{
\begin{tabular}{lrrrrrrr}
\toprule
 & \bf en & \bf avg. & \bf id & \bf sw & \bf ta & \bf tr & \bf zh \\
\midrule
\modelname Aya & \underline{85.0} & \underline{77.9} & \underline{79.5} & \underline{70.9} & \underline{73.4} & \underline{83.4} & \underline{82.4} \\
\modelname Qwen & \textbf{89.6} & \textbf{81.7} & \textbf{85.0} & \textbf{76.8} & \textbf{76.0} & \textbf{84.2} & \textbf{86.7} \\
Parrot & 63.5 & 55.1 & 56.6 & 51.2 & 50.7 & 58.6 & 58.2 \\
PALO 13B & 63.8 & 33.1 & 58.7 & 50.9 & 2.6 & 53.1 & 0.2 \\
PALO 7B & 62.7 & 24.1 & 33.6 & 47.8 & 0.4 & 38.5 & 0.0 \\
InternVL 2.5 4B & 74.9 & 59.0 & 65.7 & 50.7 & 50.9 & 64.2 & 63.5 \\
InternVL 2.5 8B & 83.0 & 63.3 & 63.2 & 51.4 & 54.6 & 67.2 & 79.9 \\
Qwen2-VL 2B & 67.9 & 55.9 & 60.9 & 51.8 & 52.2 & 59.0 & 55.8 \\
Qwen2-VL 7B & 69.8 & 60.2 & 61.1 & 53.1 & 60.9 & 65.3 & 60.7 \\
Maya & 60.3 & 56.3 & 60.3 & 50.7 & 50.6 & 58.9 & 61.2 \\
Phi 3.5 Vision & 73.4 & 46.4 & 56.4 & 51.3 & 50.8 & 58.0 & 15.7 \\
Pixtral 12B & 67.7 & 60.7 & 62.5 & 54.4 & 61.8 & 65.5 & 59.1 \\
Pangea & 75.8 & 70.5 & 74.3 & 70.9 & 66.6 & 71.1 & 69.6 \\
MiniCPM 2.6 & 70.2 & 57.9 & 57.8 & 54.2 & 57.2 & 63.3 & 57.2 \\
\bottomrule
\end{tabular}
}
    \caption{MaRVL}
    \label{tab:full:baseline:marvl}
\end{table*}

\paragraph{MaXM}
Table~\ref{tab:full:baseline:maxm}

\begin{table*}[t]
    \centering
    \footnotesize
    \centering
     \def\arraystretch{0.97}
     \resizebox{0.99\linewidth}{!}{
\begin{tabular}{lrrrrrrrr}
\toprule
 & \bf en & \bf avg. & \bf fr & \bf hi & \bf he & \bf ro & \bf th & \bf zh \\
\midrule
\modelname Aya & 55.7 & \underline{49.3} & 45.1 & \underline{58.7} & \textbf{62.9} & \underline{51.1} & 46.7 & 31.6 \\
\modelname Qwen & \underline{60.1} & 47.7 & 47.1 & 45.1 & \underline{56.8} & 47.7 & \underline{57.0} & 32.2 \\
Parrot & 28.2 & 3.6 & 2.7 & 2.9 & 1.4 & 1.2 & 3.0 & 10.7 \\
PALO 13B & 51.7 & 33.1 & 42.0 & 17.5 & 53.4 & 34.2 & 20.9 & 30.6 \\
PALO 7B & 54.0 & 22.5 & 39.9 & 9.2 & 30.6 & 16.8 & 12.3 & 26.4 \\
InternVL 2.5 4B & 46.0 & 42.5 & 45.7 & 37.1 & 38.8 & 31.5 & 51.0 & \underline{50.8} \\
InternVL 2.5 8B & 45.6 & 38.2 & \textbf{51.2} & 27.9 & 24.5 & 35.7 & 36.4 & \textbf{53.4} \\
Qwen2-VL 2B & 53.7 & 26.5 & 40.3 & 10.8 & 9.5 & 15.6 & 38.1 & 44.6 \\
Qwen2-VL 7B & 54.7 & 31.2 & 38.6 & 18.7 & 13.9 & 37.2 & 42.1 & 36.8 \\
Maya & 55.4 & 17.3 & 19.1 & 13.0 & 21.1 & 18.0 & 11.6 & 20.8 \\
Llama-Vision & 0.0 & 4.7 & 0.0 & 0.6 & 2.4 & 0.3 & 24.8 & 0.0 \\
Phi 3.5 Vision & 43.6 & 17.9 & 23.5 & 12.1 & 16.3 & 7.8 & 20.9 & 27.0 \\
Pixtral 12B & 59.4 & 43.4 & 46.8 & 31.7 & 54.4 & 44.1 & 44.4 & 39.1 \\
Pangea & \textbf{61.4} & \textbf{55.0} & \underline{47.4} & \textbf{61.0} & 53.7 & \textbf{52.9} & \textbf{67.2} & 47.9 \\
MiniCPM 2.6 & 53.4 & 22.3 & 14.3 & 12.1 & 5.1 & 19.5 & 53.6 & 29.3 \\

\bottomrule
\end{tabular}
}
    \caption{MaXM}
    \label{tab:full:baseline:maxm}
\end{table*}

\paragraph{MTVQA}
Table~\ref{tab:full:baseline:mtvqa}

\begin{table*}[t]
    \centering
    \footnotesize
    \centering
     \def\arraystretch{0.97}
     \resizebox{0.99\linewidth}{!}{
\begin{tabular}{lrrrrrrrrrr}
\toprule
 & \bf avg. & \bf ar & \bf de & \bf fr & \bf it & \bf ja & \bf ko & \bf ru & \bf th & \bf vi \\
\midrule
\modelname Aya & 11.1 & 6.7 & 19.9 & 22.5 & 16.7 & 5.0 & 9.0 & 5.2 & 5.2 & 9.7 \\
\modelname Qwen & 11.9 & 4.6 & 22.7 & 26.5 & 18.6 & 5.9 & 9.9 & 5.0 & 5.2 & 8.9 \\
Parrot & 2.0 & 1.4 & 1.9 & 0.9 & 1.6 & 1.6 & 2.7 & 2.0 & 5.2 & 0.9 \\
PALO 13B & 6.3 & 2.6 & 15.6 & 12.1 & 10.4 & 4.0 & 4.3 & 4.0 & 0.0 & 4.2 \\
PALO 7B & 5.8 & 1.8 & 14.3 & 13.3 & 8.3 & 3.4 & 3.2 & 3.6 & 0.4 & 4.1 \\
InternVL 2.5 4B & \textbf{25.1} & 11.2 & \textbf{34.4} & \textbf{38.4} & 33.5 & \underline{18.4} & \underline{29.0} & 9.8 & \underline{16.5} & \textbf{34.6} \\
InternVL 2.5 8B & \underline{25.0} & \underline{11.5} & \underline{33.8} & \underline{37.4} & \textbf{35.3} & \textbf{19.7} & \textbf{30.3} & \underline{10.4} & \underline{16.5} & 30.4 \\
Qwen2-VL 2B & 19.0 & 6.1 & 26.8 & 30.9 & 30.7 & 13.5 & 21.1 & 9.3 & 10.0 & 22.4 \\
Qwen2-VL 7B & 23.2 & \textbf{16.9} & 27.3 & 31.7 & \underline{35.2} & 16.1 & 24.6 & \textbf{10.8} & 15.6 & \underline{30.7} \\
Maya & 5.3 & 2.8 & 13.1 & 12.2 & 6.6 & 2.8 & 4.8 & 2.9 & 0.4 & 2.3 \\
Llama-Vision & 15.2 & 7.4 & 24.0 & 18.7 & 25.3 & 9.4 & 14.5 & 6.1 & 15.2 & 15.8 \\
Phi 3.5 Vision & 11.1 & 3.3 & 18.2 & 20.2 & 25.2 & 5.6 & 8.8 & 5.4 & 3.0 & 10.5 \\
Pixtral 12B & 14.1 & 4.3 & 25.7 & 27.3 & 25.2 & 5.9 & 9.1 & 7.5 & 5.2 & 16.6 \\
Pangea & 19.3 & 8.3 & 29.5 & 35.2 & 29.2 & 9.3 & 14.5 & 7.4 & 10.8 & 29.2 \\
MiniCPM 2.6 & 16.1 & 2.3 & 23.9 & 27.5 & 32.7 & 11.7 & 12.7 & 7.3 & 10.0 & 16.5 \\
\bottomrule
\end{tabular}
}
    \caption{MTVQA}
    \label{tab:full:baseline:mtvqa}
\end{table*}

\paragraph{xGQA}
Table~\ref{tab:full:baseline:xgqa}

\begin{table*}[t]
    \centering
    \footnotesize
    \centering
     \def\arraystretch{0.97}
     \resizebox{0.99\linewidth}{!}{
\begin{tabular}{lrrrrrrrrr}
\toprule
 & \bf en & \bf avg. & \bf bn & \bf de & \bf id & \bf ko & \bf pt & \bf ru & \bf zh \\
\midrule
Parrot & 37.7 & 21.2 & 20.2 & 23.2 & 19.8 & 22.8 & 21.7 & 19.7 & 21.2 \\
PALO 13B & 58.0 & 27.8 & 26.3 & 14.7 & 29.6 & 30.9 & 17.8 & 30.9 & 44.1 \\
PALO 7B & 59.1 & 36.6 & 42.8 & 34.5 & 30.0 & 40.8 & 27.7 & 32.2 & 47.9 \\
InternVL 2.5 4B & 63.6 & 28.0 & 28.1 & 29.2 & 15.4 & 38.3 & 27.2 & 31.5 & 25.9 \\
InternVL 2.5 8B & 63.4 & 32.0 & 17.4 & 23.8 & 25.0 & 38.2 & 27.6 & 36.4 & 55.2 \\
Qwen2-VL 2B & 60.5 & 38.2 & 18.6 & 43.2 & 32.6 & 39.0 & 39.9 & 44.1 & 50.3 \\
Qwen2-VL 7B & 62.5 & 49.3 & 37.4 & 51.1 & 48.4 & 50.3 & 51.8 & 52.1 & 54.1 \\
Maya & 58.2 & 49.1 & 40.1 & 53.2 & 49.7 & 47.2 & 52.5 & 50.6 & 50.1 \\
Llama-Vision & 39.3 & 27.6 & 26.0 & 29.2 & 26.8 & 24.9 & 27.9 & 30.7 & 27.9 \\
Phi 3.5 Vision & \textbf{65.2} & 38.0 & 5.0 & 51.9 & 37.3 & 35.6 & 50.6 & 45.9 & 39.5 \\
Pixtral 12B & 59.9 & 3.8 & 0.7 & 5.4 & 14.0 & 0.3 & 3.6 & 0.4 & 1.9 \\
Pangea & \underline{64.6} & \textbf{60.4} & \textbf{59.1} & \textbf{61.6} & \textbf{60.7} & \textbf{58.8} & \textbf{62.1} & \textbf{60.7} & \textbf{59.6} \\
MiniCPM 2.6 & 57.9 & 45.7 & 33.9 & 49.0 & 46.3 & 42.1 & 51.0 & 48.7 & 48.6 \\
\bottomrule
\end{tabular}
}
    \caption{xGQA}
    \label{tab:full:baseline:xgqa}
\end{table*}

\paragraph{XM3600}
Table~\ref{tab:full:baseline:xm3600}. Language fidelity Table~\ref{tab:full:baseline:xm3600_fidelity}.

\begin{table*}[t]
    \centering
    \footnotesize
    \centering
     \def\arraystretch{0.97}
     \resizebox{0.99\linewidth}{!}{
\begin{tabular}{lrrrrrrrrrrrrrrrrrrrrrrrrrrrrrrrrrrrrr}
\toprule
 & \bf en & \bf avg. & \bf ar & \bf bn & \bf cs & \bf da & \bf de & \bf el & \bf es & \bf fa & \bf fi & \bf fil & \bf fr & \bf he & \bf hi & \bf hr & \bf hu & \bf id & \bf it & \bf ja & \bf ko & \bf mi & \bf nl & \bf no & \bf pl & \bf pt & \bf quz & \bf ro & \bf ru & \bf sv & \bf sw & \bf te & \bf th & \bf tr & \bf uk & \bf vi & \bf zh \\
\midrule
\modelname Aya & 78.4 & \textbf{39.2} & \textbf{40.4} & 18.5 & \textbf{33.9} & \underline{40.0} & \underline{38.6} & \textbf{35.3} & \textbf{69.7} & \textbf{55.8} & \underline{11.0} & \textbf{34.0} & \textbf{71.3} & \textbf{47.1} & \textbf{26.3} & \textbf{24.9} & \textbf{19.6} & \underline{58.3} & \textbf{60.4} & \underline{49.1} & \textbf{21.3} & \underline{33.7} & \textbf{61.7} & \underline{42.5} & \textbf{37.9} & 59.3 & \underline{1.7} & \underline{34.6} & 38.0 & \textbf{45.9} & 29.9 & \underline{15.1} & 26.0 & \textbf{30.6} & \textbf{30.6} & \underline{72.7} & \textbf{56.9} \\
\modelname Qwen & \underline{79.1} & 34.4 & \underline{36.6} & 17.1 & \underline{29.7} & \textbf{43.1} & 32.0 & \underline{19.2} & \underline{69.2} & \underline{31.2} & \textbf{12.0} & \underline{33.6} & 67.6 & 27.6 & 20.3 & \underline{22.0} & \underline{18.7} & 50.4 & \underline{53.7} & 43.5 & 13.4 & \textbf{34.9} & \underline{56.2} & 41.4 & 30.0 & \underline{59.9} & \textbf{2.1} & 23.4 & \underline{39.2} & 42.7 & \underline{30.2} & 13.5 & \underline{42.3} & 23.3 & 20.3 & 69.4 & 33.8 \\
Parrot & 5.6 & 0.4 & 0.6 & 0.0 & 0.2 & 0.0 & 0.0 & 0.0 & 3.3 & 2.3 & 0.0 & 0.0 & 0.3 & 0.0 & 0.0 & 0.0 & 0.0 & 0.0 & 0.2 & 0.0 & 0.0 & 0.0 & 1.6 & 0.0 & 0.0 & 4.0 & 0.2 & 0.4 & 0.4 & 0.0 & 0.0 & 0.8 & 0.0 & 1.0 & 0.0 & 0.0 & 0.0 \\
PALO 13B & 67.3 & 17.0 & 23.5 & \underline{22.9} & 7.9 & 30.3 & 32.4 & 0.2 & 57.0 & 1.5 & 6.6 & 8.4 & 66.2 & 0.6 & \underline{25.0} & 9.9 & 2.7 & 22.7 & 40.4 & 19.7 & 0.2 & 0.3 & 36.5 & 31.0 & 9.1 & 13.8 & 0.8 & 14.5 & 21.3 & 33.9 & 0.8 & 0.0 & 0.5 & 0.6 & 2.6 & 15.6 & 37.0 \\
\bottomrule
\end{tabular}
}
    \caption{XM3600}
    \label{tab:full:baseline:xm3600}
\end{table*}

\begin{table*}[t]
    \centering
    \footnotesize
    \centering
     \def\arraystretch{0.97}
     \resizebox{0.99\linewidth}{!}{
\begin{tabular}{lrrrrrrrrrrrrrrrrrrrrrrrrrrrrrrrrrrrrr}
\toprule
 & \bf en & \bf avg. & \bf ar & \bf bn & \bf cs & \bf da & \bf de & \bf el & \bf es & \bf fa & \bf fi & \bf fil & \bf fr & \bf he & \bf hi & \bf hr & \bf hu & \bf id & \bf it & \bf ja & \bf ko & \bf mi & \bf nl & \bf no & \bf pl & \bf pt & \bf quz & \bf ro & \bf ru & \bf sv & \bf sw & \bf te & \bf th & \bf tr & \bf uk & \bf vi & \bf zh \\
\midrule
\modelname Aya & \underline{100.0} & \underline{95.7} & 93.6 & \underline{100.0} & 97.7 & \textbf{96.7} & \underline{100.0} & \underline{100.0} & 99.8 & \underline{100.0} & \underline{99.8} & \underline{100.0} & \underline{100.0} & \underline{99.8} & \underline{99.6} & 84.6 & 99.8 & \underline{99.2} & 99.6 & 98.8 & \underline{100.0} & 98.8 & \underline{100.0} & \textbf{97.3} & \underline{99.8} & \underline{100.0} & 1.8 & \underline{100.0} & 99.6 & 98.8 & \underline{90.6} & \underline{100.0} & 99.6 & \underline{100.0} & \underline{99.8} & \underline{100.0} & 89.6 \\
\modelname Qwen & 99.8 & 95.2 & 95.1 & \underline{100.0} & \underline{98.6} & 93.9 & \underline{100.0} & \underline{100.0} & 99.4 & \underline{100.0} & \textbf{100.0} & \underline{100.0} & \underline{100.0} & 98.8 & 99.0 & 80.9 & \underline{100.0} & 96.7 & 99.8 & 98.8 & \underline{100.0} & \textbf{100.0} & \underline{100.0} & \underline{95.7} & 99.6 & 99.4 & 3.7 & \underline{100.0} & 99.4 & 98.2 & 86.5 & \underline{100.0} & 99.8 & 99.6 & 99.2 & \underline{100.0} & 86.5 \\
Parrot & \underline{100.0} & 25.0 & \textbf{100.0} & 0.0 & 0.0 & 0.0 & 0.0 & 0.0 & \underline{100.0} & \underline{100.0} & 0.0 & 0.0 & 0.0 & 0.0 & 0.0 & 0.0 & 0.0 & 0.0 & 0.0 & 0.0 & 0.0 & 0.0 & \underline{100.0} & 0.0 & 0.0 & \underline{100.0} & 0.0 & 0.0 & \underline{100.0} & 0.0 & 0.0 & \underline{100.0} & 0.0 & \underline{100.0} & 0.0 & 0.0 & 0.0 \\
\bottomrule
\end{tabular}
}
    \caption{XM3600 Language Fidelity}
    \label{tab:full:baseline:xm3600_fidelity}
\end{table*}

\paragraph{XVNLI}
Table~\ref{tab:full:baseline:xvnli}

\begin{table*}[t]
    \centering
    \footnotesize
    \centering
     \def\arraystretch{0.97}
     \resizebox{0.99\linewidth}{!}{
\begin{tabular}{lrrrrrr}
\toprule
 & \bf en & \bf avg. & \bf ar & \bf es & \bf fr & \bf ru \\
\midrule
\modelname Aya & 65.0 & 62.4 & 61.7 & 61.0 & 64.3 & 62.7 \\
\modelname Qwen & \textbf{75.4} & \textbf{70.2} & \textbf{68.8} & \textbf{70.9} & \textbf{70.5} & \textbf{70.8} \\
Parrot & 28.7 & 31.4 & 34.0 & 24.3 & 30.0 & 37.4 \\
PALO 13B & 56.6 & 53.6 & 51.8 & 52.7 & 54.9 & 55.0 \\
PALO 7B & 58.0 & 53.4 & 52.5 & 52.3 & 53.7 & 55.1 \\
InternVL 2.5 4B & 69.0 & 58.7 & 55.7 & 58.8 & 61.4 & 59.0 \\
InternVL 2.5 8B & \underline{73.5} & \underline{66.4} & 61.8 & \underline{68.0} & \underline{68.4} & \underline{67.3} \\
Qwen2-VL 2B & 61.9 & 56.2 & 52.9 & 55.3 & 58.6 & 57.9 \\
Qwen2-VL 7B & 62.1 & 59.6 & 59.2 & 58.9 & 60.0 & 60.3 \\
Maya & 50.1 & 43.9 & 45.3 & 42.7 & 45.8 & 41.8 \\
Phi 3.5 Vision & 58.9 & 53.3 & 49.7 & 52.7 & 56.4 & 54.3 \\
Pixtral 12B & 60.9 & 52.7 & 36.0 & 57.9 & 59.0 & 58.1 \\
Pangea & 69.0 & 65.2 & \underline{64.5} & 64.3 & 66.3 & 65.7 \\
MiniCPM 2.6 & 71.9 & 65.4 & 61.1 & 67.5 & 67.0 & 66.1 \\
\bottomrule
\end{tabular}
}
    \caption{XVNLI}
    \label{tab:full:baseline:xvnli}
\end{table*}

\paragraph{xMMMU}
Table~\ref{tab:full:baseline:xmmmu}

\begin{table*}[t]
    \centering
    \footnotesize
    \centering
     \def\arraystretch{0.97}
     \resizebox{0.99\linewidth}{!}{
\begin{tabular}{lrrrrrrrr}
\toprule
 & \bf en & \bf avg. & \bf ar & \bf fr & \bf hi & \bf id & \bf ja & \bf pt \\
\midrule
\modelname Aya & 37.6 & 37.2 & 36.2 & 38.9 & 38.8 & 39.7 & 34.2 & 35.4 \\
\modelname Qwen & 46.4 & \underline{43.0} & 39.6 & 45.0 & \textbf{41.6} & \underline{44.1} & 43.5 & 44.1 \\
Parrot & 35.3 & 32.4 & 31.9 & 34.9 & 26.1 & 31.3 & 34.9 & 35.4 \\
PALO 13B & 32.4 & 28.9 & 24.2 & 34.9 & 24.2 & 31.6 & 26.4 & 32.3 \\
PALO 7B & 31.8 & 30.9 & 28.2 & 33.6 & 27.3 & 30.6 & 32.3 & 33.3 \\
InternVL 2.5 4B & \underline{49.2} & 42.7 & \textbf{41.6} & \underline{45.6} & 33.7 & 43.4 & \underline{44.2} & \underline{47.8} \\
InternVL 2.5 8B & \textbf{50.7} & \textbf{45.2} & \underline{40.3} & \textbf{48.7} & \underline{41.2} & 43.1 & \textbf{47.6} & \textbf{50.2} \\
Qwen2-VL 2B & 36.8 & 35.5 & 31.5 & 41.3 & 30.2 & 36.7 & 36.1 & 37.0 \\
Qwen2-VL 7B & 43.0 & 40.7 & 36.9 & 42.6 & 38.5 & 41.1 & 41.3 & 43.8 \\
Maya & 37.9 & 33.3 & 32.6 & 36.6 & 31.3 & 31.6 & 32.0 & 36.0 \\
Phi 3.5 Vision & 41.7 & 37.4 & 34.9 & 44.3 & 29.2 & 37.7 & 35.7 & 42.4 \\
Pixtral 12B & 30.3 & 26.2 & 19.1 & 28.5 & 19.2 & 27.3 & 28.6 & 34.7 \\
Pangea & 43.1 & 42.0 & 37.6 & 43.0 & 38.5 & \textbf{46.8} & 41.6 & 44.8 \\
MiniCPM 2.6 & 39.1 & 36.5 & 30.5 & 38.9 & 33.7 & 37.7 & 37.2 & 40.7 \\
\bottomrule
\end{tabular}
}
    \caption{xMMMU}
    \label{tab:full:baseline:xmmmu}
\end{table*}

\paragraph{CVQA}
Table~\ref{tab:full:baseline:baseline:cvqa}

\begin{table*}[t]
    \centering
    \footnotesize
    \centering
     \def\arraystretch{0.97}
     \resizebox{0.99\linewidth}{!}{
\begin{tabular}{lllllllllllllllllllllllllllllllllllllllll}
\toprule
 & \bf avg. & \bf amh-ethiopia & \bf arz-egypt & \bf ben-india & \bf bre-france & \bf bul-bulgaria & \bf fil-philippines & \bf gle-ireland & \bf hin-india & \bf ibo-nigeria & \bf ind-indonesia & \bf jav-indonesia & \bf jpn-japan & \bf kin-rwanda & \bf kor-south korea & \bf mar-india & \bf min-indonesia & \bf mon-mongolia & \bf msa-malaysia & \bf nor-norway & \bf orm-ethiopia & \bf por-brazil & \bf ron-romania & \bf rus-russia & \bf sin-sri lanka & \bf spa-argentina & \bf spa-chile & \bf spa-colombia & \bf spa-ecuador & \bf spa-mexico & \bf spa-spain & \bf spa-uruguay & \bf sun-indonesia & \bf swa-kenya & \bf tam-india & \bf tel-india & \bf urd-india & \bf urd-pakistan & \bf zho-china & \bf zho-singapore \\
\midrule
\modelname Aya & 49.4 & 32.1 & \underline{52.7} & 45.8 & 30.4 & \underline{50.1} & 48.8 & 41.7 & \textbf{67.2} & 31.0 & 53.6 & 41.1 & \underline{44.8} & 32.8 & 61.7 & \underline{56.9} & 42.6 & 29.2 & 52.7 & 55.5 & \underline{36.4} & 65.1 & \underline{61.6} & 65.5 & 28.9 & 60.0 & 58.5 & 56.0 & 55.2 & 52.6 & 68.2 & 40.3 & 42.0 & 50.2 & \underline{49.5} & 37.0 & 47.3 & 50.5 & 64.6 & 65.6 \\
\modelname Qwen & \underline{52.9} & \textbf{38.0} & \underline{52.7} & \underline{54.9} & 30.6 & 49.6 & \underline{51.7} & \underline{48.5} & 65.7 & 30.5 & \underline{54.9} & \underline{46.1} & \underline{44.8} & \textbf{42.1} & \underline{66.2} & 55.9 & 41.8 & 33.3 & \underline{55.9} & \underline{58.2} & 35.0 & \textbf{70.8} & 57.0 & \underline{67.5} & \underline{42.7} & \underline{63.8} & \underline{66.2} & \underline{59.8} & 58.8 & \textbf{61.9} & 70.4 & 42.5 & 41.5 & \underline{56.4} & 43.0 & 50.5 & \underline{52.7} & \underline{56.9} & 71.1 & \textbf{73.1} \\
\bottomrule
\end{tabular}
}
    \caption{CVQA}
    \label{tab:full:baseline:baseline:cvqa}
\end{table*}

\paragraph{SMPQA - Ground}
Table~\ref{tab:full:baseline:smpqa_ground}

\begin{table*}[t]
    \centering
    \footnotesize
    \centering
     \def\arraystretch{0.97}
     \resizebox{0.99\linewidth}{!}{
\begin{tabular}{lrrrrrrrrrrrrrrrrrrrrrrrrrr}
\toprule
 & \bf en & \bf avg. & \bf avg. Latin & \bf avg. other & \bf ar & \bf de & \bf hi & \bf id & \bf it & \bf ko & \bf ru & \bf th & \bf zh & \bf zu \\
\midrule
\modelname Aya & 83.1 & 74.2 & 80.9 & 69.7 & 75.9 & 82.1 & 80.1 & 81.4 & 80.6 & 68.8 & 73.5 & 66.5 & 53.4 & 79.5 \\
\modelname Qwen & 84.8 & 76.1 & 82.7 & 71.8 & \underline{76.9} & 83.5 & 82.4 & 83.8 & 83.1 & 72.4 & 75.6 & 64.4 & 58.9 & 80.2 \\
Parrot & 51.0 & 49.9 & 50.5 & 49.5 & 50.4 & 51.6 & 49.6 & 51.0 & 49.8 & 50.4 & 50.5 & 48.2 & 47.8 & 49.5 \\
PALO 13B & 54.0 & 51.5 & 52.7 & 50.7 & 50.9 & 53.2 & 51.2 & 52.5 & 52.8 & 51.0 & 49.5 & 51.0 & 50.7 & 52.1 \\
PALO 7B & 55.5 & 52.8 & 55.4 & 51.0 & 50.4 & 56.9 & 51.0 & 55.0 & 54.1 & 51.6 & 51.1 & 51.4 & 50.2 & 55.8 \\
InternVL 2.5 4B & 87.0 & 78.3 & 86.9 & 72.6 & 54.9 & 87.6 & 59.8 & 87.0 & 88.2 & 89.4 & 86.4 & 55.1 & 90.4 & 84.8 \\
InternVL 2.5 8B & 91.0 & 79.2 & 88.7 & 72.8 & 55.8 & 89.8 & 54.9 & 89.1 & 89.1 & \underline{92.5} & 86.9 & 53.1 & 93.6 & 86.9 \\
\bottomrule
\end{tabular}
}
    \caption{SMPQA Ground}
    \label{tab:full:baseline:smpqa_ground}
\end{table*}

\paragraph{SMPQA - Name}
Table~\ref{tab:full:baseline:smpqa_name}

\begin{table*}[t]
    \centering
    \footnotesize
    \centering
     \def\arraystretch{0.97}
     \resizebox{0.99\linewidth}{!}{
\begin{tabular}{lrrrrrrrrrrrrrrrrrrrrrrrrrr}
\toprule
 & \bf en & \bf avg. & \bf avg. Latin & \bf avg. other & \bf ar & \bf de & \bf hi & \bf id & \bf it & \bf ko & \bf ru & \bf th & \bf zh & \bf zu \\
\midrule
\modelname Aya & 60.0 & 30.1 & 49.8 & 17.0 & \textbf{29.2} & 50.2 & 17.6 & 52.6 & 51.2 & 11.2 & 38.2 & 4.8 & 0.8 & 45.2 \\
\modelname Qwen & 65.2 & 31.7 & 54.3 & 16.6 & \underline{21.4} & 53.2 & 21.4 & 55.4 & 56.6 & 16.2 & 34.8 & 5.2 & 0.6 & 52.2 \\
Parrot & 0.0 & 0.0 & 0.0 & 0.1 & 0.0 & 0.0 & 0.0 & 0.0 & 0.0 & 0.4 & 0.0 & 0.0 & 0.0 & 0.0 \\
PALO 13B & 25.6 & 4.0 & 9.9 & 0.1 & 0.0 & 12.0 & 0.0 & 10.2 & 12.4 & 0.4 & 0.0 & 0.0 & 0.0 & 5.0 \\
PALO 7B & 22.4 & 2.7 & 6.7 & 0.1 & 0.0 & 8.4 & 0.0 & 7.0 & 7.0 & 0.4 & 0.0 & 0.0 & 0.0 & 4.4 \\
InternVL 2.5 4B & 77.8 & 47.5 & 67.7 & 34.0 & 0.0 & 71.0 & 0.0 & 69.8 & 69.6 & \underline{69.0} & 54.4 & 0.2 & 80.2 & 60.4 \\
InternVL 2.5 8B & 80.6 & \underline{48.2} & 68.1 & 34.9 & 0.0 & 69.2 & 0.0 & 70.4 & 70.8 & 67.2 & 61.2 & 0.2 & 80.8 & 62.2 \\
\bottomrule
\end{tabular}
}
    \caption{SMPQA Name}
    \label{tab:full:baseline:smpqa_name}
\end{table*}

\end{document}